\pdfoutput=1
\documentclass{article}

\usepackage{iclr2027_conference_arxiv,times}

\usepackage[utf8]{inputenc}
\usepackage[T1]{fontenc}
\usepackage[hidelinks]{hyperref}
\usepackage{url}
\usepackage{booktabs}
\usepackage{amsfonts}
\usepackage{amsmath}
\usepackage{amssymb}
\usepackage{bbm}
\usepackage{nicefrac}       
\usepackage{microtype}      
\usepackage[table]{xcolor}
\usepackage{graphicx}
\usepackage{subcaption}
\usepackage{float}
\usepackage{appendix}
\usepackage{cleveref}
\crefname{appendix}{Appendix}{Appendices}
\Crefname{appendix}{Appendix}{Appendices}
\usepackage{enumitem}
\usepackage{wrapfig}
\usepackage{multirow}
\usepackage[most]{tcolorbox}
\usepackage{colortbl}
\definecolor{rowgray}{gray}{0.95}
\definecolor{taxonomyfamilyblue}{RGB}{237,242,249}
\definecolor{taxonomyblockblue}{RGB}{210,223,240}
\definecolor{probdown}{RGB}{21,128,61}
\definecolor{probup}{RGB}{185,28,28}
\newcommand{\pdown}[1]{_{\textcolor{probdown}{(#1)}}}
\newcommand{\pup}[1]{_{\textcolor{probup}{(#1)}}}
\usepackage{adjustbox}
\usepackage{array}
\newcommand{\setuplbl}[1]{\raisebox{0.9ex}[0pt][0pt]{#1}}
\newcommand{\passkrule}{\arrayrulecolor{black!20}\cmidrule{2-9}\arrayrulecolor{black}}
\newlength{\defaultbelowrulesep}
\newcommand{\deltapad}{\rule{0pt}{16.5pt}}
\newcommand{\passkruledelta}{%
  \noalign{\global\belowrulesep=0pt}\passkrule
  \noalign{\global\belowrulesep=\defaultbelowrulesep}}
\newtcolorbox{promptbox}{
  colback=gray!5,
  colframe=black!60,
  boxrule=0.4pt,
  arc=2pt,
  left=6pt,
  right=6pt,
  top=5pt,
  bottom=5pt
}

\title{Cliff Tokens: Analyzing Failure Trigger Tokens in LLM Mathematical Reasoning}

\author{
  Jaeyong Ko$^1$ \quad
  Jinu Lee$^2$ \quad
  Pilsung Kang$^1$ \quad
  Yukyung Lee$^3$\thanks{Corresponding author.} \\
  \\
  $^1$Seoul National University \\
  $^2$University of Illinois Urbana-Champaign \\
  $^3$Boston University \\
  \texttt{\{jyko22, pilsung\_kang\}@snu.ac.kr} \\
  \texttt{jinulee2@illinois.edu} \quad
  \texttt{ylee5@bu.edu}
}

\iclrfinalcopy

\begin{document}

\maketitle
\begin{abstract}
Large language models reach high accuracy in mathematical reasoning, but individual traces on the same problem diverge; some arrive at the correct answer while others fail. Prior work localizes such failures at the step, chunk, or sentence level, or identifies tokens where failure has already occurred. These approaches leave open which token triggers failure. We introduce the \textbf{cliff token}, a token at which the estimated probability of reaching the correct answer (success probability) drops beyond an adaptive threshold. Across seven models and three mathematical reasoning benchmarks (GSM1K, MATH500, AIME 2025), cliff tokens act as failure triggers. For incorrect traces containing cliff tokens, we compare resampling immediately before and after the first cliff token. Resampling before it shows higher pass@$k$ at the same sample count. We further introduce a cliff taxonomy of deterministic, uncertain, and sampled-off cliffs, defined by greedy choice and token entropy. Additionally, we show that the three types differ as training signals. Using single-token preference optimization at cliff positions (Cliff-DPO), we find that uncertain and sampled-off cliffs show larger accuracy gains than deterministic cliffs on three evaluation benchmarks. We release token-level rollout data and source code to enable further analysis without regenerating costly rollouts \url{https://github.com/beaver-22/Cliff-token}.
\end{abstract}

\section{Introduction}
\label{sec:introduction}

Large language models (LLMs) have shown strong performance on mathematical reasoning tasks \citep{CoT, DeepSeekMath, Deepseek-R1, Qwen2.5coder, openaio1}. Still, individual reasoning traces can fail to reach the correct answer despite the underlying LLM being capable of solving the problem (as reflected in the gap between pass@$1$ and pass@$k$) \citep{selfconsistency, StableReasoning}. Such failures do not necessarily stem from a global lack of capability, but from specific, critical tokens that shift a reasoning path toward an incorrect outcome \citep{CriticalTokens, 8020Rule, FailureDynamics}. Despite their impact, how to identify these tokens and analyze their characteristics remains an open question.

Existing works explore this question from three directions: (i) \textit{Broader granularity:} Several reasoning analysis studies focus on coarse-grained units such as reasoning steps \citep{CoTUnderstanding,ReasoningDrift}, chunks \citep{ThePotential, BeyondtheLastAnswer} or sentences \citep{ThoughtAnchors, MeasuringFaithfulness}, rather than at the token level at which generation occurs. (ii) \textit{Post-failure states:} While \citet{CriticalTokens} flags tokens where the estimated probability of reaching the correct answer from the current prefix reaches zero, the preceding token that triggers this drop has not been identified. (iii) \textit{Absolute thresholds:} Prior studies often use fixed thresholds for probability drops (e.g., $0.3$ in \citet{ThePotential} or $0.2$ in \citet{Phi4}), without distinguishing statistically significant shifts from rollout noise.

\begin{figure}[t]
  \centering
  \includegraphics[width=\linewidth]{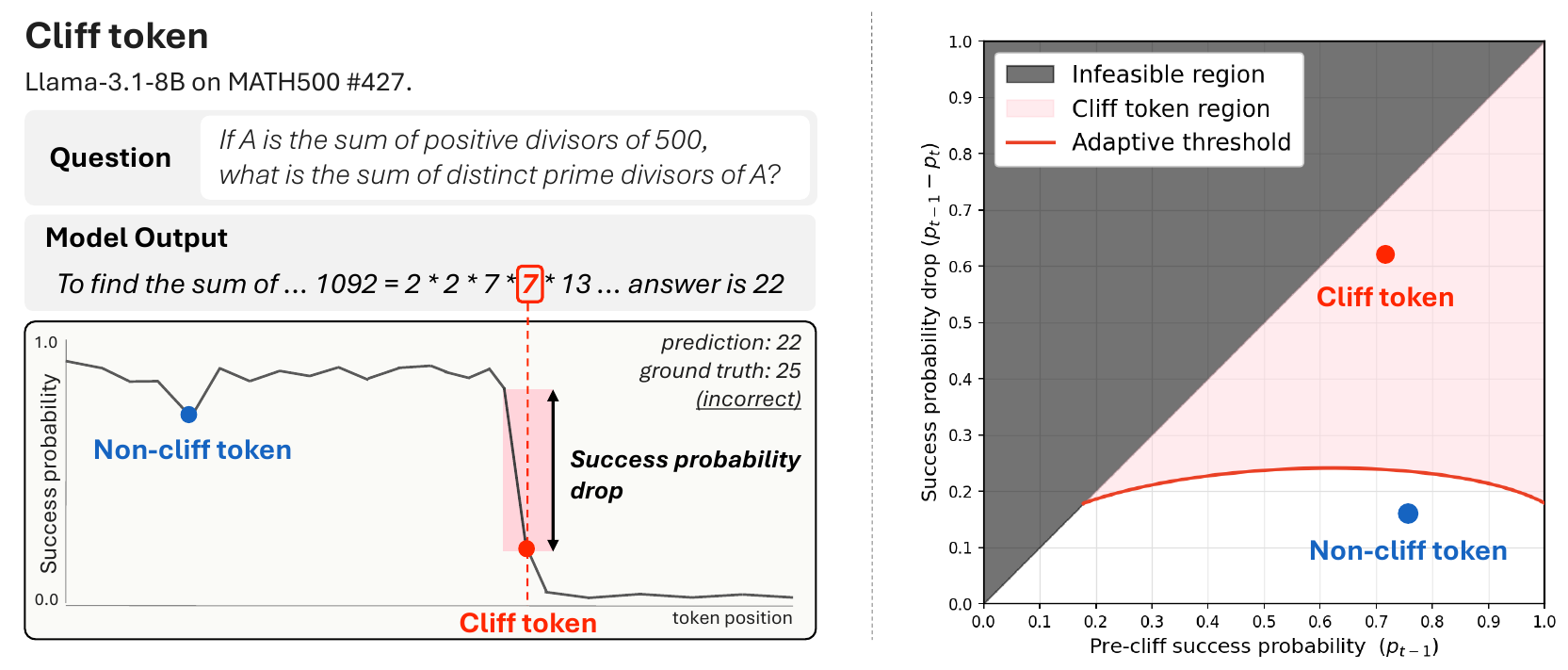}
  \caption{Cliff token identification. (Left) Example reasoning trace where the token `7' produces a drop in success probability, after which the trace proceeds to an incorrect answer. (Right) Cliff-token decision based on the adaptive threshold: the red region denotes drops satisfying the cliff-token criterion. The red point is identified as a cliff token; the blue point is not.}
  \label{fig:main_figure}
\end{figure}

To address these gaps, we identify which tokens shift a reasoning trace toward failure and analyze their characteristics. We define the \textbf{cliff token}: the token in a reasoning trace where the probability of reaching the correct answer via Monte Carlo rollouts (success probability) drops significantly. For example, in \Cref{fig:main_figure} (Left), the highlighted token `7' shifts the trace toward an incorrect factorization of $1{,}092$, inserting an extra factor of $7$ where the correct factorization requires $3$. The trace is still recoverable before this token, but once `7' is sampled, most continuations lead to incorrect answers and the success probability collapses. To distinguish statistically significant shifts from sampling noise, we apply an \textit{adaptive threshold} based on a one-sided two-proportion $z$-test (\Cref{fig:main_figure}, Right).

Using this framework, we analyze cliff tokens across seven models (\texttt{Qwen3-8B}, \texttt{Qwen3-4B}, \texttt{Qwen3-0.6B} \citep{Qwen3}, \texttt{Llama-3.1-8B}, \texttt{Llama-3.2-3B}, \texttt{Llama-3.2-1B} \citep{llama3}, and \texttt{Gemma-3-4B} \citep{gemma3}) and three benchmarks (GSM1K \citep{gsm1k}, MATH500 \citep{math500}, and AIME 2025 \citep{aime2025}). We show that cliff tokens act as failure triggers: resampling immediately before the first cliff token generally yields higher pass@$k$ than after it. We find that cliff tokens do not necessarily coincide with error tokens, and after a cliff token, traces can be recovered as the model self-corrects (\Cref{sec:analysis_1_local_failure_transitions}). We identify three types of cliff tokens---\textit{deterministic}, \textit{uncertain}, and \textit{sampled-off cliffs}---based on token entropy and whether the token is the greedy choice (\Cref{sec:cliff_taxonomy}). We show that the cliff types differ as training signals using single-token preference optimization at cliff positions (Cliff-DPO). Uncertain and sampled-off cliffs show larger accuracy gains than deterministic ones, as the models already assign near-certain probabilities to deterministic cliffs (\Cref{sec:cliff_dpo}). Finally, we release token-level rollout data to support further research on LLM reasoning. Our main contributions can be summarized as follows:
\begin{itemize}[leftmargin=*]
\item We formalize the \textit{cliff token} with a $z$-test based adaptive threshold that separates statistically significant drops in success probability from sampling noise, showing that these tokens act as failure triggers.
\item We introduce a \textit{cliff taxonomy}---\textit{deterministic}, \textit{uncertain}, and \textit{sampled-off cliffs}---showing that the proportion of each type varies across model families and sizes.
\item We show that uncertain and sampled-off cliffs are more effective as training signals than deterministic ones under Cliff-DPO, as the models assign near-certain probabilities to deterministic cliffs.
\end{itemize}

\section{Related work}
\paragraph{Localizing reasoning failures.} Prior work has studied reasoning errors in LLM traces at coarser levels of granularity, such as steps, chunks, and sentences \citep{math500,MathShepherd,OmegaPRM,Self-Explore,ThePotential,ThoughtAnchors,StepWiser}. However, a single token can shift a reasoning trace toward failure. Such tokens can be identified by a drop in the estimated probability of reaching the correct answer after the token is generated. However, existing definitions may miss tokens that shift a trace toward failure. For example, \citet{CriticalTokens} identify tokens only when the success probability reaches zero, but we show that success probability can drop sharply while remaining above zero. \citet{Phi4} use a fixed threshold, which does not incorporate sampling noise. We instead identify tokens where the success probability drops sharply, even if it remains above zero, using a statistically motivated adaptive threshold.

\paragraph{Token-level preference optimization.}
Beyond failure identification, token-level signals have been used to extend DPO \citep{DPO}. \citet{TDPO}, \citet{TIS-DPO}, and \citet{TGDPO} introduce token-level DPO objectives that incorporate KL regularization, importance weights, or per-token reward guidance. \citet{CriticalTokens} estimate critical tokens using contrastive estimation and use them as token-level weights in the DPO loss. \citet{FocusedDPO} concentrate preference optimization on tokens within error-prone regions in code generation. \citet{Phi4} construct single-token DPO pairs at pivotal positions identified by recursive subdivision. We test \textit{Cliff-DPO}, targeting cliff positions for preference optimization, and find that it improves reasoning performance with effectiveness varying across cliff types (\Cref{sec:cliff_dpo}).

\section{Cliff token}
\label{sec:cliff_token}
\subsection{Cliff token: definitions and formalization}
\label{sec:cliff_token_definition_and_formalization}
A \textbf{cliff token} is a token where the success probability drops significantly when it is generated (see \Cref{fig:main_figure}, Left).
Success probability\footnote{This rollout-based quantity has been referred to as \textit{potential}~\citep{ThePotential}, \textit{correctness score}~\citep{CriticalTokens}, and \textit{success probability}~\citep{Phi4}. We adopt the last term.} is the probability of reaching the correct answer conditioned on the reasoning trace generated up to a specific token position $t$. Formally, given a prompt $\boldsymbol{x}$, a partial reasoning sequence $\boldsymbol{c}_{\le t}$, and the ground-truth answer $y^*$, the success probability is defined as follows:
\begin{equation}
{
P(\boldsymbol{c}_{\le t}; \boldsymbol{x}) := \mathbb{P}_{(\boldsymbol{c}_{> t}, y) \sim \text{LM}_\theta(\cdot | \boldsymbol{c}_{\le t}, \boldsymbol{x})} (y = y^*).
}
\end{equation}
We empirically estimate success probability, denoted as $\widehat{P}_{N}$, by executing $N$ rollouts from every token position $t$ and computing the success rate of reaching the ground-truth answer:
\begin{equation}
{
\widehat{P}_{N}(\boldsymbol{c}_{\le t}; \boldsymbol{x}) := \frac{1}{N} \sum_{n=1}^{N} \mathbbm{1}_{\{y^{(n)}=y^*\}} \quad \text{where } \left(\boldsymbol{c}_{> t}^{(n)}, y^{(n)}\right) \sim \text{LM}_\theta(\cdot|\boldsymbol{c}_{\le t}, \boldsymbol{x}).
}
\end{equation}

With a finite number of rollouts, the estimated success probabilities are subject to sampling noise \citep{ImprovingVarianceReduction,ForkingFast,ReasoningDrift}. Thus, identifying success probability fluctuations based on absolute changes is statistically unreliable. While increasing the number of rollouts could mitigate this variance, it is computationally prohibitive given the quadratic complexity of success probability estimation (see \Cref{app:computational_complexity}). To reliably identify cliff tokens under these constraints, we use an \textit{adaptive threshold} based on one-sided two-proportion $z$-test at a 95\% confidence level. Let $\widehat{p}_t := \widehat{P}_{N}(\boldsymbol{c}_{\le t}; \boldsymbol{x})$ denote the estimate of success probability at position $t$. A token at position $t$ is identified as a cliff token if the success probability drop $\Delta_t = \widehat{p}_{t-1} - \widehat{p}_t$ satisfies the following:
\begin{equation}
{
\Delta_t > \tau + 1.645 \cdot \text{SE}_t, \quad \text{where} \quad \text{SE}_t = \sqrt{\frac{\widehat{p}_{t-1}(1 - \widehat{p}_{t-1})}{N} + \frac{\widehat{p}_t(1 - \widehat{p}_t)}{N}}.
}
\end{equation}

This adaptive threshold combines the base threshold $\tau$ and a statistical margin $1.645 \cdot \text{SE}_t$. The margin dynamically accounts for the variance in the probability estimation and reduces false-positive detections in high-variance regions.

\subsection{Experiment settings}
\label{sec:experiment_settings}
\paragraph{Models.} We use seven instruction-tuned models across different scales and model families: \texttt{Qwen3-8B/4B/0.6B}, \texttt{Llama-3.1-8B-Instruct}, \texttt{Llama-3.2-3B/1B-Instruct}, and \texttt{Gemma-3-4B-it}. For generating stem traces, we use the temperature\footnote{\texttt{Qwen3} ($T=0.7$, non-thinking mode), \texttt{Llama-3} ($T=0.6$), and \texttt{Gemma-3} ($T=1.0$).} $T$, top-$p$, and top-$k$ settings recommended by each model provider. See \Cref{app:experimental_details} for inference hyperparameters. 

\paragraph{Datasets.} We use three mathematical reasoning datasets: GSM1K, MATH500, and AIME 2025. Following \citet{CriticalTokens, ThoughtAnchors, ForkingPaths}, we use subsampling to reduce the computational cost of token-wise rollouts. We randomly sample 100 problems each from GSM1K and MATH500 and use the full set of $30$ problems from AIME 2025. Even with this subsampling, success probability estimation required $4{,}047$ A100 (80GB) GPU-hours for processing $1{,}610$ model--problem pairs ($7$ models, $230$ problems). See \Cref{app:empirical_cost} for the empirical GPU-hour breakdown. We publicly release the token-level rollout data used to estimate success probabilities so that other researchers can analyze cliff tokens without regenerating these computationally expensive rollouts.

\paragraph{Threshold.} We set the base threshold $\tau = 0.1$, following \citet{ThePotential}, who define monotonicity in reasoning as a maximum success probability drop of $0.1$ between consecutive steps. We also considered setting $\tau = 0.2$ or higher (which shifts the adaptive range to $[0.294, 0.337]$; see \Cref{app:adaptive_threshold}), but preliminary analyses showed this was too strict and missed many clear failure-triggering tokens.

\subsection{Cliff token identification}
\label{sec:cliff_token_identification}

\paragraph{Generating stem traces.} For each problem $\boldsymbol{x}$ and model $\mathrm{LM}_\theta$, we use a common zero-shot prompt (\Cref{app:prompt_template}) to sample exactly one complete response, which we call the stem trace. We neither select the best of multiple responses nor edit the stem trace, and we keep it regardless of whether its final answer is correct.

\paragraph{Rollout protocol.} To estimate the success probability, we perform $N=64$ rollouts at each evaluated position. To further reduce computational cost, we apply an early-termination heuristic inspired by similar strategies in reinforcement learning with verifiable rewards \citep{ARRoL, LATR}: if the success probability remains $0$ for $20$ consecutive tokens, we consider the reasoning trace irrecoverable and truncate all subsequent rollouts for that sequence. We also set dataset-specific maximum token lengths depending on their difficulty (see \Cref{app:experimental_details}). Using the adaptive threshold defined in \Cref{sec:cliff_token_definition_and_formalization}, we identify cliff tokens from drops in success probability between adjacent token prefixes.

\section{Do cliff tokens trigger reasoning failures?}
\label{sec:analysis_1_local_failure_transitions}

\subsection{Cliff tokens occur more often in incorrect traces}
\label{sec:cliff_frequency}
\begin{figure}[t]
  \centering
  \includegraphics[width=0.8\linewidth]{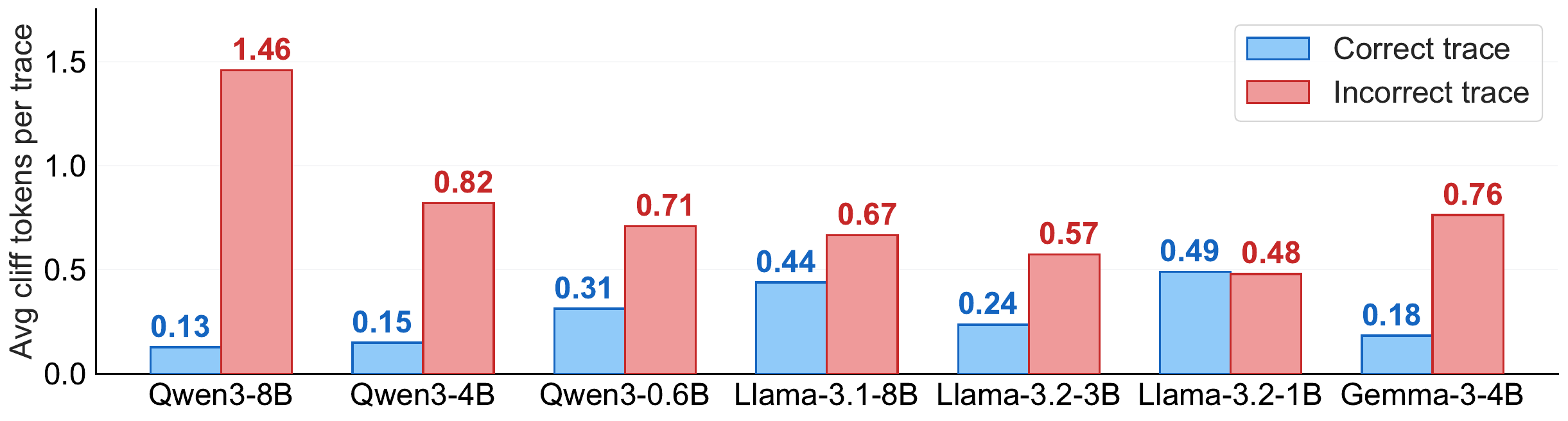}
  \caption{
  Average  cliff tokens per trace, in correct vs. incorrect cases. Aggregated across GSM1K, MATH500, and AIME 2025.}
  \label{fig:cliff_rate}
\end{figure}

\Cref{fig:cliff_rate} shows that cliff tokens occur more frequently in incorrect traces than in correct traces for most models, with \texttt{Llama-3.2-1B} as the only exception. Among incorrect traces, \texttt{Qwen3-8B} generates the most cliff tokens per trace on average, nearly twice as many as the next-highest model. The lack of separation between correct and incorrect traces for \texttt{Llama-3.2-1B} reflects two causes. First, 52.5\% of its incorrect traces never reach the minimum adaptive threshold (0.180), so cliff tokens cannot exist in incorrect traces by definition. Second, \texttt{Llama-3.2-1B} fails to retain successful prefixes due to low reasoning ability, as observed in highest cliff token rates in correct traces (0.49/trace).

\subsection{Removing a single cliff token recovers reasoning traces}
\label{sec:pass_at_k}
We investigate the effect of cliff tokens on reasoning traces by observing how model decoding behaves when these tokens are kept versus removed. Specifically, we focus on incorrect traces that contain at least one cliff token. To ensure a fair comparison and prevent traces with multiple cliff tokens from being overrepresented in the results, we restrict our evaluation to the \textit{first} occurring cliff token per trace. Let $c_{t^*}$ denote this first cliff token. Using newly sampled continuations independent of the detection rollouts, we evaluate pass@$k$ under two prefix conditions:

\begin{itemize}[leftmargin=*, nosep]
    \item \textbf{Cliff-del:} Resampling from the prefix $\boldsymbol{x} \oplus \boldsymbol{c}_{< t^*}$, excluding the cliff token $c_{t^*}$. This setup allows us to test whether the model can diverge from the failure trace when the cliff token is deleted.
    \item \textbf{Cliff-keep:} Resampling from the prefix $\boldsymbol{x} \oplus \boldsymbol{c}_{\le t^*}$, including the cliff token $c_{t^*}$. This setup forces the model to continue decoding directly from the point of the success probability drop.
\end{itemize}

\begin{figure}[t]
  \centering
  \includegraphics[width=1.0\linewidth]{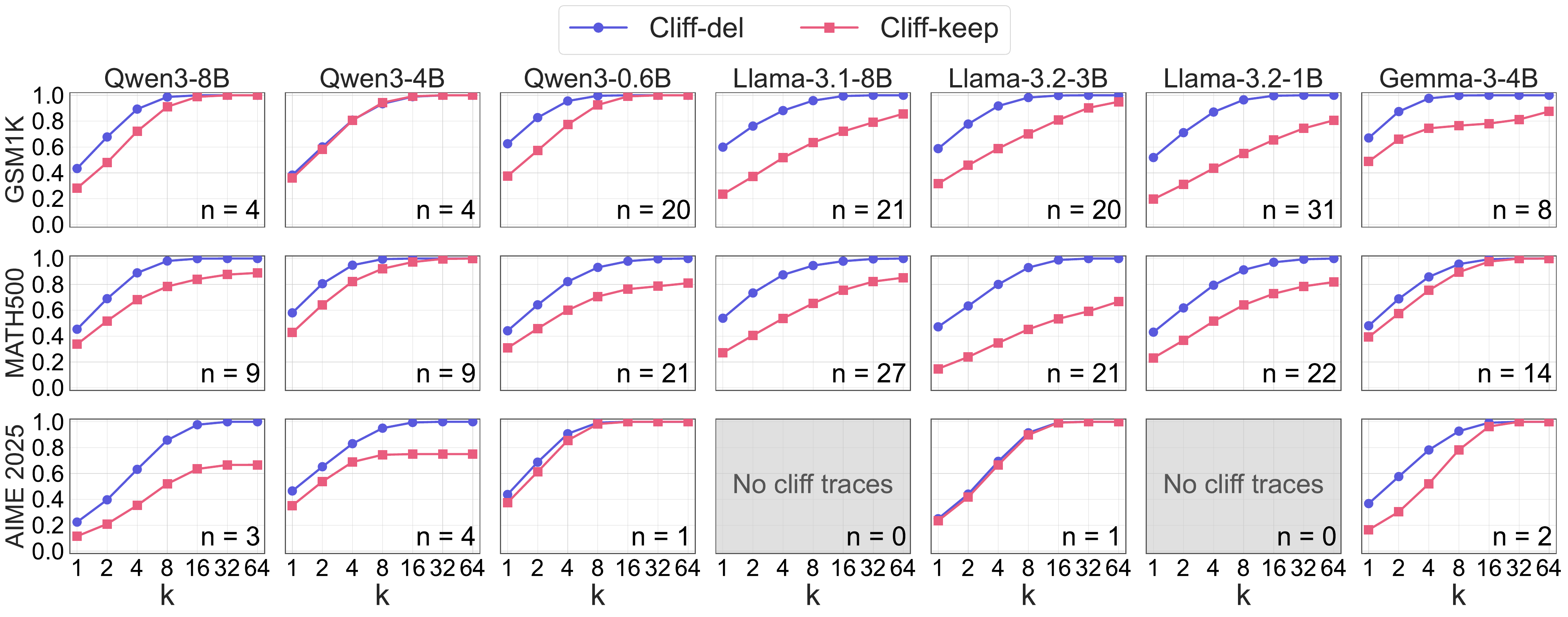}
  \caption{Cliff-del and Cliff-keep pass@$k$ results on incorrect traces. The gap between Cliff-del and Cliff-keep across pass@$k$ shows that removing a cliff token can restore reasoning performance. $n$ denotes the number of incorrect traces containing at least one cliff token. Gray panels mean no cliff tokens in that setting. See \Cref{app:cliff_del_cliff_keep} for detailed results on both incorrect and correct traces.}
  \label{fig:pass_k_comparison}
\end{figure}

\Cref{fig:pass_k_comparison} shows the pass@$k$ results for both the Cliff-del and Cliff-keep setups. Cliff-del shows higher point estimates than Cliff-keep at every $k$ until both reach $1{.}0$, except for \texttt{Qwen3-4B} on GSM1K at $k=8,16,32$, where Cliff-keep is only marginally higher. This gap is particularly notable at small $k$ (e.g., for \texttt{Llama-3.1-8B}, pass@1 is $0.60$ for Cliff-del vs. $0.23$ for Cliff-keep on GSM1K), showing that removing the cliff token leads to a substantial recovery in reasoning performance without requiring extensive resampling. Furthermore, the Cliff-del pass@64 rate reaches 1.0 across all evaluated panels, indicating that these traces are solvable when the cliff token is removed. Conversely, the Cliff-keep pass@64 rates remain between 0.67 and 1.00 across panels with cliff tokens, indicating that in some panels, even 64 resamples are insufficient to recover the trace once the cliff token is fixed. These results suggest that cliff tokens act as triggers of reasoning failures.

\subsection{Cliff tokens are local failure triggers in recoverable dynamics}
\label{sec:cliff_recoverable_dynamics}

\begin{figure}[t]
  \centering
  \includegraphics[width=\textwidth]{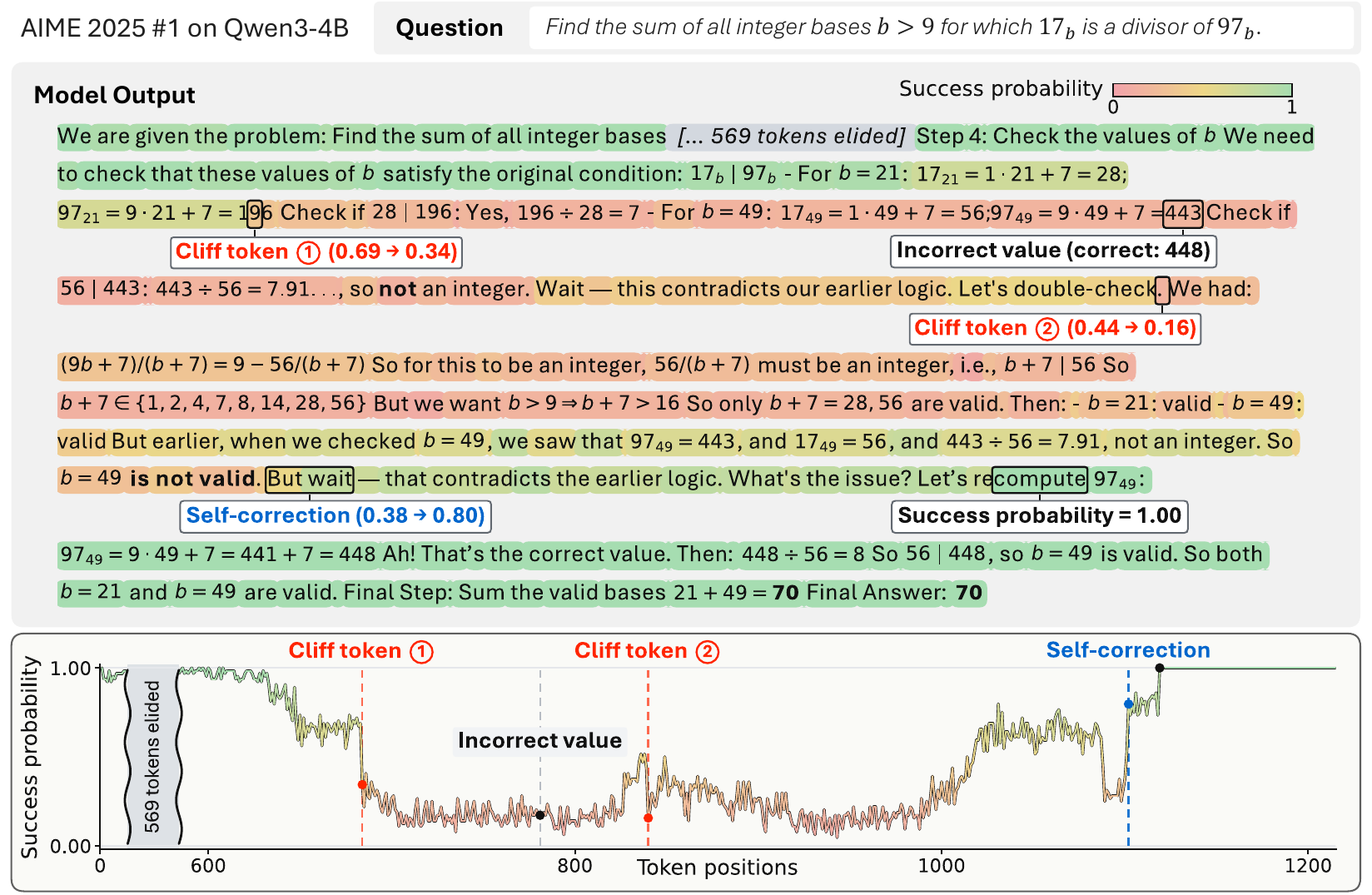}
  \caption{An example reasoning trace that reaches the correct answer despite containing two cliff tokens (\texttt{Qwen3-4B}, AIME 2025 Problem~1). Both cliff tokens produce a drop in success probability. The semantic error point (calculation error; 443) is different from the cliff tokens. Success probability recovers after the model writes ``But wait'' and corrects its reasoning.}
  \label{fig:self_correction}
\end{figure}

In contrast to critical tokens, which identify positions where success probability has already collapsed to zero \citep{CriticalTokens}, cliff tokens capture local shifts toward failure that are not necessarily irreversible. In \Cref{fig:self_correction}, a correct \texttt{Qwen3-4B} trace for AIME 2025 Problem~1 contains two cliff tokens. Success probability drops at the first cliff token ($0.69$ to $0.34$) and the second ($0.44$ to $0.16$), then recovers to $0.80$ once the model writes ``But wait'' and starts to correct its arithmetic. More broadly, among the 358 traces containing at least one cliff token, 47.2\% contain multiple cliff tokens and 32.4\% still reach the correct answer (see \Cref{app:cliff_recovery_stats}). These results reflect non-monotonic reasoning dynamics: success probability can increase after a cliff token as the model verifies, backtracks, or self-corrects. Furthermore, in \Cref{fig:self_correction}, neither cliff token coincides with the actual arithmetic error itself (the miscomputed value $443$). This aligns with findings that critical tokens do not necessarily coincide with human-annotated error tokens \citep{CriticalTokens}.

\section{Cliff taxonomy}
\label{sec:cliff_taxonomy}

\subsection{Definition of cliff taxonomy}
\label{def_cliff_taxonomy}


Are all cliff tokens equal? To analyze how the token sampling process affects success probability, we characterize cliff tokens along two criteria: whether the sampled token is the greedy token, and the entropy $H$ of the next-token distribution at that position. First, one might suspect that cliff tokens happen because the sampled token was not the top choice; we find that many of the cliff tokens are still greedy (38\%--82\%); see \Cref{app:greedy_ratio}. Second, while the entropy of ordinary reasoning tokens is concentrated near $H \approx 0$, cliff tokens show lower density in this low-entropy regime and a heavier-tailed distribution; see \Cref{app:entropy}. This indicates that many cliff tokens occur under local uncertainty. These observations motivate the following cliff taxonomy.

To formalize this taxonomy, we first define these two criteria. Let $p_t(v)=p(v \mid x, c_{<t})$ denote the next-token probability of $v \in \mathcal{V}$ at token position $t$. The sampled token $c_t$ is greedy if $c_t \in \arg\max_{v \in \mathcal{V}} p_t(v)$ and non-greedy otherwise. The token entropy at position $t$ is
\begin{equation}
    H_t = -\sum_{v \in \mathcal{V}} p_t(v) \log p_t(v).
\end{equation}

We then set an entropy threshold below which we consider a position near-deterministic. Rather than choosing an entropy value directly, we anchor the threshold to the probability of the greedy token $p_1=0.99$. Among all distributions whose greedy probability is $p_1$, the minimum token entropy is the binary entropy
\begin{equation}
    H_b(p_1) = -p_1 \log(p_1) - (1 - p_1) \log(1 - p_1),
\end{equation}
which depends only on $p_1$, making the threshold invariant to vocabulary size and top-$k$ across models. We adopt $H_b(0.99) \approx 0.0561$ nats, and an ablation over $p_1 \in \{0.90, 0.95, 0.99, 0.999\}$ confirms that our findings in \Cref{sec:cliff_taxonomy_across_seven_models} are consistent across these different thresholds (see \Cref{different_entropy_threshold}).

Using this threshold, we introduce a \textit{cliff taxonomy} with three types:
\begin{enumerate}[label=(\arabic*), leftmargin=2.0em, labelsep=0.7em, nosep]
    \item \textbf{Deterministic cliff}: a greedy token with $H_t < 0.0561$; the model samples the cliff token with near certainty.
    \item \textbf{Uncertain cliff}: a greedy token with $H_t \ge 0.0561$; the greedy cliff token is sampled despite high uncertainty.
    \item \textbf{Sampled-off cliff}: a non-greedy token with $H_t \ge 0.0561$; a non-greedy cliff token is stochastically sampled under high uncertainty.
\end{enumerate}

The fourth type (non-greedy token, $H_t < 0.0561$) is excluded from our analysis due to its extreme rarity by definition; as it requires sampling a token with probability $p \le 0.01$, only six cases were observed (three within \texttt{Gemma-3-4B} and three within \texttt{Llama-3.2-3B}).

\subsection{Cliff taxonomy varies by model family and scale}
\label{sec:cliff_taxonomy_across_seven_models}

\begin{table}[t]
  \caption{Distribution of cliff types: deterministic (D), uncertain (U), and sampled-off (S). Cliff and all-token proportions give each type’s percentage among identified cliff tokens and all reasoning-trace tokens, respectively. Darker shading indicates larger values.}
  \label{tab:cliff_taxonomy_models}
  \centering
  \small
  \setlength{\tabcolsep}{3pt}
  \begin{tabular}{@{}>{\raggedright\arraybackslash}p{0.23\linewidth}*{6}{>{\centering\arraybackslash}p{\dimexpr(0.77\linewidth-12\tabcolsep)/6\relax}}@{}}
    \toprule
    & \multicolumn{3}{c}{\textbf{Cliff proportion (\%)}} 
    & \multicolumn{3}{c}{\textbf{All-token proportion (\%)}} \\
    \cmidrule(lr){2-4}
    \cmidrule(lr){5-7}
    \textbf{Model} & D & U & S & D & U & S \\

    \specialrule{\heavyrulewidth}{0.5ex}{0.5ex}
    \texttt{Qwen3-8B}
      & \cellcolor{black!19.16}47.9
      & \cellcolor{black!13.76}34.4
      & \cellcolor{black!7.08}17.7
      & 69.1 & 27.8 & 3.1 \\
    \texttt{Qwen3-4B}
      & \cellcolor{black!17.76}44.4
      & \cellcolor{black!14.44}36.1
      & \cellcolor{black!7.76}19.4
      & 73.9 & 23.6 & 2.5 \\
    \texttt{Qwen3-0.6B}
      & \cellcolor{black!11.92}29.8
      & \cellcolor{black!16.76}41.9
      & \cellcolor{black!11.28}28.2
      & 64.3 & 32.2 & 3.5 \\

    \specialrule{\lightrulewidth}{0.5ex}{0.5ex}
    \texttt{Llama-3.1-8B}
      & \cellcolor{black!8.80}22.0
      & \cellcolor{black!10.08}25.2
      & \cellcolor{black!21.12}52.8
      & 66.8 & 28.3 & 4.8 \\
    \texttt{Llama-3.2-3B}
      & \cellcolor{black!6.72}16.8
      & \cellcolor{black!10.12}25.3
      & \cellcolor{black!21.88}54.7
      & 73.5 & 22.4 & 4.0 \\
    \texttt{Llama-3.2-1B}
      & \cellcolor{black!3.60}9.0
      & \cellcolor{black!11.52}28.8
      & \cellcolor{black!24.88}62.2
      & 74.3 & 22.0 & 3.6 \\

    \specialrule{\lightrulewidth}{0.5ex}{0.5ex}
    \texttt{Gemma-3-4B}
      & \cellcolor{black!18.08}45.2
      & \cellcolor{black!11.92}29.8
      & \cellcolor{black!8.56}21.4
      & 77.6 & 19.7 & 2.7 \\

    \specialrule{\heavyrulewidth}{0.5ex}{0pt}
  \end{tabular}
\end{table}

We first compare the cliff-token distributions against the all-token distributions. As shown in \Cref{tab:cliff_taxonomy_models}, deterministic cliffs account for a smaller proportion of cliff tokens compared to their all-token proportion, whereas sampled-off cliffs occur at a higher proportion across all models. This indicates that cliff tokens are not randomly distributed errors. Instead, they show a distinct distribution pattern from normal tokens.

Furthermore, analyzing models within the same family shows a consistent scaling effect. As model size increases in the \texttt{Qwen3} and \texttt{Llama-3} families, the proportion of sampled-off cliffs decreases, while the proportion of deterministic cliffs increases. This trend suggests that scaling up model parameters shifts the errors from stochastic sampling to deterministic ones. Beyond scale, the dominant cliff type is influenced by the model's family. Similar-sized models, such as \texttt{Qwen3-8B} and \texttt{Llama-3.1-8B}, exhibit different cliff-token distributions. \texttt{Qwen3-8B} produces more deterministic cliffs, whereas \texttt{Llama-3.1-8B} produces more sampled-off cliffs. This shows that each model family has a specific pattern in its reasoning failures. Overall, these findings show that token-level mathematical reasoning failures vary with both model family and scale. This heterogeneity has direct implications for cliff-targeted post-training, which we explore in \Cref{sec:cliff_dpo}.

\section{Do cliff types differ as training signals?}
\label{sec:cliff_dpo}
While prior work has demonstrated the effectiveness of post-training with token-level supervision \citep{CriticalTokens,Phi4}, these approaches treat token-level signals as a homogeneous supervision signal. This raises the question: Do the three cliff types differ in their effectiveness as training signals?

\subsection{Methodology}
\label{sec:cliff_dpo_methodology}
\paragraph{Cliff-DPO loss.}
Given the cliff preference dataset $\mathcal{D}_{\text{cliff}} = \{(\boldsymbol{x}_i, \boldsymbol{c}_{i,<t_i}, c_{i,t_i}^w, c_{i,t_i}^l)\}_{i=1}^M$, we adapt the standard sigmoid Direct Preference Optimization (DPO) loss \citep{DPO} to provide single-token supervision specifically at each identified cliff position $t_i$. We refer to this objective as \emph{Cliff-DPO}. The loss is formulated as follows:
\begin{equation}
\mathcal{L}_{\text{Cliff-DPO}}(\theta) = - \frac{1}{M} \sum_{i=1}^M \log \sigma \left( r_\theta(c_{i,t_i}^w ; \boldsymbol{x}_i, \boldsymbol{c}_{i,<t_i}) - r_\theta(c_{i,t_i}^l ; \boldsymbol{x}_i, \boldsymbol{c}_{i,<t_i}) \right),
\end{equation}
where $r_\theta$ denotes the implicit pointwise reward, defined as
\begin{equation}
r_\theta(c_t ; \boldsymbol{x}, \boldsymbol{c}_{<t}) := \beta \log \frac{\pi_\theta(c_t \mid \boldsymbol{x}, \boldsymbol{c}_{<t})}{\pi_{\text{ref}}(c_t \mid \boldsymbol{x}, \boldsymbol{c}_{<t})}.
\end{equation}

Here, $c_{i,t_i}^w$ and $c_{i,t_i}^l$ are the non-cliff token (chosen) and cliff token (rejected) at position $t_i$, respectively. The coefficient $\beta$ controls the scale of the implicit reward and the strength of reference-policy regularization. By applying the preference loss only to the candidate tokens at the cliff position, Cliff-DPO localizes the training signal to the point where the reasoning trace diverges into failure.

\paragraph{Constructing Cliff-DPO training pairs.} To construct $\mathcal{D}_{\text{cliff}}$, we apply the pipeline in \Cref{sec:cliff_token_identification} to the GSM8K \citep{gsm8k} training set of 7,473 problems and identify cliff positions. At each cliff position $t$, we consider the top-10 candidate tokens under the model distribution and perform $N=64$ rollouts for each candidate to estimate its success probability. We define $C_t^{\text{non-cliff}}$ as the set of candidates whose estimated success probability does not indicate a cliff token. Each non-cliff token $c_t^w \in C_t^{\text{non-cliff}}$ is then paired with the originally detected cliff token $c_t^l$, yielding a chosen--rejected preference pair. These preference pairs are then divided into three subsets according to the type of the rejected cliff token: \textit{deterministic}, \textit{uncertain}, or \textit{sampled-off}.

\begin{figure}[t]
\centering
\includegraphics[width=\textwidth]{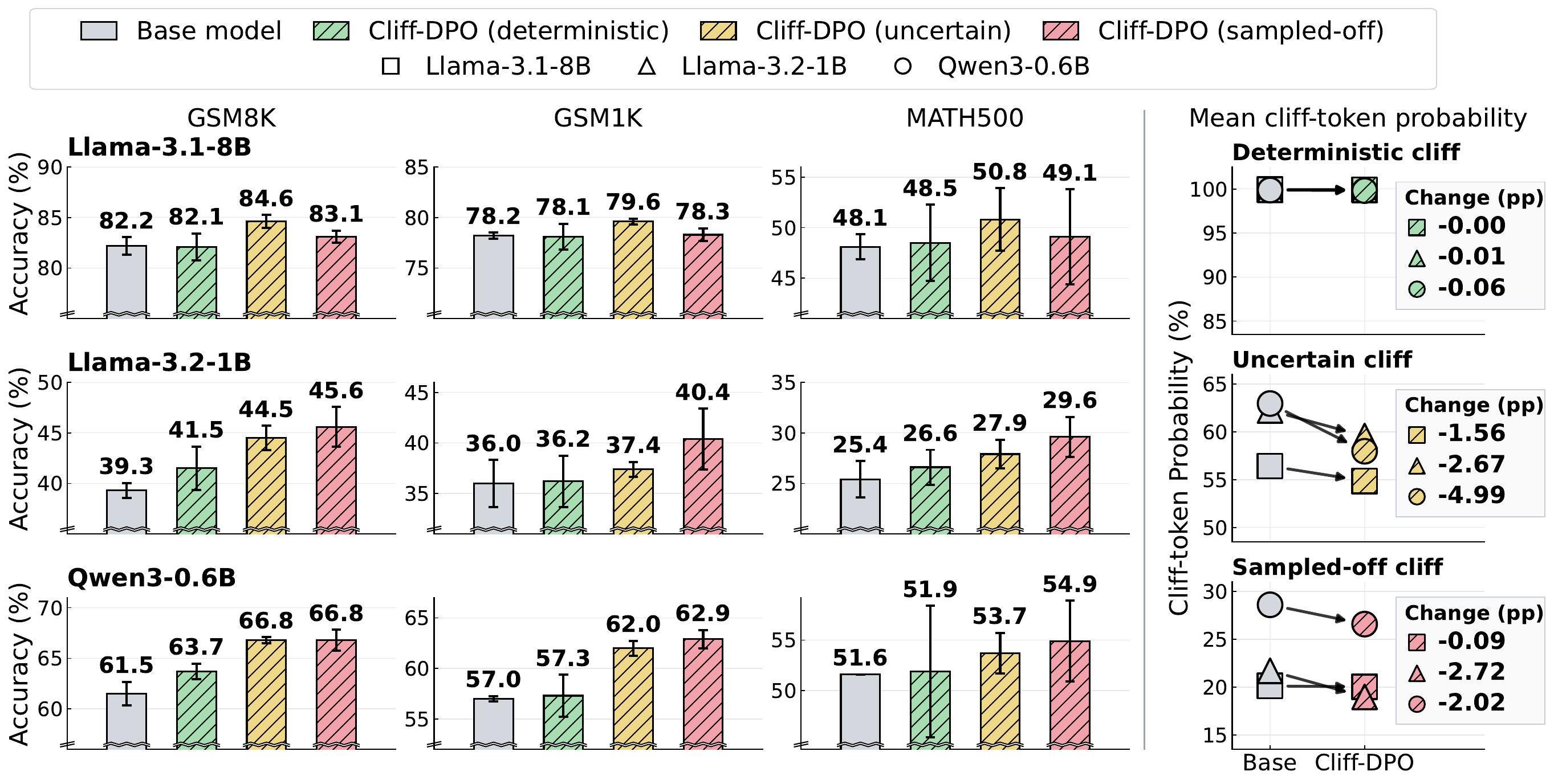}
\caption{\textbf{(Left)} Cliff-DPO performance across models and benchmarks. Mean accuracy and 95\% confidence intervals are reported. \textbf{(Right)} Mean cliff-token probability before and after Cliff-DPO on held-out cliff positions from GSM1K, MATH500, and AIME 2025. Each variant is evaluated on its corresponding cliff type; changes are in percentage points (pp). Full results are in \Cref{app:dpo_prob_shift}.}
\label{fig:cliff_dpo_main_results}
\end{figure}

\paragraph{Experimental setup.} As identifying cliff tokens across 7,473 problems requires extensive rollouts at scale, we use three models of different sizes and families: \texttt{Llama-3.1-8B}, \texttt{Llama-3.2-1B}, and \texttt{Qwen3-0.6B}. We train them across three distinct data configurations to analyze the impact of different cliff types: \textit{deterministic}, \textit{uncertain}, and \textit{sampled-off}.\footnote{To control for the differing pair counts across the cliff types, we also train size-matched variants. See \Cref{app:dpo_controlled} for further details and results.} See \Cref{app:appendix_training} for training and evaluation details and \Cref{app:dpo_full} for results on additional Cliff-DPO variants and comparisons with three baselines: the base model, DPO \citep{DPO}, and cDPO  \citep{CriticalTokens}.

\subsection{Results and discussion}
\label{sec:cliff_dpo_results}

\Cref{fig:cliff_dpo_main_results} shows that both the \textit{uncertain} variant and the \textit{sampled-off} variant achieve higher mean accuracy than the base model and the \textit{deterministic} variant on GSM8K, GSM1K, and MATH500 across all three models. The best-performing variant differs by model: on average across these benchmarks, the \textit{sampled-off} variant ranks highest for \texttt{Qwen3-0.6B} and \texttt{Llama-3.2-1B}, whereas the \textit{uncertain} variant ranks highest for \texttt{Llama-3.1-8B}. The \textit{deterministic} variant shows smaller gains for the two smaller models and leaves \texttt{Llama-3.1-8B} performance largely unchanged on these three benchmarks. Overall, the \textit{uncertain} variant and the \textit{sampled-off} variant are more effective in the evaluated settings on the three benchmarks.

The effect of Cliff-DPO depends on how strongly the model initially favors the cliff token. For deterministic cliffs, the token probability is more than 99\%, indicating that the prior is difficult to overcome via post-training. Accordingly, Cliff-DPO changes its mean cliff-token probability by at most $0.06$ pp. For uncertain and sampled-off cliffs, the model assigns lower probability to the cliff token, giving post-training more room to reduce its token probability. Cliff-DPO therefore produces larger reductions of token probability for uncertain cliffs (up to $4.99$ pp) and sampled-off cliffs (up to $2.72$ pp).



\section{Conclusion}
\label{sec:conclusion}

We introduce \textbf{cliff tokens}, single tokens where the success probability of a reasoning trace collapses. We formalize cliff tokens using a $z$-test-based adaptive threshold, separating statistically significant drops from sampling noise. Across evaluated model--benchmark settings where cliff tokens are observed, deleting the first cliff token generally outperforms retaining it, showing that cliff tokens act as failure triggers. Starting with a clear gap at $k=1$, this advantage persists across larger $k$, reaching a pass@64 of 1.0, whereas retaining the token yields 0.86 on average. We also find two distinct properties of cliff tokens: they do not necessarily coincide with explicit error tokens, and they do not guarantee failure, since the model can recover through self-correction. We further introduce a cliff taxonomy---deterministic, uncertain, and sampled-off cliffs---and show that its distribution varies by model family and scale. Finally, using Cliff-DPO, we show that single-token supervision at cliff positions improves mean accuracy, with the \textit{uncertain} and \textit{sampled-off} variants showing larger gains than the \textit{deterministic} variant across three models on GSM8K, GSM1K, and MATH500. Overall, cliff tokens offer a token-level view for diagnosing reasoning failures and a targeted training signal for improving mathematical reasoning in LLMs.

\paragraph{Limitations and future work.} Our framework characterizes failure triggers, but its scope is limited to cases where the model can reach the correct answer but loses that chance during reasoning. When a problem is beyond the model’s capacity, success probability is near zero from the start, so no cliff token appears. In $30$ incorrect \texttt{Llama-3.2-1B} traces on AIME 2025, we observed no cliff tokens because the success probability remained below $0{.}05$ throughout all $30$ traces. Our experiments are constrained by the computational cost of success probability estimation, which required $4{,}047$ A100 (80GB) GPU-hours across seven models and three benchmarks (\Cref{app:comp_cost_success_probability}). Due to the substantial compute requirements, we subsampled GSM1K and MATH500 rather than evaluating their full sets. Although Cliff-DPO uses few loss-contributing tokens (\Cref{app:dpo_full}), obtaining cliff tokens still requires costly rollout-based detection. Future work should develop efficient predictors, which could enable decoding-time resampling, backtracking, or branching before a reasoning trace collapses. Finally, our analysis is limited to mathematical reasoning. Whether these findings generalize to larger models, longer reasoning traces, and non-mathematical domains remains an open question.

\subsection*{AI use statement}

In this work, we used generative AI tools to assist in implementing the cliff detection, token-level rollout, and Cliff-DPO pipelines, checking sampling and decoding settings before running experiments, and writing scripts that compute the statistics reported in our tables from the raw rollout and evaluation outputs. We did not use them to form or refine hypotheses, to define cliff tokens or the cliff taxonomy, or to interpret results. Synthetic data generation and mathematical proofs are not applicable to this work. We also used AI tools to find and summarize related work, to write plotting code for figures we designed, to format tables and other content in LaTeX, and to polish the writing. We have reviewed all AI-assisted work. The authors ran, inspected, and tested all AI-assisted code, and checked every reported number and plotted value against the original outputs. We read every cited paper ourselves and checked AI-generated summaries against the originals. We also checked all AI-edited text for accuracy and consistency with our experiments. We take responsibility for the final content of this work, including text, claims or artifacts produced with the aid of generative AI.

\subsection*{Ethics statement}

This work does not involve experiments with human participants or any private data. All experiments use publicly available math benchmarks (GSM1K, GSM8K, MATH500, and AIME 2025) and open-weight models, following their licenses and terms of use. We release the model-generated reasoning traces and token-level rollout data used in our analysis. The released data contain only benchmark problems and model outputs, and remain subject to the original dataset licenses and model terms of use. This work analyzes where language models fail at mathematical reasoning, and we do not expect it to enable harmful applications. Although Cliff-DPO improves accuracy on math benchmarks, these gains may not generalize to other domains, so applying our methods in high-stakes settings would require further validation and human oversight.

\subsection*{Reproducibility statement}

We describe the cliff-token definition and the adaptive threshold in \Cref{sec:cliff_token_definition_and_formalization}, and the models, datasets, base threshold, and rollout protocol in \Cref{sec:experiment_settings,sec:cliff_token_identification}. \Cref{app:experimental_details} gives the dataset subsampling procedure and random seed, the sampling hyperparameters for each model, the maximum generation lengths, and the prompt template. The Cliff-del and Cliff-keep setups are defined in \Cref{sec:pass_at_k}, and the bootstrap procedure for their confidence intervals is described in \Cref{app:cliff_del_cliff_keep}. The entropy threshold for the cliff taxonomy is defined in \Cref{def_cliff_taxonomy}, with sensitivity analyses of the base thresholds and entropy thresholds in \Cref{app:adaptive_threshold,different_entropy_threshold}. For Cliff-DPO, \Cref{sec:cliff_dpo_methodology} describes the loss and the construction of preference pairs, and \Cref{app:appendix_training} lists the training and evaluation configurations for Cliff-DPO and both baselines, including hyperparameters, LoRA settings, and seeds. \Cref{app:dpo_controlled} describes the token-matched ablation, \Cref{app:comp_cost_success_probability} reports the computational cost, and \Cref{app:assets_licenses} lists the licenses of all models and datasets. Our anonymous repository (\url{https://github.com/beaver-22/Cliff-token}) contains the full source code and scripts for each experiment. It also includes the data used in this work: the stem traces and token-level success probabilities from 64 rollouts for all seven models, together with the GSM8K training traces used to build the Cliff-DPO pairs, so that our analyses can be reproduced without regenerating the costly rollouts.

\subsubsection*{Acknowledgments}
We thank Jaehee Kim for discussions that helped shape the initial idea of \textit{cliff tokens}, and Hankyeol Kim for suggesting the term. We also thank Takyoung Kim for insightful discussions on LLM reasoning, and Jinwoo Jang for helpful feedback during the early stages of the project. We are grateful to the DSBA NLP Group at Seoul National University for valuable discussions and support. The part of the computational work reported in this paper was performed on the Shared Computing Cluster administered by Boston University's Research Computing Services.

{
\small
\bibliography{references}
\bibliographystyle{iclr2027_conference}
}

\newpage
\appendix

\section{Sensitivity analysis of the adaptive threshold}
\label[appendix]{app:adaptive_threshold}

Because $\mathrm{SE}_t$ in the adaptive threshold depends on the estimated success probabilities before and after the token, the resulting adaptive threshold varies across possible pairs $(p_{t-1}, p_t)$. We conduct a sensitivity analysis by varying the base threshold from $0.1$ to $0.4$. 

In \Cref{tab:cliff-threshold-count}, the \textit{Adaptive threshold (min/max)} columns report the lower and upper bounds of this adaptive threshold for each base threshold. Detection counts are aggregated over GSM1K, MATH500, and AIME 2025 for each model. This analysis evaluates how changing the base threshold affects the number of detected cliff tokens across models and datasets.

\begin{table}[H]
  \caption{Sensitivity of cliff-token counts to the base threshold. The \textit{Total} column sums counts across the seven evaluated models.}
  \label{tab:cliff-threshold-count}
  \centering
  \footnotesize
  \setlength{\tabcolsep}{3pt}
  \resizebox{\linewidth}{!}{%
  \begin{tabular}{c c c c c c c c c c c}
    \toprule
     & \multicolumn{2}{c}{\textbf{Adaptive threshold}}
     & \multicolumn{8}{c}{\textbf{\# Cliff tokens detected}} \\
    \cmidrule(r){2-3} \cmidrule(l){4-11}
    \shortstack{\textbf{Baseline}\\\textbf{threshold}}
    & Min
    & Max
    & \shortstack{\texttt{Qwen3}\\\texttt{-8B}}
    & \shortstack{\texttt{Qwen3}\\\texttt{-4B}}
    & \shortstack{\texttt{Qwen3}\\\texttt{-0.6B}}
    & \shortstack{\texttt{Llama}\\\texttt{-3.1-8B}}
    & \shortstack{\texttt{Llama}\\\texttt{-3.2-3B}}
    & \shortstack{\texttt{Llama}\\\texttt{-3.2-1B}}
    & \shortstack{\texttt{Gemma}\\\texttt{-3-4B}}
    & \textbf{Total} \\
    \midrule
    0.1 & 0.180 & 0.241 & 96 & 72 & 124 & 127 & 95 & 111 & 84 & 709 \\
    0.2 & 0.294 & 0.337 & 29 & 14 & \phantom{0}24 & \phantom{0}43 & 40 & \phantom{0}41 & 11 & 202 \\
    0.3 & 0.401 & 0.431 & 17 & \phantom{0}6 & \phantom{0}15 & \phantom{0}30 & 21 & \phantom{0}24 & \phantom{0}7 & 120 \\
    0.4 & 0.503 & 0.523 & 11 & \phantom{0}2 & \phantom{0}10 & \phantom{0}22 & 16 & \phantom{0}15 & \phantom{0}4 & \phantom{0}80 \\
    \bottomrule
  \end{tabular}}
\end{table}

\section{Granularity of cliff-token detection}
\label[appendix]{app:granularity}

To obtain cliff tokens, we performed rollouts from \textit{every} token position. While this approach precisely localizes the cliff token, it is computationally expensive. In this section, we investigate whether rollouts can instead be performed every $c$-th token ($c>1$).

We evaluate fixed-size non-overlapping chunk detection on the same seven models and three benchmarks analyzed throughout this work, comprising $1{,}610$ traces and $709$ cliff tokens, using chunk sizes of $4$, $8$, $16$, and $32$ tokens. A chunk is flagged when the drop in success probability between its boundaries exceeds the adaptive threshold (\Cref{sec:cliff_token_definition_and_formalization}), and a flagged chunk counts as a hit when it contains a cliff token. By definition, a chunk size of $1$ corresponds to the original cliff-token detector, yielding perfect precision, recall, and F1 scores of $100\%$.

As reported in \Cref{tab:chunk_detection_granularity}, chunk-level detection misses the majority of cliff tokens, yielding persistently low recall across all chunk sizes. Reducing the chunk size does not resolve this issue. This performance limitation stems from the non-monotonicity of token-wise success probabilities \citep{ThePotential}. Because chunk-level evaluation only compares the boundary values of a chunk, it cannot capture internal token-level variations of success probability. First, a sharp drop can be masked by a subsequent recovery in success probability inside the same chunk. Second, the accumulation of minor drops across a chunk can create a large boundary difference, leading to false positive detections. Consequently, chunk-level detection cannot provide the precise localization required for our analysis.

\begin{table}[H]
  \centering
  \small
  \caption{Chunk-level detection of cliff tokens.}
  \label{tab:chunk_detection_granularity}
  \begin{tabular}{cccc}
    \toprule
    \textbf{Chunk size} & \textbf{Precision} & \textbf{Recall} & \textbf{F1} \\
    \midrule
    4  & 60.2\% & 40.7\% & 48.6\% \\
    8  & 60.8\% & 37.9\% & 46.7\% \\
    16 & 62.1\% & 35.9\% & 45.5\% \\
    32 & 62.3\% & 34.5\% & 44.4\% \\
    \bottomrule
  \end{tabular}
\end{table}

\section{\texorpdfstring{Computational cost of success probability estimation}{Computational cost of success probability estimation}}
\label[appendix]{app:comp_cost_success_probability}
\subsection{Computational complexity}
\label[appendix]{app:computational_complexity}

The computational load required to obtain the success probability of a given reasoning trace scales significantly with both the number of rollouts $N$ and the granularity of the estimation steps. Our success probability analysis demands estimation at every single token position on the reasoning trace to accurately capture sudden reasoning drifts (i.e., cliff tokens).

Suppose a generated reasoning trace consists of a total of $T$ tokens. At each token position $t \in \{1, \dots, T\}$, estimating the success probability requires performing $N$ rollouts, generating the remaining sequence to reach the final outcome. Assuming the average total length of a complete reasoning trace remains approximately $T$, the number of tokens generated for the rollouts at position $t$ is $N(T - t)$.

Therefore, the total number of tokens $T_{\text{tot}}$ produced to estimate the full success probability curve from scratch for a single reasoning trace is given by:
\begin{equation}
    T_{\text{tot}} = \sum_{t=1}^{T} N (T - t) = N \frac{T(T-1)}{2} \approx \frac{N T^2}{2}
\end{equation}

This demonstrates a quadratic scaling $\mathcal{O}(N T^2)$ with respect to the reasoning length $T$. In the context of mathematical reasoning, where $T$ is inherently large due to the multi-step nature of the CoT, this quadratic complexity imposes a severe computational bottleneck. This fundamental constraint justifies our methodological choice in \Cref{sec:cliff_token}: rather than attempting an intractable scale-up of $N$, we introduce a $z$-test to establish the statistical significance of Cliff tokens under a feasible computational budget.

\subsection{Empirical GPU cost}
\label[appendix]{app:empirical_cost}

In total, success probability estimation required approximately $4{,}047$ A100 (80GB) GPU-hours across seven models and three benchmarks. This cost is dominated by rollout generation and varies substantially across benchmarks because the number of rollout tokens grows with the length of the original reasoning trace. As discussed in \Cref{app:computational_complexity}, estimating success probability scales quadratically with the reasoning length, so benchmarks with longer reasoning traces incur much larger computational cost.

All rollout jobs were run with data parallelism over 8 A100 (80GB) GPUs. Under this setup, success probability estimation consumed approximately $392$ GPU-hours for GSM1K (700 traces), $2{,}087$ GPU-hours for MATH500 (700 traces), and $1{,}568$ GPU-hours for AIME 2025 (210 traces). The large difference across benchmarks reflects both the number of traces and the length of the generated reasoning traces. In particular, although AIME 2025 has fewer traces, its per-trace cost is substantially higher than GSM1K because AIME 2025 traces are much longer and require many more rollout tokens for success probability estimation.

\newpage

\section{Experimental details}
\label[appendix]{app:experimental_details}

\subsection{Dataset preprocessing}
For analyzing GSM1K and MATH500, we randomly sampled 100 problems each using a fixed random seed of 42, while including all 30 problems from AIME 2025.

\subsection{Sampling hyperparameters}
\label[appendix]{app:sampling_hyperparameters}
For both generating the initial reasoning traces and performing the subsequent rollouts for success probability estimation, we utilize the hyperparameters detailed in \Cref{tab:sampling-config}. These settings follow the official default configurations recommended for each respective model to ensure optimal reasoning performance.

\begin{table}[H]
  \caption{Sampling hyperparameters used for all experiments. ``--'' in the top-$k$ column denotes that top-$k$ truncation is disabled.}
  \label{tab:sampling-config}
  \centering
  \small
  \begin{tabular}{l c c c}
    \toprule
    \textbf{Model} & \textbf{Temperature} & {\boldmath\textbf{Top-$p$}} & {\boldmath\textbf{Top-$k$}} \\
    \midrule
    \texttt{Qwen3-8B}      & 0.7 & 0.80 & 20  \\
    \texttt{Qwen3-4B}      & 0.7 & 0.80 & 20  \\
    \texttt{Qwen3-0.6B}    & 0.7 & 0.80 & 20  \\
    \texttt{Llama-3.1-8B}  & 0.6 & 0.90 & -- \\
    \texttt{Llama-3.2-3B}  & 0.6 & 0.90 & -- \\
    \texttt{Llama-3.2-1B}  & 0.6 & 0.90 & -- \\
    \texttt{Gemma-3-4B}    & 1.0 & 0.95 & 64  \\
    \bottomrule
  \end{tabular}
\end{table}

\subsection{Maximum token lengths}
As mentioned in \Cref{sec:cliff_token_identification}, to maintain computational efficiency while accommodating the varying reasoning complexities of the datasets, we set dataset-specific maximum token limits. As detailed in \Cref{tab:max-tokens}, the maximum generation lengths during rollouts are strictly bounded to $1{,}024$ tokens for GSM1K/GSM8K, and $2{,}048$ tokens for MATH500. For AIME 2025, the rollout budget is specifically restricted to $4{,}096$ tokens to prevent models generating overly long traces that make rollout sampling infeasible, while allowing a reasonable budget for solving the challenging problems.

\begin{table}[H]
  \caption{Maximum generation tokens per dataset. \textit{Stem traces} denotes the budget for generating the initial stem trace per problem. \textit{Rollouts} indicates the budget during success probability estimation. \textit{Evaluation} represents the extended budget used when evaluating the final accuracy of each Cliff-DPO model in \Cref{sec:cliff_dpo}.}
  \label{tab:max-tokens}
  \centering
  \small
  \begin{tabular}{l r r r}
    \toprule
    \textbf{Dataset} & \textbf{Stem traces} & \textbf{Rollouts} & \textbf{Evaluation} \\
    \midrule
    GSM8K     & 1{,}024 & 1{,}024 &  8{,}192 \\
    GSM1K     & 1{,}024 & 1{,}024 &  8{,}192 \\
    MATH500   & 2{,}048 & 2{,}048 & 16{,}384 \\
    AIME 2025 & 8{,}192 & 4{,}096 & 32{,}768 \\
    \bottomrule
  \end{tabular}
\end{table}

\subsection{Prompt template}
\label[appendix]{app:prompt_template}
For all seven models across all datasets, we employ the same zero-shot prompt. We append the following instruction to the end of every input query:

\begin{promptbox}
\textbf{Prompt.} \textit{Please reason step by step, and put your final answer within \texttt{\textbackslash boxed\{\}}.}
\end{promptbox}

\newpage

\section{Cliff tokens vs. critical tokens}
\label[appendix]{app:critical_cliff}

\paragraph{Comparison setup.}
We compare cliff tokens with critical tokens from \citet{CriticalTokens}. In their definition, a critical token is a token whose success probability is $0$ and whose success probabilities for all subsequent tokens remain below $0.05$. In contrast, a cliff token is a token at which the drop in success probability exceeds our $z$-test-based adaptive threshold with $\tau = 0.1$. We evaluate both notions on the same incorrect subset of traces using deletion-based recovery:
\begin{itemize}[leftmargin=*, nosep]
    \item \textbf{Cliff-del:} resampling from the prefix before the first cliff token.
    \item \textbf{Critical-del:} resampling from the prefix before the critical token.
\end{itemize}
For each intervention, avg@64 is computed as the fraction of 64 temperature-sampled rollouts that yield a correct answer, and each table entry reports the mean avg@64 across traces in the corresponding incorrect subset. If an incorrect trace does not contain the corresponding target token, we assign avg@64 $=0$ for that intervention. See \Cref{app:experimental_details} for experimental details.

\paragraph{Results.}
\Cref{tab:exp2-avg64-allfail} shows that Cliff-del outperforms Critical-del in 18 of the 19 comparable model--dataset settings. Two additional settings, \texttt{Llama-3.1-8B} on AIME 2025 and \texttt{Llama-3.2-1B} on AIME 2025, are not directly comparable because no cliff token was identified in any incorrect trace. The only comparable exception is \texttt{Llama-3.2-3B} on AIME 2025, where the avg@$64$ under both interventions are at most $0.011$, indicating that neither deletion substantially restores correctness.

Overall, deleting the first cliff token leads to higher recovery than deleting the critical token in the majority of comparable settings. This supports the interpretation that cliff tokens identify the local point where the model shifts into an incorrect trace. Thus, the two notions capture different stages of failure: critical tokens correspond to points where an error has become persistent, whereas cliff tokens identify an earlier trigger associated with the collapse of the reasoning trace.

\begin{table}[h]
  \caption{Mean avg@$64$ after Cliff-del vs. Critical-del on the incorrect subset. \textbf{Bold} marks the larger value within each (model, dataset) block. `-' indicates no cliff token was identified in any incorrect trace for that model and dataset pair.}
  \label{tab:exp2-avg64-allfail}
  \centering
  \small
  \setlength{\tabcolsep}{4pt}
  \begin{tabular}{l cc cc cc}
    \toprule
     & \multicolumn{2}{c}{\textbf{GSM1K}}
     & \multicolumn{2}{c}{\textbf{MATH500}}
     & \multicolumn{2}{c}{\textbf{AIME 2025}} \\
    \cmidrule(lr){2-3} \cmidrule(lr){4-5} \cmidrule(lr){6-7}
    Model
    & Cliff-del
    & Critical-del
    & Cliff-del
    & Critical-del
    & Cliff-del
    & Critical-del \\
    \midrule
    \texttt{Qwen3-8B}     & \textbf{0.161} & 0.073          & \textbf{0.284} & 0.078          & \textbf{0.037} & 0.029 \\
    \texttt{Qwen3-4B}     & \textbf{0.132} & 0.047          & \textbf{0.312} & 0.101          & \textbf{0.072} & 0.021 \\
    \texttt{Qwen3-0.6B}   & \textbf{0.238} & 0.106          & \textbf{0.187} & 0.084          & \textbf{0.012} & 0.002 \\
    \texttt{Llama-3.1-8B} & \textbf{0.434} & 0.115          & \textbf{0.267} & 0.098          & -            & 0.008 \\
    \texttt{Llama-3.2-3B} & \textbf{0.336} & 0.149          & \textbf{0.173} & 0.098          & 0.006          & \textbf{0.011} \\
    \texttt{Llama-3.2-1B} & \textbf{0.230} & 0.090          & \textbf{0.125} & 0.051          & -            & 0.001 \\
    \texttt{Gemma-3-4B}   & \textbf{0.308} & 0.076          & \textbf{0.239} & 0.054          & \textbf{0.026} & 0.010 \\
    \bottomrule
  \end{tabular}
\end{table}

\newpage

\section{Cliff-token multiplicity and downstream correctness}
\label[appendix]{app:cliff_recovery_stats}

To complement the analysis in \Cref{sec:cliff_recoverable_dynamics}, we report model-wise evidence that cliff tokens occur within non-monotonic reasoning dynamics, where the estimated success probability fluctuates along the trace \citep{ThePotential}. Based on \Cref{tab:cliff_recovery_stats}, we identify two primary patterns in traces containing at least one cliff token. First, nearly half of these traces ($47.2\%$) contain multiple cliff tokens, indicating that reasoning errors frequently occur as repeated local failures. Second, a cliff token does not strictly determine an incorrect outcome; approximately one-third of the traces with cliff tokens ($32.4\%$) successfully recover to reach the correct final answer.

The proportion of traces that still reach the correct answer after cliff tokens increases with model size in both the \texttt{Qwen3} and \texttt{Llama-3} families. This pattern is consistent with greater downstream recovery capacity in larger models \citep{RobustIntervention, SingleUtteranceSelfCorrection}, potentially reflecting stronger self-correction.

\begin{table}[H]
  \centering
  \small
  \caption{Multiplicity and downstream correctness among traces containing at least one cliff token. Percentages are relative to the number of traces with any cliff token. Multiple-cliff traces contain at least two cliff tokens, and correct traces end with the correct answer.}
  \label{tab:cliff_recovery_stats}
  \setlength{\tabcolsep}{5pt}
  \begin{tabular}{lccc}
    \toprule
    \textbf{Model}
    & \begin{tabular}[c]{@{}c@{}}\textbf{Traces with any}\\ \textbf{cliff token}\end{tabular}
    & \begin{tabular}[c]{@{}c@{}}\textbf{Multiple-cliff traces,}\\ {\boldmath\textbf{$n$ (\%)}}\end{tabular}
    & \begin{tabular}[c]{@{}c@{}}\textbf{Correct traces,}\\ {\boldmath\textbf{$n$ (\%)}}\end{tabular}
    \\
    \midrule
    \texttt{Qwen3-8B}       & 33 & 14 (42.4\%) & 17 (51.5\%) \\
    \texttt{Qwen3-4B}       & 29 & 20 (69.0\%) & 12 (41.4\%) \\
    \texttt{Qwen3-0.6B}     & 58 & 36 (62.1\%) & 16 (27.6\%) \\
    \texttt{Llama-3.1-8B}   & 74 & 29 (39.2\%) & 26 (35.1\%) \\
    \texttt{Llama-3.2-3B}   & 56 & 25 (44.6\%) & 14 (25.0\%) \\
    \texttt{Llama-3.2-1B}   & 69 & 25 (36.2\%) & 16 (23.2\%) \\
    \texttt{Gemma-3-4B}     & 39 & 20 (51.3\%) & 15 (38.5\%) \\
    \midrule
    \textbf{Total} & \textbf{358} & \textbf{169 (47.2\%)} & \textbf{116 (32.4\%)} \\
    \bottomrule
  \end{tabular}
\end{table}

\section{Detailed Cliff-del and Cliff-keep results}
\label[appendix]{app:cliff_del_cliff_keep}

This section reports the complete pass@$k$ results for the Cliff-del and Cliff-keep interventions described in \Cref{sec:pass_at_k}. The tables provide mean pass@$k$ values, pointwise 95\% two-level bootstrap confidence intervals (CIs), paired differences between the interventions ($\Delta$), and the number of traces included in the analysis for each dataset--model combination. We repeat the following procedure $10{,}000$ times to calculate CIs: At each bootstrap iteration, we first resample traces with replacement and then resample the 64 rollout outcomes within each selected trace.

\subsection{Results for incorrect traces}

In \Cref{sec:pass_at_k}, the main analysis focuses on incorrect traces, where deleting the first cliff token recovers reasoning performance. \Cref{tab:passk_incorrect_firstcliff_gsm1k,tab:passk_incorrect_firstcliff_math500,tab:passk_incorrect_firstcliff_aime25} show detailed results for GSM1K, MATH500, and AIME 2025, respectively. Cliff-del yields equal or higher mean pass@$k$ compared to Cliff-keep in most settings, with a few exceptions. Cliff-del reaches pass@$64$ of $1{.}0$ in every model and benchmark setting included in the analysis, while Cliff-keep ranges from $0{.}67$ to $1{.}0$.

\begin{table}[H]
\centering
\caption{\textbf{GSM1K:} Cliff-del and Cliff-keep pass@$k$ on incorrect traces. Cells report mean pass@$k$ across traces and pointwise 95\% CIs. $\Delta$ is Cliff-del minus Cliff-keep and uses paired trace resamples. $n$ is the number of traces that contain at least one cliff token.}
\label{tab:passk_incorrect_firstcliff_gsm1k}
\label{tab:passk_incorrect_firstcliff}
\small
\setlength{\tabcolsep}{1.5pt}
\renewcommand{\arraystretch}{1.05}
\begin{adjustbox}{max width=\linewidth}
\begin{tabular}{l>{\centering\arraybackslash}m{0.65in}*{7}{c}}
\toprule
\textbf{Model} & \textbf{Setup} & {\boldmath$k=1$} & {\boldmath$k=2$} & {\boldmath$k=4$} & {\boldmath$k=8$} & {\boldmath$k=16$} & {\boldmath$k=32$} & {\boldmath$k=64$} \\
\midrule
\multirow{3}{*}{\shortstack[l]{\texttt{Qwen3-8B}\\{\footnotesize $(n = 4)$}}} & \setuplbl{Cliff-del} & \shortstack{0.43\\{\scriptsize [0.34,0.52]}} & \shortstack{0.68\\{\scriptsize [0.56,0.77]}} & \shortstack{0.89\\{\scriptsize [0.79,0.95]}} & \shortstack{0.99\\{\scriptsize [0.95,1.00]}} & \shortstack{1.00\\{\scriptsize [1.00,1.00]}} & \shortstack{1.00\\{\scriptsize [1.00,1.00]}} & \shortstack{1.00\\{\scriptsize [1.00,1.00]}} \\
\passkrule
 & \setuplbl{Cliff-keep} & \shortstack{0.28\\{\scriptsize [0.18,0.38]}} & \shortstack{0.48\\{\scriptsize [0.33,0.61]}} & \shortstack{0.72\\{\scriptsize [0.54,0.85]}} & \shortstack{0.91\\{\scriptsize [0.77,0.98]}} & \shortstack{0.99\\{\scriptsize [0.92,1.00]}} & \shortstack{1.00\\{\scriptsize [0.99,1.00]}} & \shortstack{1.00\\{\scriptsize [1.00,1.00]}} \\
\passkruledelta
 & \cellcolor{rowgray}\setuplbl{$\Delta$} & \cellcolor{rowgray}\deltapad\shortstack{0.15\\{\scriptsize [-0.01,0.31]}} & \cellcolor{rowgray}\shortstack{0.20\\{\scriptsize [-0.01,0.41]}} & \cellcolor{rowgray}\shortstack{0.17\\{\scriptsize [-0.02,0.38]}} & \cellcolor{rowgray}\shortstack{0.08\\{\scriptsize [-0.01,0.22]}} & \cellcolor{rowgray}\shortstack{0.01\\{\scriptsize [0.00,0.08]}} & \cellcolor{rowgray}\shortstack{0.00\\{\scriptsize [0.00,0.01]}} & \cellcolor{rowgray}\shortstack{0.00\\{\scriptsize [0.00,0.00]}} \\
\midrule
\multirow{3}{*}{\shortstack[l]{\texttt{Qwen3-4B}\\{\footnotesize $(n = 4)$}}} & \setuplbl{Cliff-del} & \shortstack{0.38\\{\scriptsize [0.22,0.53]}} & \shortstack{0.60\\{\scriptsize [0.38,0.77]}} & \shortstack{0.81\\{\scriptsize [0.59,0.95]}} & \shortstack{0.94\\{\scriptsize [0.78,1.00]}} & \shortstack{0.99\\{\scriptsize [0.92,1.00]}} & \shortstack{1.00\\{\scriptsize [0.99,1.00]}} & \shortstack{1.00\\{\scriptsize [1.00,1.00]}} \\
\passkrule
 & \setuplbl{Cliff-keep} & \shortstack{0.36\\{\scriptsize [0.23,0.47]}} & \shortstack{0.58\\{\scriptsize [0.39,0.72]}} & \shortstack{0.81\\{\scriptsize [0.61,0.92]}} & \shortstack{0.94\\{\scriptsize [0.81,1.00]}} & \shortstack{0.99\\{\scriptsize [0.94,1.00]}} & \shortstack{1.00\\{\scriptsize [0.99,1.00]}} & \shortstack{1.00\\{\scriptsize [1.00,1.00]}} \\
\passkruledelta
 & \cellcolor{rowgray}\setuplbl{$\Delta$} & \cellcolor{rowgray}\deltapad\shortstack{0.02\\{\scriptsize [-0.08,0.13]}} & \cellcolor{rowgray}\shortstack{0.02\\{\scriptsize [-0.11,0.15]}} & \cellcolor{rowgray}\shortstack{0.00\\{\scriptsize [-0.11,0.11]}} & \cellcolor{rowgray}\shortstack{-0.01\\{\scriptsize [-0.10,0.07]}} & \cellcolor{rowgray}\shortstack{0.00\\{\scriptsize [-0.06,0.04]}} & \cellcolor{rowgray}\shortstack{0.00\\{\scriptsize [-0.01,0.00]}} & \cellcolor{rowgray}\shortstack{0.00\\{\scriptsize [0.00,0.00]}} \\
\midrule
\multirow{3}{*}{\shortstack[l]{\texttt{Qwen3-0.6B}\\{\footnotesize $(n = 20)$}}} & \setuplbl{Cliff-del} & \shortstack{0.63\\{\scriptsize [0.54,0.71]}} & \shortstack{0.83\\{\scriptsize [0.76,0.88]}} & \shortstack{0.96\\{\scriptsize [0.92,0.98]}} & \shortstack{1.00\\{\scriptsize [0.99,1.00]}} & \shortstack{1.00\\{\scriptsize [1.00,1.00]}} & \shortstack{1.00\\{\scriptsize [1.00,1.00]}} & \shortstack{1.00\\{\scriptsize [1.00,1.00]}} \\
\passkrule
 & \setuplbl{Cliff-keep} & \shortstack{0.38\\{\scriptsize [0.29,0.47]}} & \shortstack{0.57\\{\scriptsize [0.47,0.67]}} & \shortstack{0.77\\{\scriptsize [0.68,0.85]}} & \shortstack{0.93\\{\scriptsize [0.87,0.96]}} & \shortstack{0.99\\{\scriptsize [0.97,1.00]}} & \shortstack{1.00\\{\scriptsize [1.00,1.00]}} & \shortstack{1.00\\{\scriptsize [1.00,1.00]}} \\
\passkruledelta
 & \cellcolor{rowgray}\setuplbl{$\Delta$} & \cellcolor{rowgray}\deltapad\shortstack{0.25\\{\scriptsize [0.17,0.34]}} & \cellcolor{rowgray}\shortstack{0.25\\{\scriptsize [0.17,0.34]}} & \cellcolor{rowgray}\shortstack{0.18\\{\scriptsize [0.11,0.27]}} & \cellcolor{rowgray}\shortstack{0.07\\{\scriptsize [0.04,0.13]}} & \cellcolor{rowgray}\shortstack{0.01\\{\scriptsize [0.00,0.03]}} & \cellcolor{rowgray}\shortstack{0.00\\{\scriptsize [0.00,0.00]}} & \cellcolor{rowgray}\shortstack{0.00\\{\scriptsize [0.00,0.00]}} \\
\midrule
\multirow{3}{*}{\shortstack[l]{\texttt{Llama-3.1-8B}\\{\footnotesize $(n = 21)$}}} & \setuplbl{Cliff-del} & \shortstack{0.60\\{\scriptsize [0.47,0.72]}} & \shortstack{0.76\\{\scriptsize [0.64,0.87]}} & \shortstack{0.88\\{\scriptsize [0.79,0.95]}} & \shortstack{0.96\\{\scriptsize [0.91,0.99]}} & \shortstack{0.99\\{\scriptsize [0.97,1.00]}} & \shortstack{1.00\\{\scriptsize [1.00,1.00]}} & \shortstack{1.00\\{\scriptsize [1.00,1.00]}} \\
\passkrule
 & \setuplbl{Cliff-keep} & \shortstack{0.23\\{\scriptsize [0.15,0.33]}} & \shortstack{0.37\\{\scriptsize [0.24,0.50]}} & \shortstack{0.52\\{\scriptsize [0.35,0.67]}} & \shortstack{0.63\\{\scriptsize [0.45,0.79]}} & \shortstack{0.72\\{\scriptsize [0.53,0.86]}} & \shortstack{0.79\\{\scriptsize [0.59,0.92]}} & \shortstack{0.86\\{\scriptsize [0.62,0.95]}} \\
\passkruledelta
 & \cellcolor{rowgray}\setuplbl{$\Delta$} & \cellcolor{rowgray}\deltapad\shortstack{0.37\\{\scriptsize [0.23,0.50]}} & \cellcolor{rowgray}\shortstack{0.39\\{\scriptsize [0.25,0.54]}} & \cellcolor{rowgray}\shortstack{0.37\\{\scriptsize [0.21,0.53]}} & \cellcolor{rowgray}\shortstack{0.32\\{\scriptsize [0.16,0.50]}} & \cellcolor{rowgray}\shortstack{0.27\\{\scriptsize [0.12,0.46]}} & \cellcolor{rowgray}\shortstack{0.21\\{\scriptsize [0.08,0.41]}} & \cellcolor{rowgray}\shortstack{0.14\\{\scriptsize [0.05,0.38]}} \\
\midrule
\multirow{3}{*}{\shortstack[l]{\texttt{Llama-3.2-3B}\\{\footnotesize $(n = 20)$}}} & \setuplbl{Cliff-del} & \shortstack{0.59\\{\scriptsize [0.48,0.70]}} & \shortstack{0.78\\{\scriptsize [0.68,0.86]}} & \shortstack{0.92\\{\scriptsize [0.85,0.96]}} & \shortstack{0.98\\{\scriptsize [0.95,1.00]}} & \shortstack{1.00\\{\scriptsize [0.99,1.00]}} & \shortstack{1.00\\{\scriptsize [1.00,1.00]}} & \shortstack{1.00\\{\scriptsize [1.00,1.00]}} \\
\passkrule
 & \setuplbl{Cliff-keep} & \shortstack{0.32\\{\scriptsize [0.20,0.44]}} & \shortstack{0.46\\{\scriptsize [0.31,0.61]}} & \shortstack{0.59\\{\scriptsize [0.42,0.74]}} & \shortstack{0.70\\{\scriptsize [0.53,0.84]}} & \shortstack{0.81\\{\scriptsize [0.64,0.91]}} & \shortstack{0.90\\{\scriptsize [0.74,0.97]}} & \shortstack{0.95\\{\scriptsize [0.80,1.00]}} \\
\passkruledelta
 & \cellcolor{rowgray}\setuplbl{$\Delta$} & \cellcolor{rowgray}\deltapad\shortstack{0.27\\{\scriptsize [0.15,0.41]}} & \cellcolor{rowgray}\shortstack{0.32\\{\scriptsize [0.18,0.46]}} & \cellcolor{rowgray}\shortstack{0.33\\{\scriptsize [0.18,0.48]}} & \cellcolor{rowgray}\shortstack{0.28\\{\scriptsize [0.15,0.44]}} & \cellcolor{rowgray}\shortstack{0.19\\{\scriptsize [0.08,0.35]}} & \cellcolor{rowgray}\shortstack{0.10\\{\scriptsize [0.03,0.26]}} & \cellcolor{rowgray}\shortstack{0.05\\{\scriptsize [0.00,0.20]}} \\
\midrule
\multirow{3}{*}{\shortstack[l]{\texttt{Llama-3.2-1B}\\{\footnotesize $(n = 31)$}}} & \setuplbl{Cliff-del} & \shortstack{0.52\\{\scriptsize [0.43,0.61]}} & \shortstack{0.71\\{\scriptsize [0.63,0.79]}} & \shortstack{0.87\\{\scriptsize [0.81,0.92]}} & \shortstack{0.97\\{\scriptsize [0.93,0.98]}} & \shortstack{1.00\\{\scriptsize [0.98,1.00]}} & \shortstack{1.00\\{\scriptsize [1.00,1.00]}} & \shortstack{1.00\\{\scriptsize [1.00,1.00]}} \\
\passkrule
 & \setuplbl{Cliff-keep} & \shortstack{0.20\\{\scriptsize [0.12,0.28]}} & \shortstack{0.31\\{\scriptsize [0.20,0.42]}} & \shortstack{0.44\\{\scriptsize [0.30,0.56]}} & \shortstack{0.55\\{\scriptsize [0.40,0.68]}} & \shortstack{0.66\\{\scriptsize [0.49,0.78]}} & \shortstack{0.74\\{\scriptsize [0.56,0.85]}} & \shortstack{0.81\\{\scriptsize [0.61,0.90]}} \\
\passkruledelta
 & \cellcolor{rowgray}\setuplbl{$\Delta$} & \cellcolor{rowgray}\deltapad\shortstack{0.32\\{\scriptsize [0.23,0.42]}} & \cellcolor{rowgray}\shortstack{0.40\\{\scriptsize [0.29,0.51]}} & \cellcolor{rowgray}\shortstack{0.44\\{\scriptsize [0.31,0.56]}} & \cellcolor{rowgray}\shortstack{0.41\\{\scriptsize [0.28,0.56]}} & \cellcolor{rowgray}\shortstack{0.34\\{\scriptsize [0.22,0.50]}} & \cellcolor{rowgray}\shortstack{0.26\\{\scriptsize [0.15,0.44]}} & \cellcolor{rowgray}\shortstack{0.19\\{\scriptsize [0.10,0.39]}} \\
\midrule
\multirow{3}{*}{\shortstack[l]{\texttt{Gemma-3-4B}\\{\footnotesize $(n = 8)$}}} & \setuplbl{Cliff-del} & \shortstack{0.67\\{\scriptsize [0.56,0.77]}} & \shortstack{0.87\\{\scriptsize [0.79,0.94]}} & \shortstack{0.98\\{\scriptsize [0.94,1.00]}} & \shortstack{1.00\\{\scriptsize [0.99,1.00]}} & \shortstack{1.00\\{\scriptsize [1.00,1.00]}} & \shortstack{1.00\\{\scriptsize [1.00,1.00]}} & \shortstack{1.00\\{\scriptsize [1.00,1.00]}} \\
\passkrule
 & \setuplbl{Cliff-keep} & \shortstack{0.49\\{\scriptsize [0.26,0.67]}} & \shortstack{0.66\\{\scriptsize [0.35,0.89]}} & \shortstack{0.74\\{\scriptsize [0.40,0.99]}} & \shortstack{0.77\\{\scriptsize [0.44,1.00]}} & \shortstack{0.78\\{\scriptsize [0.49,1.00]}} & \shortstack{0.81\\{\scriptsize [0.50,1.00]}} & \shortstack{0.88\\{\scriptsize [0.50,1.00]}} \\
\passkruledelta
 & \cellcolor{rowgray}\setuplbl{$\Delta$} & \cellcolor{rowgray}\deltapad\shortstack{0.18\\{\scriptsize [0.02,0.37]}} & \cellcolor{rowgray}\shortstack{0.22\\{\scriptsize [0.01,0.47]}} & \cellcolor{rowgray}\shortstack{0.23\\{\scriptsize [0.00,0.53]}} & \cellcolor{rowgray}\shortstack{0.23\\{\scriptsize [0.00,0.55]}} & \cellcolor{rowgray}\shortstack{0.22\\{\scriptsize [0.00,0.50]}} & \cellcolor{rowgray}\shortstack{0.19\\{\scriptsize [0.00,0.50]}} & \cellcolor{rowgray}\shortstack{0.13\\{\scriptsize [0.00,0.50]}} \\
\bottomrule
\end{tabular}
\end{adjustbox}
\end{table}

\begin{table}[H]
\centering
\caption{\textbf{MATH500:} Cliff-del and Cliff-keep pass@$k$ on incorrect traces. Notations follow \Cref{tab:passk_incorrect_firstcliff_gsm1k}.}
\label{tab:passk_incorrect_firstcliff_math500}
\small
\setlength{\tabcolsep}{1.5pt}
\renewcommand{\arraystretch}{1.05}
\begin{adjustbox}{max width=\linewidth}
\begin{tabular}{l>{\centering\arraybackslash}m{0.65in}*{7}{c}}
\toprule
\textbf{Model} & \textbf{Setup} & {\boldmath$k=1$} & {\boldmath$k=2$} & {\boldmath$k=4$} & {\boldmath$k=8$} & {\boldmath$k=16$} & {\boldmath$k=32$} & {\boldmath$k=64$} \\
\midrule
\multirow{3}{*}{\shortstack[l]{\texttt{Qwen3-8B}\\{\footnotesize $(n = 9)$}}} & \setuplbl{Cliff-del} & \shortstack{0.45\\{\scriptsize [0.37,0.54]}} & \shortstack{0.69\\{\scriptsize [0.59,0.78]}} & \shortstack{0.89\\{\scriptsize [0.80,0.95]}} & \shortstack{0.98\\{\scriptsize [0.94,1.00]}} & \shortstack{1.00\\{\scriptsize [0.99,1.00]}} & \shortstack{1.00\\{\scriptsize [1.00,1.00]}} & \shortstack{1.00\\{\scriptsize [1.00,1.00]}} \\
\passkrule
 & \setuplbl{Cliff-keep} & \shortstack{0.34\\{\scriptsize [0.19,0.48]}} & \shortstack{0.52\\{\scriptsize [0.31,0.70]}} & \shortstack{0.68\\{\scriptsize [0.43,0.88]}} & \shortstack{0.78\\{\scriptsize [0.53,0.97]}} & \shortstack{0.84\\{\scriptsize [0.58,1.00]}} & \shortstack{0.88\\{\scriptsize [0.64,1.00]}} & \shortstack{0.89\\{\scriptsize [0.67,1.00]}} \\
\passkruledelta
 & \cellcolor{rowgray}\setuplbl{$\Delta$} & \cellcolor{rowgray}\deltapad\shortstack{0.12\\{\scriptsize [-0.01,0.27]}} & \cellcolor{rowgray}\shortstack{0.17\\{\scriptsize [0.00,0.38]}} & \cellcolor{rowgray}\shortstack{0.21\\{\scriptsize [0.02,0.44]}} & \cellcolor{rowgray}\shortstack{0.20\\{\scriptsize [0.01,0.45]}} & \cellcolor{rowgray}\shortstack{0.16\\{\scriptsize [0.00,0.41]}} & \cellcolor{rowgray}\shortstack{0.12\\{\scriptsize [0.00,0.36]}} & \cellcolor{rowgray}\shortstack{0.11\\{\scriptsize [0.00,0.33]}} \\
\midrule
\multirow{3}{*}{\shortstack[l]{\texttt{Qwen3-4B}\\{\footnotesize $(n = 9)$}}} & \setuplbl{Cliff-del} & \shortstack{0.58\\{\scriptsize [0.48,0.68]}} & \shortstack{0.81\\{\scriptsize [0.71,0.89]}} & \shortstack{0.95\\{\scriptsize [0.89,0.98]}} & \shortstack{1.00\\{\scriptsize [0.98,1.00]}} & \shortstack{1.00\\{\scriptsize [1.00,1.00]}} & \shortstack{1.00\\{\scriptsize [1.00,1.00]}} & \shortstack{1.00\\{\scriptsize [1.00,1.00]}} \\
\passkrule
 & \setuplbl{Cliff-keep} & \shortstack{0.43\\{\scriptsize [0.30,0.54]}} & \shortstack{0.64\\{\scriptsize [0.47,0.78]}} & \shortstack{0.82\\{\scriptsize [0.64,0.94]}} & \shortstack{0.92\\{\scriptsize [0.78,1.00]}} & \shortstack{0.97\\{\scriptsize [0.88,1.00]}} & \shortstack{1.00\\{\scriptsize [0.94,1.00]}} & \shortstack{1.00\\{\scriptsize [1.00,1.00]}} \\
\passkruledelta
 & \cellcolor{rowgray}\setuplbl{$\Delta$} & \cellcolor{rowgray}\deltapad\shortstack{0.15\\{\scriptsize [0.06,0.25]}} & \cellcolor{rowgray}\shortstack{0.16\\{\scriptsize [0.06,0.28]}} & \cellcolor{rowgray}\shortstack{0.13\\{\scriptsize [0.03,0.27]}} & \cellcolor{rowgray}\shortstack{0.07\\{\scriptsize [0.00,0.21]}} & \cellcolor{rowgray}\shortstack{0.03\\{\scriptsize [0.00,0.12]}} & \cellcolor{rowgray}\shortstack{0.00\\{\scriptsize [0.00,0.06]}} & \cellcolor{rowgray}\shortstack{0.00\\{\scriptsize [0.00,0.00]}} \\
\midrule
\multirow{3}{*}{\shortstack[l]{\texttt{Qwen3-0.6B}\\{\footnotesize $(n = 21)$}}} & \setuplbl{Cliff-del} & \shortstack{0.44\\{\scriptsize [0.35,0.54]}} & \shortstack{0.64\\{\scriptsize [0.54,0.74]}} & \shortstack{0.82\\{\scriptsize [0.72,0.90]}} & \shortstack{0.93\\{\scriptsize [0.85,0.98]}} & \shortstack{0.98\\{\scriptsize [0.92,1.00]}} & \shortstack{1.00\\{\scriptsize [0.96,1.00]}} & \shortstack{1.00\\{\scriptsize [1.00,1.00]}} \\
\passkrule
 & \setuplbl{Cliff-keep} & \shortstack{0.31\\{\scriptsize [0.20,0.42]}} & \shortstack{0.46\\{\scriptsize [0.31,0.60]}} & \shortstack{0.60\\{\scriptsize [0.43,0.75]}} & \shortstack{0.71\\{\scriptsize [0.52,0.86]}} & \shortstack{0.76\\{\scriptsize [0.57,0.92]}} & \shortstack{0.79\\{\scriptsize [0.60,0.95]}} & \shortstack{0.81\\{\scriptsize [0.62,0.95]}} \\
\passkruledelta
 & \cellcolor{rowgray}\setuplbl{$\Delta$} & \cellcolor{rowgray}\deltapad\shortstack{0.13\\{\scriptsize [0.05,0.23]}} & \cellcolor{rowgray}\shortstack{0.18\\{\scriptsize [0.06,0.31]}} & \cellcolor{rowgray}\shortstack{0.22\\{\scriptsize [0.07,0.38]}} & \cellcolor{rowgray}\shortstack{0.23\\{\scriptsize [0.07,0.40]}} & \cellcolor{rowgray}\shortstack{0.22\\{\scriptsize [0.05,0.40]}} & \cellcolor{rowgray}\shortstack{0.21\\{\scriptsize [0.05,0.39]}} & \cellcolor{rowgray}\shortstack{0.19\\{\scriptsize [0.05,0.38]}} \\
\midrule
\multirow{3}{*}{\shortstack[l]{\texttt{Llama-3.1-8B}\\{\footnotesize $(n = 27)$}}} & \setuplbl{Cliff-del} & \shortstack{0.54\\{\scriptsize [0.45,0.63]}} & \shortstack{0.73\\{\scriptsize [0.63,0.82]}} & \shortstack{0.87\\{\scriptsize [0.78,0.94]}} & \shortstack{0.95\\{\scriptsize [0.88,0.99]}} & \shortstack{0.98\\{\scriptsize [0.93,1.00]}} & \shortstack{1.00\\{\scriptsize [0.96,1.00]}} & \shortstack{1.00\\{\scriptsize [1.00,1.00]}} \\
\passkrule
 & \setuplbl{Cliff-keep} & \shortstack{0.27\\{\scriptsize [0.18,0.37]}} & \shortstack{0.41\\{\scriptsize [0.28,0.53]}} & \shortstack{0.54\\{\scriptsize [0.39,0.67]}} & \shortstack{0.65\\{\scriptsize [0.50,0.78]}} & \shortstack{0.76\\{\scriptsize [0.59,0.87]}} & \shortstack{0.82\\{\scriptsize [0.66,0.93]}} & \shortstack{0.85\\{\scriptsize [0.70,0.96]}} \\
\passkruledelta
 & \cellcolor{rowgray}\setuplbl{$\Delta$} & \cellcolor{rowgray}\deltapad\shortstack{0.27\\{\scriptsize [0.15,0.39]}} & \cellcolor{rowgray}\shortstack{0.33\\{\scriptsize [0.19,0.47]}} & \cellcolor{rowgray}\shortstack{0.34\\{\scriptsize [0.19,0.48]}} & \cellcolor{rowgray}\shortstack{0.29\\{\scriptsize [0.15,0.45]}} & \cellcolor{rowgray}\shortstack{0.22\\{\scriptsize [0.10,0.39]}} & \cellcolor{rowgray}\shortstack{0.18\\{\scriptsize [0.06,0.34]}} & \cellcolor{rowgray}\shortstack{0.15\\{\scriptsize [0.04,0.30]}} \\
\midrule
\multirow{3}{*}{\shortstack[l]{\texttt{Llama-3.2-3B}\\{\footnotesize $(n = 21)$}}} & \setuplbl{Cliff-del} & \shortstack{0.47\\{\scriptsize [0.35,0.61]}} & \shortstack{0.63\\{\scriptsize [0.51,0.75]}} & \shortstack{0.80\\{\scriptsize [0.70,0.88]}} & \shortstack{0.93\\{\scriptsize [0.86,0.97]}} & \shortstack{0.99\\{\scriptsize [0.96,1.00]}} & \shortstack{1.00\\{\scriptsize [0.99,1.00]}} & \shortstack{1.00\\{\scriptsize [1.00,1.00]}} \\
\passkrule
 & \setuplbl{Cliff-keep} & \shortstack{0.15\\{\scriptsize [0.07,0.24]}} & \shortstack{0.24\\{\scriptsize [0.12,0.36]}} & \shortstack{0.35\\{\scriptsize [0.20,0.50]}} & \shortstack{0.45\\{\scriptsize [0.27,0.63]}} & \shortstack{0.53\\{\scriptsize [0.33,0.71]}} & \shortstack{0.59\\{\scriptsize [0.37,0.77]}} & \shortstack{0.67\\{\scriptsize [0.43,0.81]}} \\
\passkruledelta
 & \cellcolor{rowgray}\setuplbl{$\Delta$} & \cellcolor{rowgray}\deltapad\shortstack{0.32\\{\scriptsize [0.19,0.47]}} & \cellcolor{rowgray}\shortstack{0.39\\{\scriptsize [0.24,0.55]}} & \cellcolor{rowgray}\shortstack{0.45\\{\scriptsize [0.28,0.62]}} & \cellcolor{rowgray}\shortstack{0.48\\{\scriptsize [0.30,0.66]}} & \cellcolor{rowgray}\shortstack{0.46\\{\scriptsize [0.27,0.66]}} & \cellcolor{rowgray}\shortstack{0.41\\{\scriptsize [0.23,0.63]}} & \cellcolor{rowgray}\shortstack{0.33\\{\scriptsize [0.19,0.62]}} \\
\midrule
\multirow{3}{*}{\shortstack[l]{\texttt{Llama-3.2-1B}\\{\footnotesize $(n = 22)$}}} & \setuplbl{Cliff-del} & \shortstack{0.43\\{\scriptsize [0.33,0.54]}} & \shortstack{0.62\\{\scriptsize [0.50,0.72]}} & \shortstack{0.79\\{\scriptsize [0.68,0.88]}} & \shortstack{0.91\\{\scriptsize [0.82,0.97]}} & \shortstack{0.97\\{\scriptsize [0.90,1.00]}} & \shortstack{0.99\\{\scriptsize [0.95,1.00]}} & \shortstack{1.00\\{\scriptsize [0.96,1.00]}} \\
\passkrule
 & \setuplbl{Cliff-keep} & \shortstack{0.23\\{\scriptsize [0.14,0.33]}} & \shortstack{0.37\\{\scriptsize [0.24,0.50]}} & \shortstack{0.51\\{\scriptsize [0.35,0.66]}} & \shortstack{0.64\\{\scriptsize [0.46,0.79]}} & \shortstack{0.73\\{\scriptsize [0.54,0.88]}} & \shortstack{0.79\\{\scriptsize [0.59,0.93]}} & \shortstack{0.82\\{\scriptsize [0.64,0.96]}} \\
\passkruledelta
 & \cellcolor{rowgray}\setuplbl{$\Delta$} & \cellcolor{rowgray}\deltapad\shortstack{0.20\\{\scriptsize [0.12,0.28]}} & \cellcolor{rowgray}\shortstack{0.25\\{\scriptsize [0.14,0.37]}} & \cellcolor{rowgray}\shortstack{0.28\\{\scriptsize [0.15,0.42]}} & \cellcolor{rowgray}\shortstack{0.27\\{\scriptsize [0.12,0.43]}} & \cellcolor{rowgray}\shortstack{0.24\\{\scriptsize [0.09,0.42]}} & \cellcolor{rowgray}\shortstack{0.21\\{\scriptsize [0.06,0.40]}} & \cellcolor{rowgray}\shortstack{0.18\\{\scriptsize [0.05,0.36]}} \\
\midrule
\multirow{3}{*}{\shortstack[l]{\texttt{Gemma-3-4B}\\{\footnotesize $(n = 14)$}}} & \setuplbl{Cliff-del} & \shortstack{0.48\\{\scriptsize [0.37,0.60]}} & \shortstack{0.69\\{\scriptsize [0.56,0.80]}} & \shortstack{0.86\\{\scriptsize [0.75,0.94]}} & \shortstack{0.96\\{\scriptsize [0.89,0.99]}} & \shortstack{0.99\\{\scriptsize [0.97,1.00]}} & \shortstack{1.00\\{\scriptsize [1.00,1.00]}} & \shortstack{1.00\\{\scriptsize [1.00,1.00]}} \\
\passkrule
 & \setuplbl{Cliff-keep} & \shortstack{0.39\\{\scriptsize [0.27,0.53]}} & \shortstack{0.58\\{\scriptsize [0.43,0.72]}} & \shortstack{0.75\\{\scriptsize [0.62,0.87]}} & \shortstack{0.90\\{\scriptsize [0.79,0.96]}} & \shortstack{0.98\\{\scriptsize [0.92,1.00]}} & \shortstack{1.00\\{\scriptsize [0.98,1.00]}} & \shortstack{1.00\\{\scriptsize [1.00,1.00]}} \\
\passkruledelta
 & \cellcolor{rowgray}\setuplbl{$\Delta$} & \cellcolor{rowgray}\deltapad\shortstack{0.09\\{\scriptsize [-0.01,0.19]}} & \cellcolor{rowgray}\shortstack{0.11\\{\scriptsize [-0.01,0.24]}} & \cellcolor{rowgray}\shortstack{0.10\\{\scriptsize [-0.01,0.23]}} & \cellcolor{rowgray}\shortstack{0.06\\{\scriptsize [-0.01,0.16]}} & \cellcolor{rowgray}\shortstack{0.02\\{\scriptsize [-0.01,0.07]}} & \cellcolor{rowgray}\shortstack{0.00\\{\scriptsize [0.00,0.02]}} & \cellcolor{rowgray}\shortstack{0.00\\{\scriptsize [0.00,0.00]}} \\
\bottomrule
\end{tabular}
\end{adjustbox}
\end{table}

\begin{table}[H]
\centering
\caption{\textbf{AIME 2025:} Cliff-del and Cliff-keep pass@$k$ on incorrect traces. \texttt{n/e} indicates that the CI is not estimated since $n<2$. \texttt{--} indicates that no traces were included in the analysis. Notations follow \Cref{tab:passk_incorrect_firstcliff_gsm1k}.}
\label{tab:passk_incorrect_firstcliff_aime25}
\small
\setlength{\tabcolsep}{1.5pt}
\renewcommand{\arraystretch}{1.05}
\begin{adjustbox}{max width=\linewidth}
\begin{tabular}{l>{\centering\arraybackslash}m{0.65in}*{7}{c}}
\toprule
\textbf{Model} & \textbf{Setup} & {\boldmath$k=1$} & {\boldmath$k=2$} & {\boldmath$k=4$} & {\boldmath$k=8$} & {\boldmath$k=16$} & {\boldmath$k=32$} & {\boldmath$k=64$} \\
\midrule
\multirow{3}{*}{\shortstack[l]{\texttt{Qwen3-8B}\\{\footnotesize $(n = 3)$}}} & \setuplbl{Cliff-del} & \shortstack{0.22\\{\scriptsize [0.14,0.32]}} & \shortstack{0.40\\{\scriptsize [0.25,0.54]}} & \shortstack{0.63\\{\scriptsize [0.43,0.79]}} & \shortstack{0.86\\{\scriptsize [0.67,0.96]}} & \shortstack{0.98\\{\scriptsize [0.87,1.00]}} & \shortstack{1.00\\{\scriptsize [0.98,1.00]}} & \shortstack{1.00\\{\scriptsize [1.00,1.00]}} \\
\passkrule
 & \setuplbl{Cliff-keep} & \shortstack{0.12\\{\scriptsize [0.00,0.23]}} & \shortstack{0.21\\{\scriptsize [0.00,0.40]}} & \shortstack{0.35\\{\scriptsize [0.00,0.64]}} & \shortstack{0.52\\{\scriptsize [0.00,0.87]}} & \shortstack{0.64\\{\scriptsize [0.00,0.99]}} & \shortstack{0.67\\{\scriptsize [0.00,1.00]}} & \shortstack{0.67\\{\scriptsize [0.00,1.00]}} \\
\passkruledelta
 & \cellcolor{rowgray}\setuplbl{$\Delta$} & \cellcolor{rowgray}\deltapad\shortstack{0.11\\{\scriptsize [-0.01,0.24]}} & \cellcolor{rowgray}\shortstack{0.19\\{\scriptsize [-0.02,0.42]}} & \cellcolor{rowgray}\shortstack{0.28\\{\scriptsize [-0.03,0.64]}} & \cellcolor{rowgray}\shortstack{0.34\\{\scriptsize [-0.05,0.87]}} & \cellcolor{rowgray}\shortstack{0.34\\{\scriptsize [-0.04,0.99]}} & \cellcolor{rowgray}\shortstack{0.33\\{\scriptsize [-0.01,1.00]}} & \cellcolor{rowgray}\shortstack{0.33\\{\scriptsize [0.00,1.00]}} \\
\midrule
\multirow{3}{*}{\shortstack[l]{\texttt{Qwen3-4B}\\{\footnotesize $(n = 4)$}}} & \setuplbl{Cliff-del} & \shortstack{0.47\\{\scriptsize [0.24,0.74]}} & \shortstack{0.65\\{\scriptsize [0.41,0.89]}} & \shortstack{0.83\\{\scriptsize [0.63,0.97]}} & \shortstack{0.95\\{\scriptsize [0.83,1.00]}} & \shortstack{1.00\\{\scriptsize [0.95,1.00]}} & \shortstack{1.00\\{\scriptsize [1.00,1.00]}} & \shortstack{1.00\\{\scriptsize [1.00,1.00]}} \\
\passkrule
 & \setuplbl{Cliff-keep} & \shortstack{0.35\\{\scriptsize [0.11,0.54]}} & \shortstack{0.54\\{\scriptsize [0.18,0.78]}} & \shortstack{0.69\\{\scriptsize [0.23,0.95]}} & \shortstack{0.75\\{\scriptsize [0.25,1.00]}} & \shortstack{0.75\\{\scriptsize [0.25,1.00]}} & \shortstack{0.75\\{\scriptsize [0.25,1.00]}} & \shortstack{0.75\\{\scriptsize [0.25,1.00]}} \\
\passkruledelta
 & \cellcolor{rowgray}\setuplbl{$\Delta$} & \cellcolor{rowgray}\deltapad\shortstack{0.11\\{\scriptsize [-0.06,0.28]}} & \cellcolor{rowgray}\shortstack{0.12\\{\scriptsize [-0.06,0.29]}} & \cellcolor{rowgray}\shortstack{0.14\\{\scriptsize [-0.05,0.43]}} & \cellcolor{rowgray}\shortstack{0.21\\{\scriptsize [-0.02,0.61]}} & \cellcolor{rowgray}\shortstack{0.25\\{\scriptsize [0.00,0.73]}} & \cellcolor{rowgray}\shortstack{0.25\\{\scriptsize [0.00,0.75]}} & \cellcolor{rowgray}\shortstack{0.25\\{\scriptsize [0.00,0.75]}} \\
\midrule
\multirow{3}{*}{\shortstack[l]{\texttt{Qwen3-0.6B}\\{\footnotesize $(n = 1)$}}} & \setuplbl{Cliff-del} & \shortstack{0.44\\{\scriptsize [n/e]}} & \shortstack{0.69\\{\scriptsize [n/e]}} & \shortstack{0.91\\{\scriptsize [n/e]}} & \shortstack{0.99\\{\scriptsize [n/e]}} & \shortstack{1.00\\{\scriptsize [n/e]}} & \shortstack{1.00\\{\scriptsize [n/e]}} & \shortstack{1.00\\{\scriptsize [n/e]}} \\
\passkrule
 & \setuplbl{Cliff-keep} & \shortstack{0.38\\{\scriptsize [n/e]}} & \shortstack{0.61\\{\scriptsize [n/e]}} & \shortstack{0.86\\{\scriptsize [n/e]}} & \shortstack{0.98\\{\scriptsize [n/e]}} & \shortstack{1.00\\{\scriptsize [n/e]}} & \shortstack{1.00\\{\scriptsize [n/e]}} & \shortstack{1.00\\{\scriptsize [n/e]}} \\
\passkruledelta
 & \cellcolor{rowgray}\setuplbl{$\Delta$} & \cellcolor{rowgray}\deltapad\shortstack{0.06\\{\scriptsize [n/e]}} & \cellcolor{rowgray}\shortstack{0.07\\{\scriptsize [n/e]}} & \cellcolor{rowgray}\shortstack{0.05\\{\scriptsize [n/e]}} & \cellcolor{rowgray}\shortstack{0.01\\{\scriptsize [n/e]}} & \cellcolor{rowgray}\shortstack{0.00\\{\scriptsize [n/e]}} & \cellcolor{rowgray}\shortstack{0.00\\{\scriptsize [n/e]}} & \cellcolor{rowgray}\shortstack{0.00\\{\scriptsize [n/e]}} \\
\midrule
\multirow{3}{*}{\shortstack[l]{\texttt{Llama-3.1-8B}\\{\footnotesize $(n = 0)$}}} & Cliff-del & -- & -- & -- & -- & -- & -- & -- \\
\passkrule
 & Cliff-keep & -- & -- & -- & -- & -- & -- & -- \\
\passkruledelta
 & \cellcolor{rowgray}$\Delta$ & \cellcolor{rowgray}-- & \cellcolor{rowgray}-- & \cellcolor{rowgray}-- & \cellcolor{rowgray}-- & \cellcolor{rowgray}-- & \cellcolor{rowgray}-- & \cellcolor{rowgray}-- \\
\midrule
\multirow{3}{*}{\shortstack[l]{\texttt{Llama-3.2-3B}\\{\footnotesize $(n = 1)$}}} & \setuplbl{Cliff-del} & \shortstack{0.25\\{\scriptsize [n/e]}} & \shortstack{0.44\\{\scriptsize [n/e]}} & \shortstack{0.69\\{\scriptsize [n/e]}} & \shortstack{0.92\\{\scriptsize [n/e]}} & \shortstack{1.00\\{\scriptsize [n/e]}} & \shortstack{1.00\\{\scriptsize [n/e]}} & \shortstack{1.00\\{\scriptsize [n/e]}} \\
\passkrule
 & \setuplbl{Cliff-keep} & \shortstack{0.23\\{\scriptsize [n/e]}} & \shortstack{0.42\\{\scriptsize [n/e]}} & \shortstack{0.67\\{\scriptsize [n/e]}} & \shortstack{0.90\\{\scriptsize [n/e]}} & \shortstack{0.99\\{\scriptsize [n/e]}} & \shortstack{1.00\\{\scriptsize [n/e]}} & \shortstack{1.00\\{\scriptsize [n/e]}} \\
\passkruledelta
 & \cellcolor{rowgray}\setuplbl{$\Delta$} & \cellcolor{rowgray}\deltapad\shortstack{0.02\\{\scriptsize [n/e]}} & \cellcolor{rowgray}\shortstack{0.02\\{\scriptsize [n/e]}} & \cellcolor{rowgray}\shortstack{0.03\\{\scriptsize [n/e]}} & \cellcolor{rowgray}\shortstack{0.02\\{\scriptsize [n/e]}} & \cellcolor{rowgray}\shortstack{0.00\\{\scriptsize [n/e]}} & \cellcolor{rowgray}\shortstack{0.00\\{\scriptsize [n/e]}} & \cellcolor{rowgray}\shortstack{0.00\\{\scriptsize [n/e]}} \\
\midrule
\multirow{3}{*}{\shortstack[l]{\texttt{Llama-3.2-1B}\\{\footnotesize $(n = 0)$}}} & Cliff-del & -- & -- & -- & -- & -- & -- & -- \\
\passkrule
 & Cliff-keep & -- & -- & -- & -- & -- & -- & -- \\
\passkruledelta
 & \cellcolor{rowgray}$\Delta$ & \cellcolor{rowgray}-- & \cellcolor{rowgray}-- & \cellcolor{rowgray}-- & \cellcolor{rowgray}-- & \cellcolor{rowgray}-- & \cellcolor{rowgray}-- & \cellcolor{rowgray}-- \\
\midrule
\multirow{3}{*}{\shortstack[l]{\texttt{Gemma-3-4B}\\{\footnotesize $(n = 2)$}}} & \setuplbl{Cliff-del} & \shortstack{0.37\\{\scriptsize [0.16,0.59]}} & \shortstack{0.58\\{\scriptsize [0.29,0.83]}} & \shortstack{0.78\\{\scriptsize [0.50,0.97]}} & \shortstack{0.93\\{\scriptsize [0.75,1.00]}} & \shortstack{0.99\\{\scriptsize [0.93,1.00]}} & \shortstack{1.00\\{\scriptsize [1.00,1.00]}} & \shortstack{1.00\\{\scriptsize [1.00,1.00]}} \\
\passkrule
 & \setuplbl{Cliff-keep} & \shortstack{0.16\\{\scriptsize [0.10,0.23]}} & \shortstack{0.30\\{\scriptsize [0.19,0.42]}} & \shortstack{0.52\\{\scriptsize [0.35,0.66]}} & \shortstack{0.78\\{\scriptsize [0.59,0.89]}} & \shortstack{0.96\\{\scriptsize [0.83,0.99]}} & \shortstack{1.00\\{\scriptsize [0.97,1.00]}} & \shortstack{1.00\\{\scriptsize [1.00,1.00]}} \\
\passkruledelta
 & \cellcolor{rowgray}\setuplbl{$\Delta$} & \cellcolor{rowgray}\deltapad\shortstack{0.20\\{\scriptsize [-0.03,0.45]}} & \cellcolor{rowgray}\shortstack{0.27\\{\scriptsize [-0.05,0.58]}} & \cellcolor{rowgray}\shortstack{0.26\\{\scriptsize [-0.07,0.56]}} & \cellcolor{rowgray}\shortstack{0.15\\{\scriptsize [-0.07,0.38]}} & \cellcolor{rowgray}\shortstack{0.03\\{\scriptsize [-0.03,0.16]}} & \cellcolor{rowgray}\shortstack{0.00\\{\scriptsize [0.00,0.03]}} & \cellcolor{rowgray}\shortstack{0.00\\{\scriptsize [0.00,0.00]}} \\
\bottomrule
\end{tabular}
\end{adjustbox}
\end{table}

\clearpage
\subsection{Results for correct traces}

In addition to \Cref{sec:pass_at_k}, we report the complementary analysis on correct traces. \Cref{tab:passk_correct_firstcliff_gsm1k,tab:passk_correct_firstcliff_math500,tab:passk_correct_firstcliff_aime25} show the corresponding results for GSM1K, MATH500, and AIME 2025, respectively. Cliff-del yields equal or higher mean pass@$k$ compared to Cliff-keep in most settings, with only a few exceptions. Overall, these results show that cliff tokens reduce success probability even in traces that reach the correct answer.

\begin{table}[H]
\centering
\caption{\textbf{GSM1K:} Cliff-del and Cliff-keep pass@$k$ on correct traces. Notations follow \Cref{tab:passk_incorrect_firstcliff_gsm1k}.}
\label{tab:passk_correct_firstcliff_gsm1k}
\label{tab:passk_correct_firstcliff}
\small
\setlength{\tabcolsep}{1.5pt}
\renewcommand{\arraystretch}{1.05}
\begin{adjustbox}{max width=\linewidth}
\begin{tabular}{l>{\centering\arraybackslash}m{0.65in}*{7}{c}}
\toprule
\textbf{Model} & \textbf{Setup} & {\boldmath$k=1$} & {\boldmath$k=2$} & {\boldmath$k=4$} & {\boldmath$k=8$} & {\boldmath$k=16$} & {\boldmath$k=32$} & {\boldmath$k=64$} \\
\midrule
\multirow{3}{*}{\shortstack[l]{\texttt{Qwen3-8B}\\{\footnotesize $(n = 2)$}}} & \setuplbl{Cliff-del} & \shortstack{0.72\\{\scriptsize [0.54,0.88]}} & \shortstack{0.91\\{\scriptsize [0.79,0.99]}} & \shortstack{0.99\\{\scriptsize [0.96,1.00]}} & \shortstack{1.00\\{\scriptsize [1.00,1.00]}} & \shortstack{1.00\\{\scriptsize [1.00,1.00]}} & \shortstack{1.00\\{\scriptsize [1.00,1.00]}} & \shortstack{1.00\\{\scriptsize [1.00,1.00]}} \\
\passkrule
 & \setuplbl{Cliff-keep} & \shortstack{0.73\\{\scriptsize [0.56,0.88]}} & \shortstack{0.91\\{\scriptsize [0.80,0.99]}} & \shortstack{0.99\\{\scriptsize [0.96,1.00]}} & \shortstack{1.00\\{\scriptsize [1.00,1.00]}} & \shortstack{1.00\\{\scriptsize [1.00,1.00]}} & \shortstack{1.00\\{\scriptsize [1.00,1.00]}} & \shortstack{1.00\\{\scriptsize [1.00,1.00]}} \\
\passkruledelta
 & \cellcolor{rowgray}\setuplbl{$\Delta$} & \cellcolor{rowgray}\deltapad\shortstack{-0.01\\{\scriptsize [-0.12,0.09]}} & \cellcolor{rowgray}\shortstack{-0.01\\{\scriptsize [-0.08,0.07]}} & \cellcolor{rowgray}\shortstack{0.00\\{\scriptsize [-0.03,0.02]}} & \cellcolor{rowgray}\shortstack{0.00\\{\scriptsize [0.00,0.00]}} & \cellcolor{rowgray}\shortstack{0.00\\{\scriptsize [0.00,0.00]}} & \cellcolor{rowgray}\shortstack{0.00\\{\scriptsize [0.00,0.00]}} & \cellcolor{rowgray}\shortstack{0.00\\{\scriptsize [0.00,0.00]}} \\
\midrule
\multirow{3}{*}{\shortstack[l]{\texttt{Qwen3-4B}\\{\footnotesize $(n = 2)$}}} & \setuplbl{Cliff-del} & \shortstack{0.72\\{\scriptsize [0.48,0.94]}} & \shortstack{0.89\\{\scriptsize [0.73,1.00]}} & \shortstack{0.98\\{\scriptsize [0.92,1.00]}} & \shortstack{1.00\\{\scriptsize [0.99,1.00]}} & \shortstack{1.00\\{\scriptsize [1.00,1.00]}} & \shortstack{1.00\\{\scriptsize [1.00,1.00]}} & \shortstack{1.00\\{\scriptsize [1.00,1.00]}} \\
\passkrule
 & \setuplbl{Cliff-keep} & \shortstack{0.50\\{\scriptsize [0.06,0.94]}} & \shortstack{0.59\\{\scriptsize [0.12,1.00]}} & \shortstack{0.67\\{\scriptsize [0.23,1.00]}} & \shortstack{0.78\\{\scriptsize [0.40,1.00]}} & \shortstack{0.92\\{\scriptsize [0.63,1.00]}} & \shortstack{0.99\\{\scriptsize [0.87,1.00]}} & \shortstack{1.00\\{\scriptsize [1.00,1.00]}} \\
\passkruledelta
 & \cellcolor{rowgray}\setuplbl{$\Delta$} & \cellcolor{rowgray}\deltapad\shortstack{0.22\\{\scriptsize [-0.05,0.50]}} & \cellcolor{rowgray}\shortstack{0.30\\{\scriptsize [-0.01,0.68]}} & \cellcolor{rowgray}\shortstack{0.31\\{\scriptsize [0.00,0.73]}} & \cellcolor{rowgray}\shortstack{0.22\\{\scriptsize [0.00,0.60]}} & \cellcolor{rowgray}\shortstack{0.08\\{\scriptsize [0.00,0.38]}} & \cellcolor{rowgray}\shortstack{0.01\\{\scriptsize [0.00,0.13]}} & \cellcolor{rowgray}\shortstack{0.00\\{\scriptsize [0.00,0.00]}} \\
\midrule
\multirow{3}{*}{\shortstack[l]{\texttt{Qwen3-0.6B}\\{\footnotesize $(n = 5)$}}} & \setuplbl{Cliff-del} & \shortstack{0.67\\{\scriptsize [0.47,0.85]}} & \shortstack{0.85\\{\scriptsize [0.69,0.98]}} & \shortstack{0.96\\{\scriptsize [0.88,1.00]}} & \shortstack{1.00\\{\scriptsize [0.98,1.00]}} & \shortstack{1.00\\{\scriptsize [1.00,1.00]}} & \shortstack{1.00\\{\scriptsize [1.00,1.00]}} & \shortstack{1.00\\{\scriptsize [1.00,1.00]}} \\
\passkrule
 & \setuplbl{Cliff-keep} & \shortstack{0.63\\{\scriptsize [0.41,0.83]}} & \shortstack{0.81\\{\scriptsize [0.62,0.97]}} & \shortstack{0.93\\{\scriptsize [0.82,1.00]}} & \shortstack{0.99\\{\scriptsize [0.96,1.00]}} & \shortstack{1.00\\{\scriptsize [1.00,1.00]}} & \shortstack{1.00\\{\scriptsize [1.00,1.00]}} & \shortstack{1.00\\{\scriptsize [1.00,1.00]}} \\
\passkruledelta
 & \cellcolor{rowgray}\setuplbl{$\Delta$} & \cellcolor{rowgray}\deltapad\shortstack{0.04\\{\scriptsize [-0.03,0.11]}} & \cellcolor{rowgray}\shortstack{0.04\\{\scriptsize [-0.02,0.12]}} & \cellcolor{rowgray}\shortstack{0.03\\{\scriptsize [-0.02,0.10]}} & \cellcolor{rowgray}\shortstack{0.01\\{\scriptsize [-0.01,0.04]}} & \cellcolor{rowgray}\shortstack{0.00\\{\scriptsize [0.00,0.00]}} & \cellcolor{rowgray}\shortstack{0.00\\{\scriptsize [0.00,0.00]}} & \cellcolor{rowgray}\shortstack{0.00\\{\scriptsize [0.00,0.00]}} \\
\midrule
\multirow{3}{*}{\shortstack[l]{\texttt{Llama-3.1-8B}\\{\footnotesize $(n = 10)$}}} & \setuplbl{Cliff-del} & \shortstack{0.71\\{\scriptsize [0.57,0.85]}} & \shortstack{0.87\\{\scriptsize [0.75,0.96]}} & \shortstack{0.96\\{\scriptsize [0.88,1.00]}} & \shortstack{0.99\\{\scriptsize [0.96,1.00]}} & \shortstack{1.00\\{\scriptsize [0.99,1.00]}} & \shortstack{1.00\\{\scriptsize [1.00,1.00]}} & \shortstack{1.00\\{\scriptsize [1.00,1.00]}} \\
\passkrule
 & \setuplbl{Cliff-keep} & \shortstack{0.51\\{\scriptsize [0.38,0.64]}} & \shortstack{0.73\\{\scriptsize [0.58,0.85]}} & \shortstack{0.89\\{\scriptsize [0.78,0.96]}} & \shortstack{0.97\\{\scriptsize [0.92,1.00]}} & \shortstack{1.00\\{\scriptsize [0.99,1.00]}} & \shortstack{1.00\\{\scriptsize [1.00,1.00]}} & \shortstack{1.00\\{\scriptsize [1.00,1.00]}} \\
\passkruledelta
 & \cellcolor{rowgray}\setuplbl{$\Delta$} & \cellcolor{rowgray}\deltapad\shortstack{0.20\\{\scriptsize [0.09,0.31]}} & \cellcolor{rowgray}\shortstack{0.15\\{\scriptsize [0.05,0.26]}} & \cellcolor{rowgray}\shortstack{0.07\\{\scriptsize [0.00,0.17]}} & \cellcolor{rowgray}\shortstack{0.02\\{\scriptsize [-0.01,0.07]}} & \cellcolor{rowgray}\shortstack{0.00\\{\scriptsize [0.00,0.01]}} & \cellcolor{rowgray}\shortstack{0.00\\{\scriptsize [0.00,0.00]}} & \cellcolor{rowgray}\shortstack{0.00\\{\scriptsize [0.00,0.00]}} \\
\midrule
\multirow{3}{*}{\shortstack[l]{\texttt{Llama-3.2-3B}\\{\footnotesize $(n = 6)$}}} & \setuplbl{Cliff-del} & \shortstack{0.54\\{\scriptsize [0.38,0.73]}} & \shortstack{0.75\\{\scriptsize [0.59,0.89]}} & \shortstack{0.91\\{\scriptsize [0.81,0.98]}} & \shortstack{0.99\\{\scriptsize [0.95,1.00]}} & \shortstack{1.00\\{\scriptsize [0.99,1.00]}} & \shortstack{1.00\\{\scriptsize [1.00,1.00]}} & \shortstack{1.00\\{\scriptsize [1.00,1.00]}} \\
\passkrule
 & \setuplbl{Cliff-keep} & \shortstack{0.48\\{\scriptsize [0.32,0.64]}} & \shortstack{0.69\\{\scriptsize [0.52,0.84]}} & \shortstack{0.87\\{\scriptsize [0.74,0.96]}} & \shortstack{0.97\\{\scriptsize [0.90,1.00]}} & \shortstack{1.00\\{\scriptsize [0.98,1.00]}} & \shortstack{1.00\\{\scriptsize [1.00,1.00]}} & \shortstack{1.00\\{\scriptsize [1.00,1.00]}} \\
\passkruledelta
 & \cellcolor{rowgray}\setuplbl{$\Delta$} & \cellcolor{rowgray}\deltapad\shortstack{0.07\\{\scriptsize [-0.13,0.27]}} & \cellcolor{rowgray}\shortstack{0.06\\{\scriptsize [-0.14,0.25]}} & \cellcolor{rowgray}\shortstack{0.04\\{\scriptsize [-0.11,0.19]}} & \cellcolor{rowgray}\shortstack{0.02\\{\scriptsize [-0.04,0.09]}} & \cellcolor{rowgray}\shortstack{0.00\\{\scriptsize [0.00,0.02]}} & \cellcolor{rowgray}\shortstack{0.00\\{\scriptsize [0.00,0.00]}} & \cellcolor{rowgray}\shortstack{0.00\\{\scriptsize [0.00,0.00]}} \\
\midrule
\multirow{3}{*}{\shortstack[l]{\texttt{Llama-3.2-1B}\\{\footnotesize $(n = 7)$}}} & \setuplbl{Cliff-del} & \shortstack{0.67\\{\scriptsize [0.57,0.77]}} & \shortstack{0.88\\{\scriptsize [0.80,0.94]}} & \shortstack{0.98\\{\scriptsize [0.95,1.00]}} & \shortstack{1.00\\{\scriptsize [0.99,1.00]}} & \shortstack{1.00\\{\scriptsize [1.00,1.00]}} & \shortstack{1.00\\{\scriptsize [1.00,1.00]}} & \shortstack{1.00\\{\scriptsize [1.00,1.00]}} \\
\passkrule
 & \setuplbl{Cliff-keep} & \shortstack{0.42\\{\scriptsize [0.27,0.55]}} & \shortstack{0.63\\{\scriptsize [0.42,0.79]}} & \shortstack{0.81\\{\scriptsize [0.55,0.95]}} & \shortstack{0.88\\{\scriptsize [0.64,1.00]}} & \shortstack{0.92\\{\scriptsize [0.71,1.00]}} & \shortstack{0.97\\{\scriptsize [0.79,1.00]}} & \shortstack{1.00\\{\scriptsize [0.86,1.00]}} \\
\passkruledelta
 & \cellcolor{rowgray}\setuplbl{$\Delta$} & \cellcolor{rowgray}\deltapad\shortstack{0.25\\{\scriptsize [0.12,0.38]}} & \cellcolor{rowgray}\shortstack{0.25\\{\scriptsize [0.09,0.44]}} & \cellcolor{rowgray}\shortstack{0.17\\{\scriptsize [0.03,0.42]}} & \cellcolor{rowgray}\shortstack{0.12\\{\scriptsize [0.00,0.36]}} & \cellcolor{rowgray}\shortstack{0.08\\{\scriptsize [0.00,0.29]}} & \cellcolor{rowgray}\shortstack{0.04\\{\scriptsize [0.00,0.21]}} & \cellcolor{rowgray}\shortstack{0.00\\{\scriptsize [0.00,0.14]}} \\
\midrule
\multirow{3}{*}{\shortstack[l]{\texttt{Gemma-3-4B}\\{\footnotesize $(n = 7)$}}} & \setuplbl{Cliff-del} & \shortstack{0.70\\{\scriptsize [0.60,0.80]}} & \shortstack{0.90\\{\scriptsize [0.83,0.95]}} & \shortstack{0.99\\{\scriptsize [0.96,1.00]}} & \shortstack{1.00\\{\scriptsize [1.00,1.00]}} & \shortstack{1.00\\{\scriptsize [1.00,1.00]}} & \shortstack{1.00\\{\scriptsize [1.00,1.00]}} & \shortstack{1.00\\{\scriptsize [1.00,1.00]}} \\
\passkrule
 & \setuplbl{Cliff-keep} & \shortstack{0.57\\{\scriptsize [0.45,0.71]}} & \shortstack{0.79\\{\scriptsize [0.68,0.89]}} & \shortstack{0.94\\{\scriptsize [0.89,0.98]}} & \shortstack{1.00\\{\scriptsize [0.98,1.00]}} & \shortstack{1.00\\{\scriptsize [1.00,1.00]}} & \shortstack{1.00\\{\scriptsize [1.00,1.00]}} & \shortstack{1.00\\{\scriptsize [1.00,1.00]}} \\
\passkruledelta
 & \cellcolor{rowgray}\setuplbl{$\Delta$} & \cellcolor{rowgray}\deltapad\shortstack{0.13\\{\scriptsize [0.03,0.24]}} & \cellcolor{rowgray}\shortstack{0.11\\{\scriptsize [0.03,0.20]}} & \cellcolor{rowgray}\shortstack{0.04\\{\scriptsize [0.01,0.09]}} & \cellcolor{rowgray}\shortstack{0.00\\{\scriptsize [0.00,0.02]}} & \cellcolor{rowgray}\shortstack{0.00\\{\scriptsize [0.00,0.00]}} & \cellcolor{rowgray}\shortstack{0.00\\{\scriptsize [0.00,0.00]}} & \cellcolor{rowgray}\shortstack{0.00\\{\scriptsize [0.00,0.00]}} \\
\bottomrule
\end{tabular}
\end{adjustbox}
\end{table}

\begin{table}[H]
\centering
\caption{\textbf{MATH500:} Cliff-del and Cliff-keep pass@$k$ on correct traces. Notations follow \Cref{tab:passk_incorrect_firstcliff_gsm1k}.}
\label{tab:passk_correct_firstcliff_math500}
\small
\setlength{\tabcolsep}{1.5pt}
\renewcommand{\arraystretch}{1.05}
\begin{adjustbox}{max width=\linewidth}
\begin{tabular}{l>{\centering\arraybackslash}m{0.65in}*{7}{c}}
\toprule
\textbf{Model} & \textbf{Setup} & {\boldmath$k=1$} & {\boldmath$k=2$} & {\boldmath$k=4$} & {\boldmath$k=8$} & {\boldmath$k=16$} & {\boldmath$k=32$} & {\boldmath$k=64$} \\
\midrule
\multirow{3}{*}{\shortstack[l]{\texttt{Qwen3-8B}\\{\footnotesize $(n = 13)$}}} & \setuplbl{Cliff-del} & \shortstack{0.68\\{\scriptsize [0.58,0.77]}} & \shortstack{0.87\\{\scriptsize [0.79,0.93]}} & \shortstack{0.97\\{\scriptsize [0.92,0.99]}} & \shortstack{1.00\\{\scriptsize [0.98,1.00]}} & \shortstack{1.00\\{\scriptsize [1.00,1.00]}} & \shortstack{1.00\\{\scriptsize [1.00,1.00]}} & \shortstack{1.00\\{\scriptsize [1.00,1.00]}} \\
\passkrule
 & \setuplbl{Cliff-keep} & \shortstack{0.63\\{\scriptsize [0.54,0.71]}} & \shortstack{0.85\\{\scriptsize [0.77,0.90]}} & \shortstack{0.97\\{\scriptsize [0.92,0.99]}} & \shortstack{1.00\\{\scriptsize [0.98,1.00]}} & \shortstack{1.00\\{\scriptsize [1.00,1.00]}} & \shortstack{1.00\\{\scriptsize [1.00,1.00]}} & \shortstack{1.00\\{\scriptsize [1.00,1.00]}} \\
\passkruledelta
 & \cellcolor{rowgray}\setuplbl{$\Delta$} & \cellcolor{rowgray}\deltapad\shortstack{0.05\\{\scriptsize [-0.03,0.14]}} & \cellcolor{rowgray}\shortstack{0.03\\{\scriptsize [-0.05,0.11]}} & \cellcolor{rowgray}\shortstack{0.01\\{\scriptsize [-0.04,0.05]}} & \cellcolor{rowgray}\shortstack{0.00\\{\scriptsize [-0.02,0.01]}} & \cellcolor{rowgray}\shortstack{0.00\\{\scriptsize [0.00,0.00]}} & \cellcolor{rowgray}\shortstack{0.00\\{\scriptsize [0.00,0.00]}} & \cellcolor{rowgray}\shortstack{0.00\\{\scriptsize [0.00,0.00]}} \\
\midrule
\multirow{3}{*}{\shortstack[l]{\texttt{Qwen3-4B}\\{\footnotesize $(n = 9)$}}} & \setuplbl{Cliff-del} & \shortstack{0.62\\{\scriptsize [0.48,0.75]}} & \shortstack{0.82\\{\scriptsize [0.69,0.92]}} & \shortstack{0.94\\{\scriptsize [0.87,0.99]}} & \shortstack{0.99\\{\scriptsize [0.97,1.00]}} & \shortstack{1.00\\{\scriptsize [1.00,1.00]}} & \shortstack{1.00\\{\scriptsize [1.00,1.00]}} & \shortstack{1.00\\{\scriptsize [1.00,1.00]}} \\
\passkrule
 & \setuplbl{Cliff-keep} & \shortstack{0.61\\{\scriptsize [0.48,0.74]}} & \shortstack{0.81\\{\scriptsize [0.69,0.92]}} & \shortstack{0.94\\{\scriptsize [0.86,0.99]}} & \shortstack{0.99\\{\scriptsize [0.96,1.00]}} & \shortstack{1.00\\{\scriptsize [1.00,1.00]}} & \shortstack{1.00\\{\scriptsize [1.00,1.00]}} & \shortstack{1.00\\{\scriptsize [1.00,1.00]}} \\
\passkruledelta
 & \cellcolor{rowgray}\setuplbl{$\Delta$} & \cellcolor{rowgray}\deltapad\shortstack{0.01\\{\scriptsize [-0.08,0.12]}} & \cellcolor{rowgray}\shortstack{0.00\\{\scriptsize [-0.06,0.07]}} & \cellcolor{rowgray}\shortstack{0.00\\{\scriptsize [-0.05,0.05]}} & \cellcolor{rowgray}\shortstack{0.00\\{\scriptsize [-0.02,0.02]}} & \cellcolor{rowgray}\shortstack{0.00\\{\scriptsize [0.00,0.00]}} & \cellcolor{rowgray}\shortstack{0.00\\{\scriptsize [0.00,0.00]}} & \cellcolor{rowgray}\shortstack{0.00\\{\scriptsize [0.00,0.00]}} \\
\midrule
\multirow{3}{*}{\shortstack[l]{\texttt{Qwen3-0.6B}\\{\footnotesize $(n = 10)$}}} & \setuplbl{Cliff-del} & \shortstack{0.50\\{\scriptsize [0.38,0.63]}} & \shortstack{0.72\\{\scriptsize [0.59,0.83]}} & \shortstack{0.89\\{\scriptsize [0.80,0.96]}} & \shortstack{0.98\\{\scriptsize [0.93,1.00]}} & \shortstack{1.00\\{\scriptsize [0.99,1.00]}} & \shortstack{1.00\\{\scriptsize [1.00,1.00]}} & \shortstack{1.00\\{\scriptsize [1.00,1.00]}} \\
\passkrule
 & \setuplbl{Cliff-keep} & \shortstack{0.45\\{\scriptsize [0.32,0.59]}} & \shortstack{0.65\\{\scriptsize [0.50,0.79]}} & \shortstack{0.84\\{\scriptsize [0.71,0.93]}} & \shortstack{0.95\\{\scriptsize [0.86,0.99]}} & \shortstack{0.99\\{\scriptsize [0.96,1.00]}} & \shortstack{1.00\\{\scriptsize [0.99,1.00]}} & \shortstack{1.00\\{\scriptsize [1.00,1.00]}} \\
\passkruledelta
 & \cellcolor{rowgray}\setuplbl{$\Delta$} & \cellcolor{rowgray}\deltapad\shortstack{0.05\\{\scriptsize [-0.04,0.15]}} & \cellcolor{rowgray}\shortstack{0.06\\{\scriptsize [-0.03,0.17]}} & \cellcolor{rowgray}\shortstack{0.06\\{\scriptsize [-0.01,0.14]}} & \cellcolor{rowgray}\shortstack{0.03\\{\scriptsize [-0.01,0.09]}} & \cellcolor{rowgray}\shortstack{0.01\\{\scriptsize [0.00,0.04]}} & \cellcolor{rowgray}\shortstack{0.00\\{\scriptsize [0.00,0.01]}} & \cellcolor{rowgray}\shortstack{0.00\\{\scriptsize [0.00,0.00]}} \\
\midrule
\multirow{3}{*}{\shortstack[l]{\texttt{Llama-3.1-8B}\\{\footnotesize $(n = 15)$}}} & \setuplbl{Cliff-del} & \shortstack{0.64\\{\scriptsize [0.52,0.76]}} & \shortstack{0.82\\{\scriptsize [0.73,0.91]}} & \shortstack{0.94\\{\scriptsize [0.89,0.98]}} & \shortstack{0.99\\{\scriptsize [0.97,1.00]}} & \shortstack{1.00\\{\scriptsize [1.00,1.00]}} & \shortstack{1.00\\{\scriptsize [1.00,1.00]}} & \shortstack{1.00\\{\scriptsize [1.00,1.00]}} \\
\passkrule
 & \setuplbl{Cliff-keep} & \shortstack{0.48\\{\scriptsize [0.34,0.62]}} & \shortstack{0.66\\{\scriptsize [0.50,0.80]}} & \shortstack{0.81\\{\scriptsize [0.65,0.93]}} & \shortstack{0.89\\{\scriptsize [0.76,0.98]}} & \shortstack{0.95\\{\scriptsize [0.84,1.00]}} & \shortstack{0.98\\{\scriptsize [0.90,1.00]}} & \shortstack{1.00\\{\scriptsize [0.93,1.00]}} \\
\passkruledelta
 & \cellcolor{rowgray}\setuplbl{$\Delta$} & \cellcolor{rowgray}\deltapad\shortstack{0.16\\{\scriptsize [0.03,0.30]}} & \cellcolor{rowgray}\shortstack{0.16\\{\scriptsize [0.03,0.31]}} & \cellcolor{rowgray}\shortstack{0.14\\{\scriptsize [0.02,0.28]}} & \cellcolor{rowgray}\shortstack{0.10\\{\scriptsize [0.01,0.22]}} & \cellcolor{rowgray}\shortstack{0.05\\{\scriptsize [0.00,0.16]}} & \cellcolor{rowgray}\shortstack{0.02\\{\scriptsize [0.00,0.10]}} & \cellcolor{rowgray}\shortstack{0.00\\{\scriptsize [0.00,0.07]}} \\
\midrule
\multirow{3}{*}{\shortstack[l]{\texttt{Llama-3.2-3B}\\{\footnotesize $(n = 8)$}}} & \setuplbl{Cliff-del} & \shortstack{0.61\\{\scriptsize [0.43,0.75]}} & \shortstack{0.80\\{\scriptsize [0.60,0.93]}} & \shortstack{0.90\\{\scriptsize [0.72,0.99]}} & \shortstack{0.95\\{\scriptsize [0.81,1.00]}} & \shortstack{0.98\\{\scriptsize [0.90,1.00]}} & \shortstack{1.00\\{\scriptsize [0.97,1.00]}} & \shortstack{1.00\\{\scriptsize [1.00,1.00]}} \\
\passkrule
 & \setuplbl{Cliff-keep} & \shortstack{0.50\\{\scriptsize [0.32,0.66]}} & \shortstack{0.70\\{\scriptsize [0.46,0.87]}} & \shortstack{0.82\\{\scriptsize [0.57,0.98]}} & \shortstack{0.87\\{\scriptsize [0.62,1.00]}} & \shortstack{0.88\\{\scriptsize [0.63,1.00]}} & \shortstack{0.88\\{\scriptsize [0.63,1.00]}} & \shortstack{0.88\\{\scriptsize [0.63,1.00]}} \\
\passkruledelta
 & \cellcolor{rowgray}\setuplbl{$\Delta$} & \cellcolor{rowgray}\deltapad\shortstack{0.10\\{\scriptsize [0.00,0.21]}} & \cellcolor{rowgray}\shortstack{0.10\\{\scriptsize [0.01,0.20]}} & \cellcolor{rowgray}\shortstack{0.08\\{\scriptsize [0.01,0.18]}} & \cellcolor{rowgray}\shortstack{0.08\\{\scriptsize [0.00,0.23]}} & \cellcolor{rowgray}\shortstack{0.11\\{\scriptsize [0.00,0.32]}} & \cellcolor{rowgray}\shortstack{0.12\\{\scriptsize [0.00,0.37]}} & \cellcolor{rowgray}\shortstack{0.13\\{\scriptsize [0.00,0.38]}} \\
\midrule
\multirow{3}{*}{\shortstack[l]{\texttt{Llama-3.2-1B}\\{\footnotesize $(n = 9)$}}} & \setuplbl{Cliff-del} & \shortstack{0.73\\{\scriptsize [0.59,0.86]}} & \shortstack{0.89\\{\scriptsize [0.78,0.96]}} & \shortstack{0.97\\{\scriptsize [0.91,1.00]}} & \shortstack{1.00\\{\scriptsize [0.98,1.00]}} & \shortstack{1.00\\{\scriptsize [1.00,1.00]}} & \shortstack{1.00\\{\scriptsize [1.00,1.00]}} & \shortstack{1.00\\{\scriptsize [1.00,1.00]}} \\
\passkrule
 & \setuplbl{Cliff-keep} & \shortstack{0.56\\{\scriptsize [0.44,0.67]}} & \shortstack{0.78\\{\scriptsize [0.66,0.88]}} & \shortstack{0.93\\{\scriptsize [0.85,0.98]}} & \shortstack{0.99\\{\scriptsize [0.96,1.00]}} & \shortstack{1.00\\{\scriptsize [1.00,1.00]}} & \shortstack{1.00\\{\scriptsize [1.00,1.00]}} & \shortstack{1.00\\{\scriptsize [1.00,1.00]}} \\
\passkruledelta
 & \cellcolor{rowgray}\setuplbl{$\Delta$} & \cellcolor{rowgray}\deltapad\shortstack{0.17\\{\scriptsize [0.06,0.30]}} & \cellcolor{rowgray}\shortstack{0.11\\{\scriptsize [0.04,0.20]}} & \cellcolor{rowgray}\shortstack{0.04\\{\scriptsize [0.01,0.09]}} & \cellcolor{rowgray}\shortstack{0.01\\{\scriptsize [-0.01,0.03]}} & \cellcolor{rowgray}\shortstack{0.00\\{\scriptsize [0.00,0.00]}} & \cellcolor{rowgray}\shortstack{0.00\\{\scriptsize [0.00,0.00]}} & \cellcolor{rowgray}\shortstack{0.00\\{\scriptsize [0.00,0.00]}} \\
\midrule
\multirow{3}{*}{\shortstack[l]{\texttt{Gemma-3-4B}\\{\footnotesize $(n = 7)$}}} & \setuplbl{Cliff-del} & \shortstack{0.80\\{\scriptsize [0.70,0.87]}} & \shortstack{0.95\\{\scriptsize [0.90,0.98]}} & \shortstack{1.00\\{\scriptsize [0.98,1.00]}} & \shortstack{1.00\\{\scriptsize [1.00,1.00]}} & \shortstack{1.00\\{\scriptsize [1.00,1.00]}} & \shortstack{1.00\\{\scriptsize [1.00,1.00]}} & \shortstack{1.00\\{\scriptsize [1.00,1.00]}} \\
\passkrule
 & \setuplbl{Cliff-keep} & \shortstack{0.66\\{\scriptsize [0.54,0.78]}} & \shortstack{0.86\\{\scriptsize [0.77,0.94]}} & \shortstack{0.97\\{\scriptsize [0.94,0.99]}} & \shortstack{1.00\\{\scriptsize [0.99,1.00]}} & \shortstack{1.00\\{\scriptsize [1.00,1.00]}} & \shortstack{1.00\\{\scriptsize [1.00,1.00]}} & \shortstack{1.00\\{\scriptsize [1.00,1.00]}} \\
\passkruledelta
 & \cellcolor{rowgray}\setuplbl{$\Delta$} & \cellcolor{rowgray}\deltapad\shortstack{0.14\\{\scriptsize [0.03,0.25]}} & \cellcolor{rowgray}\shortstack{0.09\\{\scriptsize [0.02,0.17]}} & \cellcolor{rowgray}\shortstack{0.02\\{\scriptsize [0.00,0.06]}} & \cellcolor{rowgray}\shortstack{0.00\\{\scriptsize [0.00,0.01]}} & \cellcolor{rowgray}\shortstack{0.00\\{\scriptsize [0.00,0.00]}} & \cellcolor{rowgray}\shortstack{0.00\\{\scriptsize [0.00,0.00]}} & \cellcolor{rowgray}\shortstack{0.00\\{\scriptsize [0.00,0.00]}} \\
\bottomrule
\end{tabular}
\end{adjustbox}
\end{table}

\begin{table}[H]
\centering
\caption{\textbf{AIME 2025:} Cliff-del and Cliff-keep pass@$k$ on correct traces. \texttt{n/e} indicates that the CI is not estimated since $n<2$. \texttt{--} indicates that no traces were included in the analysis. Notations follow \Cref{tab:passk_incorrect_firstcliff_gsm1k}.}
\label{tab:passk_correct_firstcliff_aime25}
\small
\setlength{\tabcolsep}{1.5pt}
\renewcommand{\arraystretch}{1.05}
\begin{adjustbox}{max width=\linewidth}
\begin{tabular}{l>{\centering\arraybackslash}m{0.65in}*{7}{c}}
\toprule
\textbf{Model} & \textbf{Setup} & {\boldmath$k=1$} & {\boldmath$k=2$} & {\boldmath$k=4$} & {\boldmath$k=8$} & {\boldmath$k=16$} & {\boldmath$k=32$} & {\boldmath$k=64$} \\
\midrule
\multirow{3}{*}{\shortstack[l]{\texttt{Qwen3-8B}\\{\footnotesize $(n = 2)$}}} & \setuplbl{Cliff-del} & \shortstack{0.61\\{\scriptsize [0.32,0.88]}} & \shortstack{0.80\\{\scriptsize [0.54,0.99]}} & \shortstack{0.93\\{\scriptsize [0.79,1.00]}} & \shortstack{0.99\\{\scriptsize [0.96,1.00]}} & \shortstack{1.00\\{\scriptsize [1.00,1.00]}} & \shortstack{1.00\\{\scriptsize [1.00,1.00]}} & \shortstack{1.00\\{\scriptsize [1.00,1.00]}} \\
\passkrule
 & \setuplbl{Cliff-keep} & \shortstack{0.62\\{\scriptsize [0.38,0.84]}} & \shortstack{0.82\\{\scriptsize [0.62,0.98]}} & \shortstack{0.95\\{\scriptsize [0.86,1.00]}} & \shortstack{1.00\\{\scriptsize [0.98,1.00]}} & \shortstack{1.00\\{\scriptsize [1.00,1.00]}} & \shortstack{1.00\\{\scriptsize [1.00,1.00]}} & \shortstack{1.00\\{\scriptsize [1.00,1.00]}} \\
\passkruledelta
 & \cellcolor{rowgray}\setuplbl{$\Delta$} & \cellcolor{rowgray}\deltapad\shortstack{-0.01\\{\scriptsize [-0.15,0.12]}} & \cellcolor{rowgray}\shortstack{-0.03\\{\scriptsize [-0.17,0.07]}} & \cellcolor{rowgray}\shortstack{-0.03\\{\scriptsize [-0.13,0.04]}} & \cellcolor{rowgray}\shortstack{-0.01\\{\scriptsize [-0.04,0.01]}} & \cellcolor{rowgray}\shortstack{0.00\\{\scriptsize [0.00,0.00]}} & \cellcolor{rowgray}\shortstack{0.00\\{\scriptsize [0.00,0.00]}} & \cellcolor{rowgray}\shortstack{0.00\\{\scriptsize [0.00,0.00]}} \\
\midrule
\multirow{3}{*}{\shortstack[l]{\texttt{Qwen3-4B}\\{\footnotesize $(n = 1)$}}} & \setuplbl{Cliff-del} & \shortstack{0.63\\{\scriptsize [n/e]}} & \shortstack{0.86\\{\scriptsize [n/e]}} & \shortstack{0.98\\{\scriptsize [n/e]}} & \shortstack{1.00\\{\scriptsize [n/e]}} & \shortstack{1.00\\{\scriptsize [n/e]}} & \shortstack{1.00\\{\scriptsize [n/e]}} & \shortstack{1.00\\{\scriptsize [n/e]}} \\
\passkrule
 & \setuplbl{Cliff-keep} & \shortstack{0.25\\{\scriptsize [n/e]}} & \shortstack{0.44\\{\scriptsize [n/e]}} & \shortstack{0.69\\{\scriptsize [n/e]}} & \shortstack{0.92\\{\scriptsize [n/e]}} & \shortstack{1.00\\{\scriptsize [n/e]}} & \shortstack{1.00\\{\scriptsize [n/e]}} & \shortstack{1.00\\{\scriptsize [n/e]}} \\
\passkruledelta
 & \cellcolor{rowgray}\setuplbl{$\Delta$} & \cellcolor{rowgray}\deltapad\shortstack{0.38\\{\scriptsize [n/e]}} & \cellcolor{rowgray}\shortstack{0.42\\{\scriptsize [n/e]}} & \cellcolor{rowgray}\shortstack{0.29\\{\scriptsize [n/e]}} & \cellcolor{rowgray}\shortstack{0.09\\{\scriptsize [n/e]}} & \cellcolor{rowgray}\shortstack{0.01\\{\scriptsize [n/e]}} & \cellcolor{rowgray}\shortstack{0.00\\{\scriptsize [n/e]}} & \cellcolor{rowgray}\shortstack{0.00\\{\scriptsize [n/e]}} \\
\midrule
\multirow{3}{*}{\shortstack[l]{\texttt{Qwen3-0.6B}\\{\footnotesize $(n = 1)$}}} & \setuplbl{Cliff-del} & \shortstack{0.48\\{\scriptsize [n/e]}} & \shortstack{0.74\\{\scriptsize [n/e]}} & \shortstack{0.94\\{\scriptsize [n/e]}} & \shortstack{1.00\\{\scriptsize [n/e]}} & \shortstack{1.00\\{\scriptsize [n/e]}} & \shortstack{1.00\\{\scriptsize [n/e]}} & \shortstack{1.00\\{\scriptsize [n/e]}} \\
\passkrule
 & \setuplbl{Cliff-keep} & \shortstack{0.44\\{\scriptsize [n/e]}} & \shortstack{0.69\\{\scriptsize [n/e]}} & \shortstack{0.91\\{\scriptsize [n/e]}} & \shortstack{0.99\\{\scriptsize [n/e]}} & \shortstack{1.00\\{\scriptsize [n/e]}} & \shortstack{1.00\\{\scriptsize [n/e]}} & \shortstack{1.00\\{\scriptsize [n/e]}} \\
\passkruledelta
 & \cellcolor{rowgray}\setuplbl{$\Delta$} & \cellcolor{rowgray}\deltapad\shortstack{0.05\\{\scriptsize [n/e]}} & \cellcolor{rowgray}\shortstack{0.05\\{\scriptsize [n/e]}} & \cellcolor{rowgray}\shortstack{0.03\\{\scriptsize [n/e]}} & \cellcolor{rowgray}\shortstack{0.00\\{\scriptsize [n/e]}} & \cellcolor{rowgray}\shortstack{0.00\\{\scriptsize [n/e]}} & \cellcolor{rowgray}\shortstack{0.00\\{\scriptsize [n/e]}} & \cellcolor{rowgray}\shortstack{0.00\\{\scriptsize [n/e]}} \\
\midrule
\multirow{3}{*}{\shortstack[l]{\texttt{Llama-3.1-8B}\\{\footnotesize $(n = 1)$}}} & \setuplbl{Cliff-del} & \shortstack{0.48\\{\scriptsize [n/e]}} & \shortstack{0.74\\{\scriptsize [n/e]}} & \shortstack{0.94\\{\scriptsize [n/e]}} & \shortstack{1.00\\{\scriptsize [n/e]}} & \shortstack{1.00\\{\scriptsize [n/e]}} & \shortstack{1.00\\{\scriptsize [n/e]}} & \shortstack{1.00\\{\scriptsize [n/e]}} \\
\passkrule
 & \setuplbl{Cliff-keep} & \shortstack{0.38\\{\scriptsize [n/e]}} & \shortstack{0.61\\{\scriptsize [n/e]}} & \shortstack{0.86\\{\scriptsize [n/e]}} & \shortstack{0.98\\{\scriptsize [n/e]}} & \shortstack{1.00\\{\scriptsize [n/e]}} & \shortstack{1.00\\{\scriptsize [n/e]}} & \shortstack{1.00\\{\scriptsize [n/e]}} \\
\passkruledelta
 & \cellcolor{rowgray}\setuplbl{$\Delta$} & \cellcolor{rowgray}\deltapad\shortstack{0.11\\{\scriptsize [n/e]}} & \cellcolor{rowgray}\shortstack{0.13\\{\scriptsize [n/e]}} & \cellcolor{rowgray}\shortstack{0.08\\{\scriptsize [n/e]}} & \cellcolor{rowgray}\shortstack{0.01\\{\scriptsize [n/e]}} & \cellcolor{rowgray}\shortstack{0.00\\{\scriptsize [n/e]}} & \cellcolor{rowgray}\shortstack{0.00\\{\scriptsize [n/e]}} & \cellcolor{rowgray}\shortstack{0.00\\{\scriptsize [n/e]}} \\
\midrule
\multirow{3}{*}{\shortstack[l]{\texttt{Llama-3.2-3B}\\{\footnotesize $(n = 0)$}}} & Cliff-del & -- & -- & -- & -- & -- & -- & -- \\
\passkrule
 & Cliff-keep & -- & -- & -- & -- & -- & -- & -- \\
\passkruledelta
 & \cellcolor{rowgray}$\Delta$ & \cellcolor{rowgray}-- & \cellcolor{rowgray}-- & \cellcolor{rowgray}-- & \cellcolor{rowgray}-- & \cellcolor{rowgray}-- & \cellcolor{rowgray}-- & \cellcolor{rowgray}-- \\
\midrule
\multirow{3}{*}{\shortstack[l]{\texttt{Llama-3.2-1B}\\{\footnotesize $(n = 0)$}}} & Cliff-del & -- & -- & -- & -- & -- & -- & -- \\
\passkrule
 & Cliff-keep & -- & -- & -- & -- & -- & -- & -- \\
\passkruledelta
 & \cellcolor{rowgray}$\Delta$ & \cellcolor{rowgray}-- & \cellcolor{rowgray}-- & \cellcolor{rowgray}-- & \cellcolor{rowgray}-- & \cellcolor{rowgray}-- & \cellcolor{rowgray}-- & \cellcolor{rowgray}-- \\
\midrule
\multirow{3}{*}{\shortstack[l]{\texttt{Gemma-3-4B}\\{\footnotesize $(n = 1)$}}} & \setuplbl{Cliff-del} & \shortstack{0.97\\{\scriptsize [n/e]}} & \shortstack{1.00\\{\scriptsize [n/e]}} & \shortstack{1.00\\{\scriptsize [n/e]}} & \shortstack{1.00\\{\scriptsize [n/e]}} & \shortstack{1.00\\{\scriptsize [n/e]}} & \shortstack{1.00\\{\scriptsize [n/e]}} & \shortstack{1.00\\{\scriptsize [n/e]}} \\
\passkrule
 & \setuplbl{Cliff-keep} & \shortstack{0.84\\{\scriptsize [n/e]}} & \shortstack{0.98\\{\scriptsize [n/e]}} & \shortstack{1.00\\{\scriptsize [n/e]}} & \shortstack{1.00\\{\scriptsize [n/e]}} & \shortstack{1.00\\{\scriptsize [n/e]}} & \shortstack{1.00\\{\scriptsize [n/e]}} & \shortstack{1.00\\{\scriptsize [n/e]}} \\
\passkruledelta
 & \cellcolor{rowgray}\setuplbl{$\Delta$} & \cellcolor{rowgray}\deltapad\shortstack{0.13\\{\scriptsize [n/e]}} & \cellcolor{rowgray}\shortstack{0.02\\{\scriptsize [n/e]}} & \cellcolor{rowgray}\shortstack{0.00\\{\scriptsize [n/e]}} & \cellcolor{rowgray}\shortstack{0.00\\{\scriptsize [n/e]}} & \cellcolor{rowgray}\shortstack{0.00\\{\scriptsize [n/e]}} & \cellcolor{rowgray}\shortstack{0.00\\{\scriptsize [n/e]}} & \cellcolor{rowgray}\shortstack{0.00\\{\scriptsize [n/e]}} \\
\bottomrule
\end{tabular}
\end{adjustbox}
\end{table}

\clearpage

\section{Token entropy distribution}
\label[appendix]{app:entropy}

\begin{figure}[H]
  \centering
  \includegraphics[width=0.8\linewidth]{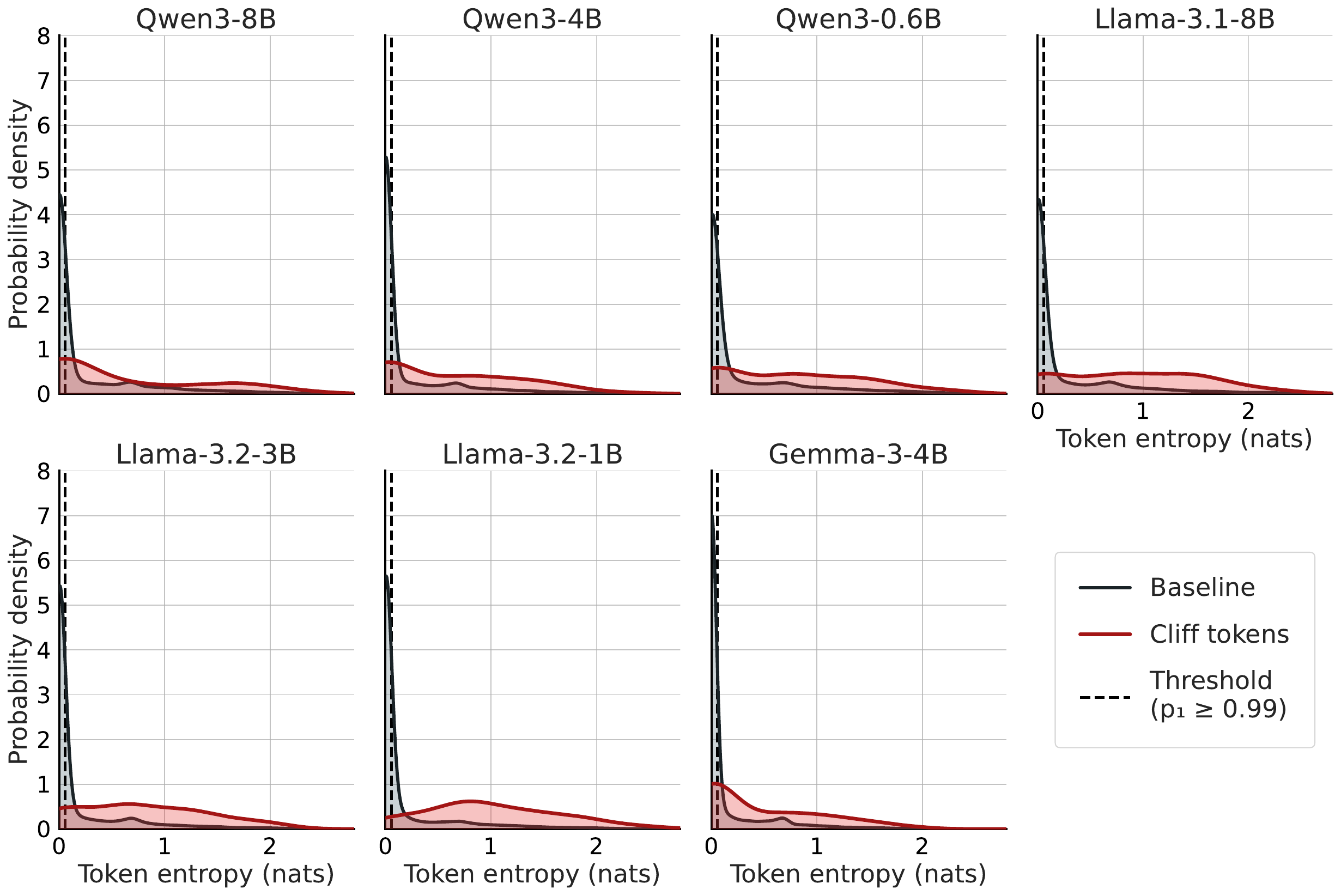}
  \caption{Probability density distributions of token entropy across the seven models. Token entropy is aggregated across the GSM1K, MATH500, and AIME 2025 datasets. The baseline represents the token entropy computed over all generated tokens within the reasoning traces, whereas the cliff token distribution represents the token entropy at the cliff token positions.}
  \label{fig:entropy_density}
\end{figure}

As illustrated in \Cref{fig:entropy_density}, the overall token entropy during mathematical reasoning exhibits a high probability density near zero, peaking at $H \approx 0$. In contrast, the distribution for cliff tokens shows a lower density in this low-entropy regime. Furthermore, the cliff token entropy displays a heavy-tailed distribution. This empirical evidence indicates that models frequently sample cliff tokens under elevated uncertainty.

\section{Greedy token ratios}
\label[appendix]{app:greedy_ratio}

As detailed in \Cref{tab:greedy-ratio}, the greedy-token ratio for cliff token is significantly lower than the overall baseline across all evaluated models. This reduction is largest in the Llama variants, whose cliff-token greedy ratios fall below 0.5. In contrast, a majority of cliff tokens remain greedy for \texttt{Qwen3} family and \texttt{Gemma-3-4B}.

\begin{table}[H]
  \centering
  \caption{Comparison of greedy token sampling rates. Ratios are aggregated across the GSM1K, MATH500, and AIME 2025 datasets. The baseline represents the average greedy token ratio computed over all generated tokens within the reasoning traces.}
  \label{tab:greedy-ratio}
  \begin{tabular}{lcc}
    \toprule
    \textbf{Model} & \textbf{Cliff Tokens} & \textbf{Baseline} \\
    \midrule
    \texttt{Qwen3-8B}              & 0.8229 & 0.9685 \\
    \texttt{Qwen3-4B}              & 0.8056 & 0.9747 \\
    \texttt{Qwen3-0.6B}            & 0.7177 & 0.9650 \\
    \texttt{Llama-3.1-8B}          & 0.4724 & 0.9516 \\
    \texttt{Llama-3.2-3B}          & 0.4211 & 0.9599 \\
    \texttt{Llama-3.2-1B}          & 0.3784 & 0.9636 \\
    \texttt{Gemma-3-4B}            & 0.7500 & 0.9732 \\
    \bottomrule
  \end{tabular}
\end{table}

\section{Robustness of the cliff taxonomy to the entropy threshold}
\label[appendix]{different_entropy_threshold}

To evaluate the robustness of the cliff taxonomy, we conduct an ablation study using alternative token entropy thresholds calculated on binary entropy with $p_1 \in \{0.90, 0.95, 0.999\}$ in \Cref{def_cliff_taxonomy}. As detailed in \Cref{tab:cliff_taxonomy_thresholds}, the qualitative trends reported in \Cref{sec:cliff_taxonomy_across_seven_models} remain consistent across all thresholds. The deterministic cliff proportion is lower than the all-token proportion, while sampled-off cliffs occur at a higher proportion across all models. Furthermore, the scaling behavior within the \texttt{Qwen3} and \texttt{Llama-3} families is preserved across all settings, except for sampled-off cliffs on \texttt{Llama-3.2-3B}. These results indicate that the findings on cliff tokens are not sensitive to the specific boundary of the entropy split.

\begin{table}[H]
  \caption{Distribution of cliff types under alternative entropy splits
  ($p_1 \in \{0.90, 0.95, 0.999\}$). Column definitions follow
  \Cref{tab:cliff_taxonomy_models}.}
  \label{tab:cliff_taxonomy_thresholds}
  \label{tab:cliff_taxonomy_p90}
  \label{tab:cliff_taxonomy_p95}
  \label{tab:cliff_taxonomy_p999}
  \centering
  \small
  \setlength{\tabcolsep}{3pt}
  \renewcommand{\arraystretch}{1.08}

  \begin{tabular}{@{}>{\raggedright\arraybackslash}p{0.23\linewidth}*{6}{>{\centering\arraybackslash}p{\dimexpr(0.77\linewidth-12\tabcolsep)/6\relax}}@{}}
    \toprule
    & \multicolumn{3}{c}{\textbf{Cliff proportion (\%)}}
    & \multicolumn{3}{c}{\textbf{All-token proportion (\%)}} \\
    \cmidrule(lr){2-4}
    \cmidrule(lr){5-7}
    \textbf{Model} & D & U & S & D & U & S \\

    \specialrule{\heavyrulewidth}{0.5ex}{0pt}
    \multicolumn{7}{@{}>{\columncolor{taxonomyblockblue}[0pt][0pt]}l@{}}{
      \rule[-0.75ex]{0pt}{3.0ex}
      \hspace{1em}\textbf{$H_b \approx 0.325$ nats ($p_1 = 0.90$)}
    } \\
    \specialrule{\lightrulewidth}{0pt}{0.5ex}

    \texttt{Qwen3-8B}
      & \cellcolor{black!22.48}56.2
      & \cellcolor{black!10.40}26.0
      & \cellcolor{black!7.08}17.7
      & 78.4 & 18.5 & 3.1 \\
    \texttt{Qwen3-4B}
      & \cellcolor{black!18.88}47.2
      & \cellcolor{black!13.32}33.3
      & \cellcolor{black!7.76}19.4
      & 81.9 & 15.5 & 2.5 \\
    \texttt{Qwen3-0.6B}
      & \cellcolor{black!15.16}37.9
      & \cellcolor{black!13.56}33.9
      & \cellcolor{black!11.28}28.2
      & 75.5 & 21.0 & 3.5 \\

    \specialrule{\lightrulewidth}{0.5ex}{0.5ex}
    \texttt{Llama-3.1-8B}
      & \cellcolor{black!10.08}25.2
      & \cellcolor{black!8.80}22.0
      & \cellcolor{black!20.16}50.4
      & 77.8 & 17.3 & 4.8 \\
    \texttt{Llama-3.2-3B}
      & \cellcolor{black!8.44}21.1
      & \cellcolor{black!8.44}21.1
      & \cellcolor{black!20.64}51.6
      & 82.3 & 13.7 & 3.9 \\
    \texttt{Llama-3.2-1B}
      & \cellcolor{black!5.40}13.5
      & \cellcolor{black!9.72}24.3
      & \cellcolor{black!24.16}60.4
      & 84.3 & 12.1 & 3.6 \\

    \specialrule{\lightrulewidth}{0.5ex}{0.5ex}
    \texttt{Gemma-3-4B}
      & \cellcolor{black!20.48}51.2
      & \cellcolor{black!9.52}23.8
      & \cellcolor{black!7.16}17.9
      & 86.2 & 11.2 & 2.6 \\

    \specialrule{\heavyrulewidth}{0.5ex}{0pt}
    \multicolumn{7}{@{}>{\columncolor{taxonomyblockblue}[0pt][0pt]}l@{}}{
      \rule[-0.75ex]{0pt}{3.0ex}
      \hspace{1em}\textbf{$H_b \approx 0.199$ nats ($p_1 = 0.95$)}
    } \\
    \specialrule{\lightrulewidth}{0pt}{0.5ex}

    \texttt{Qwen3-8B}
      & \cellcolor{black!21.24}53.1
      & \cellcolor{black!11.68}29.2
      & \cellcolor{black!7.08}17.7
      & 75.2 & 21.6 & 3.1 \\
    \texttt{Qwen3-4B}
      & \cellcolor{black!17.76}44.4
      & \cellcolor{black!14.44}36.1
      & \cellcolor{black!7.76}19.4
      & 79.2 & 18.3 & 2.5 \\
    \texttt{Qwen3-0.6B}
      & \cellcolor{black!13.88}34.7
      & \cellcolor{black!14.84}37.1
      & \cellcolor{black!11.28}28.2
      & 71.9 & 24.6 & 3.5 \\

    \specialrule{\lightrulewidth}{0.5ex}{0.5ex}
    \texttt{Llama-3.1-8B}
      & \cellcolor{black!9.76}24.4
      & \cellcolor{black!9.12}22.8
      & \cellcolor{black!20.80}52.0
      & 74.5 & 20.7 & 4.8 \\
    \texttt{Llama-3.2-3B}
      & \cellcolor{black!7.56}18.9
      & \cellcolor{black!9.28}23.2
      & \cellcolor{black!20.64}51.6
      & 79.6 & 16.4 & 4.0 \\
    \texttt{Llama-3.2-1B}
      & \cellcolor{black!4.32}10.8
      & \cellcolor{black!10.80}27.0
      & \cellcolor{black!24.16}60.4
      & 81.6 & 14.7 & 3.6 \\

    \specialrule{\lightrulewidth}{0.5ex}{0.5ex}
    \texttt{Gemma-3-4B}
      & \cellcolor{black!20.48}51.2
      & \cellcolor{black!9.52}23.8
      & \cellcolor{black!7.60}19.0
      & 83.4 & 13.9 & 2.7 \\

    \specialrule{\heavyrulewidth}{0.5ex}{0pt}
    \multicolumn{7}{@{}>{\columncolor{taxonomyblockblue}[0pt][0pt]}l@{}}{
      \rule[-0.75ex]{0pt}{3.0ex}
      \hspace{1em}\textbf{$H_b \approx 0.008$ nats ($p_1 = 0.999$)}
    } \\
    \specialrule{\lightrulewidth}{0pt}{0.5ex}

    \texttt{Qwen3-8B}
      & \cellcolor{black!18.76}46.9
      & \cellcolor{black!14.16}35.4
      & \cellcolor{black!7.08}17.7
      & 61.9 & 35.0 & 3.1 \\
    \texttt{Qwen3-4B}
      & \cellcolor{black!15.56}38.9
      & \cellcolor{black!16.68}41.7
      & \cellcolor{black!7.76}19.4
      & 67.4 & 30.1 & 2.5 \\
    \texttt{Qwen3-0.6B}
      & \cellcolor{black!8.08}20.2
      & \cellcolor{black!20.64}51.6
      & \cellcolor{black!11.28}28.2
      & 52.4 & 44.1 & 3.5 \\

    \specialrule{\lightrulewidth}{0.5ex}{0.5ex}
    \texttt{Llama-3.1-8B}
      & \cellcolor{black!7.24}18.1
      & \cellcolor{black!11.64}29.1
      & \cellcolor{black!21.12}52.8
      & 52.0 & 43.1 & 4.8 \\
    \texttt{Llama-3.2-3B}
      & \cellcolor{black!6.32}15.8
      & \cellcolor{black!10.52}26.3
      & \cellcolor{black!22.72}56.8
      & 61.3 & 34.6 & 4.0 \\
    \texttt{Llama-3.2-1B}
      & \cellcolor{black!3.24}8.1
      & \cellcolor{black!11.88}29.7
      & \cellcolor{black!24.88}62.2
      & 56.7 & 39.7 & 3.6 \\

    \specialrule{\lightrulewidth}{0.5ex}{0.5ex}
    \texttt{Gemma-3-4B}
      & \cellcolor{black!17.16}42.9
      & \cellcolor{black!12.84}32.1
      & \cellcolor{black!9.04}22.6
      & 70.4 & 26.9 & 2.7 \\

    \specialrule{\heavyrulewidth}{0.5ex}{0pt}
  \end{tabular}
\end{table}

\newpage

\section{Counterfactual analysis on sampled-off cliffs}
\label[appendix]{app:counterfactual}

To verify that sampled-off cliffs are induced by stochastic sampling noise rather than an inherent systemic preference toward failure, we conduct a counterfactual analysis on \texttt{Qwen3-8B}. Specifically, we override the sampled-off cliff token with its corresponding greedy token and measure the resulting success probability.

\begin{figure}[H]
  \centering
  \includegraphics[width=0.45\linewidth]{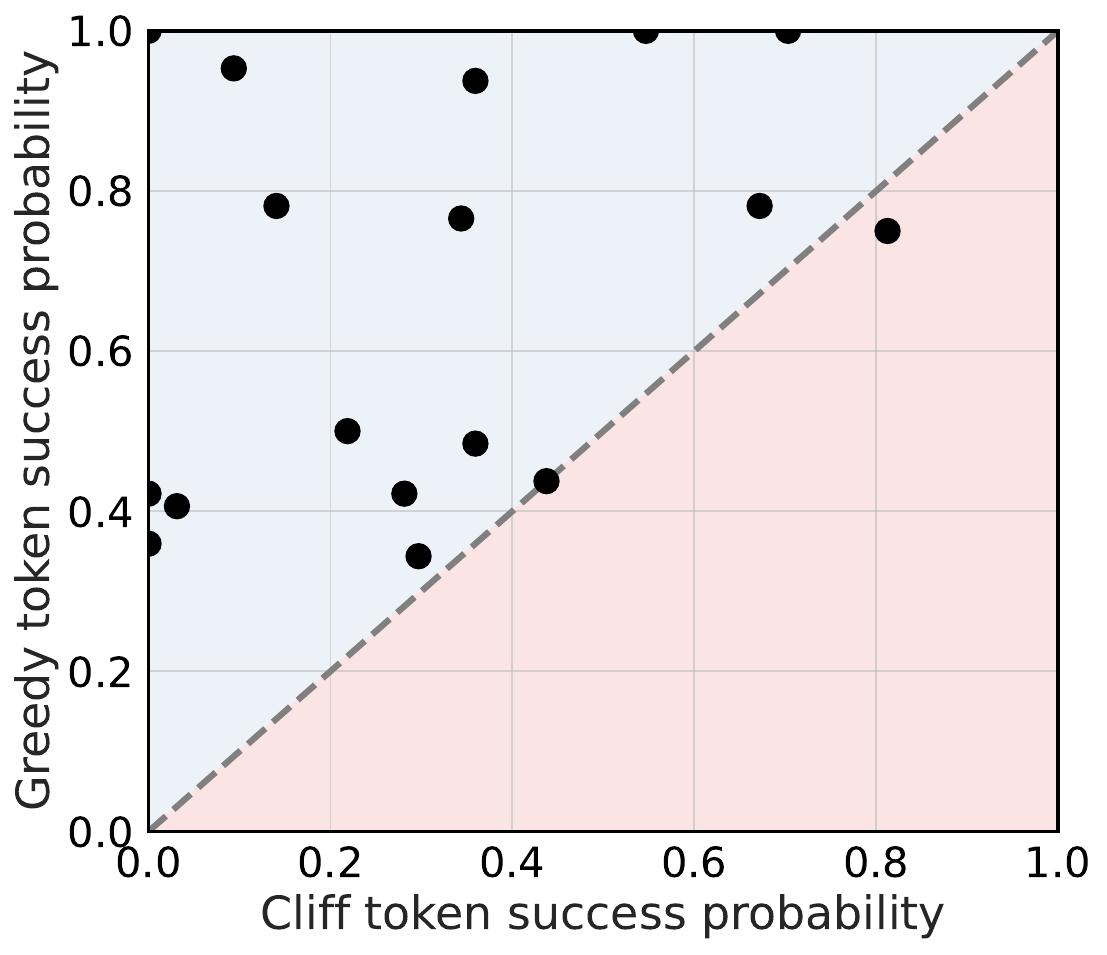}
  \caption{Counterfactual analysis on sampled-off cliffs using \texttt{Qwen3-8B}. The experiment evaluates 17 sampled-off cliffs identified across the GSM1K, MATH500, and AIME 2025 datasets. Greedy-token success probability refers to the success probability when these sampled-off cliffs are replaced with greedy tokens. Points above the diagonal dashed line indicate an increase in success probability.}
  \label{fig:counterfactual_qwen3}
\end{figure}

As illustrated in \Cref{fig:counterfactual_qwen3}, the vast majority of data points lie significantly above the $y=x$ reference line. Out of all evaluated instances, we observe only one case where the success probability remains unchanged and one case where it slightly decreases. In all other instances, substituting with the greedy token successfully increases the success probability. This confirms that sampled-off cliffs are indeed suboptimal choices drawn by chance, and replacing them with the greedy token effectively mitigates the success probability drop.

\newpage

\section{Cross-scale cliff probability mass transfer}
\label[appendix]{app:cross_model_transfer}

\begin{figure}[H]
  \centering
  \includegraphics[width=0.6\linewidth]{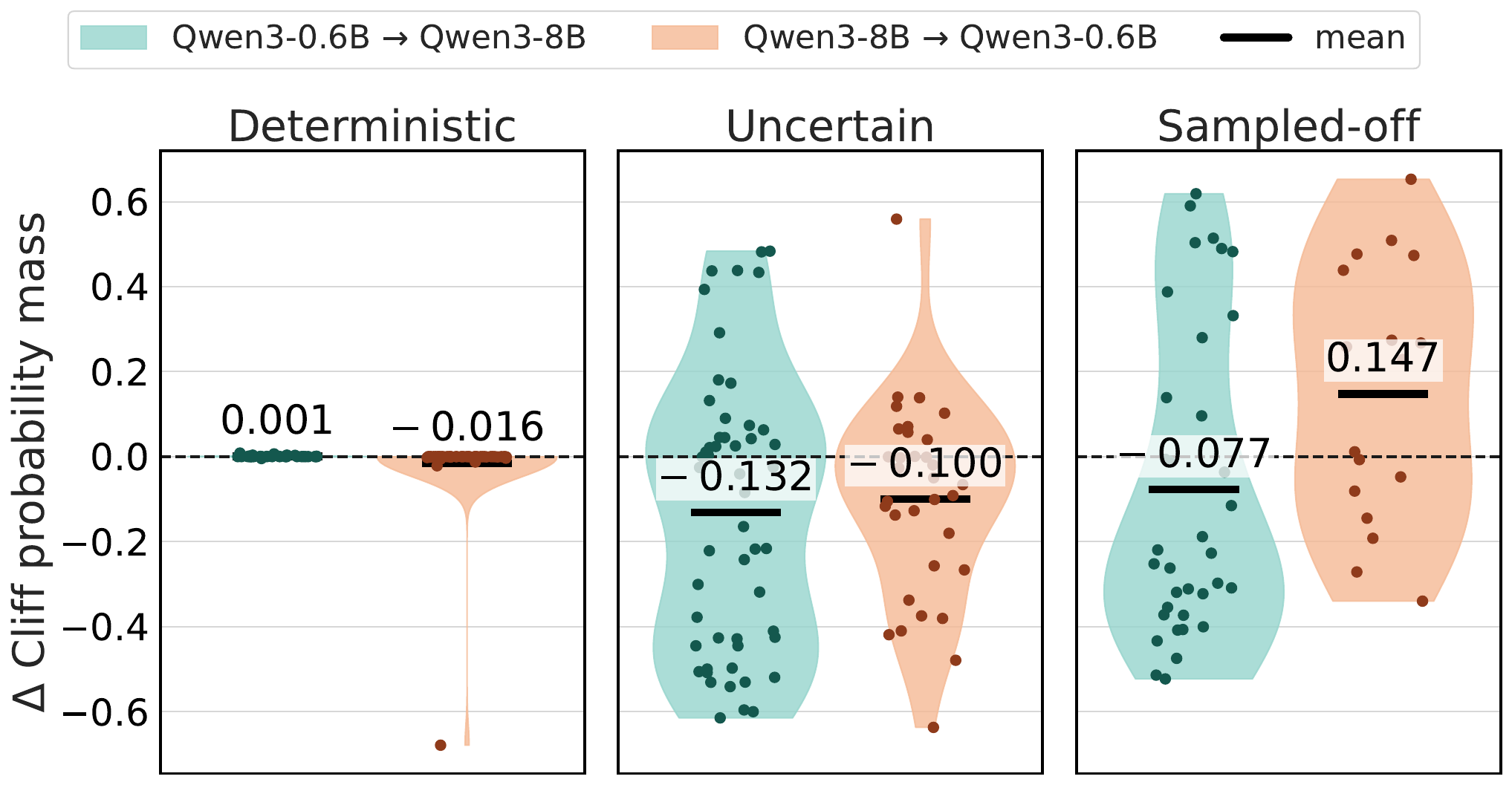}
  \caption{
Cliff probability mass shifts ($\Delta$) at identified cliff positions upon cross-scale transfer between \texttt{Qwen3-0.6B} and \texttt{Qwen3-8B}. Deterministic cliffs are scale-invariant ($\Delta \approx 0$). Uncertain cliffs show overall mass decrease. Sampled-off cliffs exhibit scale-asymmetry.}
  \label{fig:violin_qwen}
\end{figure}

While \Cref{sec:cliff_taxonomy_across_seven_models} shows macro-level distributional shifts across model families and scales, it remains unclear whether scale-dependent cliff behavior persists at exact token positions within a family. To investigate this token-level transferability, we conduct cross-scale transfer experiments.  Given a cliff token $c_{t^*}$ identified by a source model, we evaluate a target model on the same source-generated cliff-del prefix ($\boldsymbol{x} \oplus \boldsymbol{c}_{< t^*}$). We then observe how the cliff probability mass at this decoding step $t^*$ shifts between the two models. Because exact position alignment requires a shared tokenizer, we restrict this evaluation to two models of different sizes within the same family: \texttt{Qwen3-0.6B} and \texttt{Qwen3-8B}. Our experiments show distinct distributions of cliff probability mass difference ($\Delta$) for each type (\Cref{fig:violin_qwen}):

\begin{itemize}[leftmargin=*,nosep]
\item \textit{Deterministic cliffs: Scale-invariance}. We observe that the difference in cliff probability mass is near zero ($\Delta \approx 0$) in both \texttt{Qwen3-0.6B} $\rightarrow$ \texttt{8B} and \texttt{8B} $\rightarrow$ \texttt{0.6B} transfers. At these exact reasoning positions, both models sample the identical deterministic cliffs. Specifically, at 44 out of 46 deterministic cliff positions identified by \texttt{Qwen3-8B}, \texttt{Qwen3-0.6B} also samples the same cliff tokens. Conversely, at all 37 deterministic cliff positions identified by \texttt{Qwen3-0.6B}, \texttt{Qwen3-8B} mirrors this selection (detailed in \Cref{app:rank_difference}). The bidirectional overlap indicates that deterministic cliffs are largely scale-invariant within the \texttt{Qwen3} family. Given the same reasoning prefix, both model sizes select the same failure-triggering token, suggesting that these cliff tokens reflect a shared family-level bias.

\item \textit{Uncertain cliffs: Model-specific failure patterns}. Regardless of the transfer direction, the probability mass of these cliff tokens decreases during cross-model transfers ($\Delta \approx -0.13$ and $-0.10$). This overall drop suggests that uncertain cliffs reflect model-specific failure patterns, rather than shared reasoning bottlenecks across the \texttt{Qwen3} family. Consequently, uncertain cliffs expose the unique failure spots of individual models.

\item \textit{Sampled-off cliffs: Scale-asymmetry}. We observe a clear asymmetry in cliff probability mass shifts: the mean $\Delta$ is negative for \texttt{Qwen3-0.6B} $\rightarrow$ \texttt{8B}, but positive for \texttt{8B} $\rightarrow$ \texttt{0.6B}. Positions identified as sampled-off cliffs by \texttt{Qwen3-8B} act as high-probability cliff regions for \texttt{Qwen3-0.6B}. This connects to the scaling trend in \Cref{sec:cliff_taxonomy_across_seven_models}, where the sampled-off cliff proportion decreases ($28{.}2 \rightarrow 17{.}7$) with \texttt{Qwen3} model scaling.
\end{itemize}

\section{Rank difference of cliff token in cross-model experiment}
\label[appendix]{app:rank_difference}

\begin{figure}[H]
  \centering
  \includegraphics[width=0.8\linewidth]{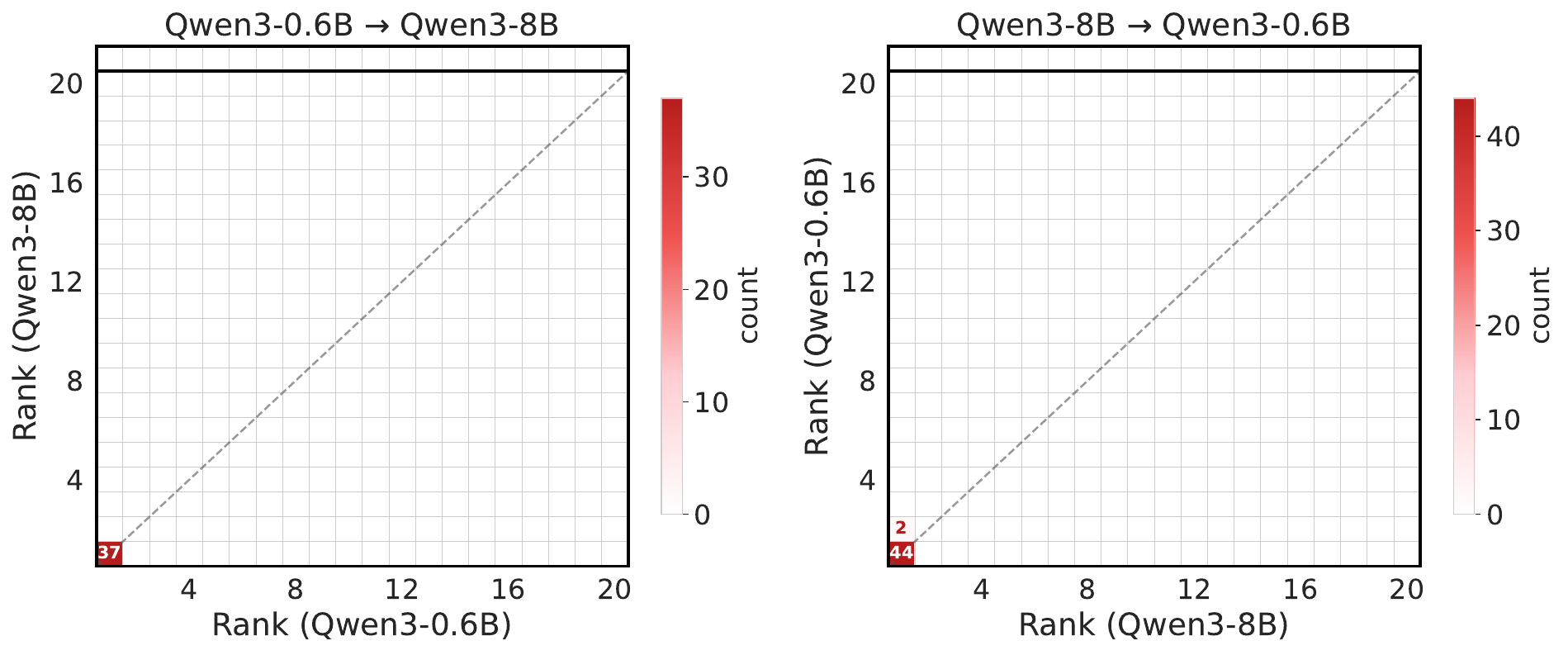}
  \caption{Rank transfer of \emph{deterministic cliffs} between \texttt{Qwen3-0.6B} and \texttt{Qwen3-8B}. The cliff token's rank is preserved at 1 in both transfer directions, with only 2 deterministic cliffs moving to rank 2 in the \texttt{Qwen3-8B} $\rightarrow$ \texttt{Qwen3-0.6B} direction.}
  \label{fig:rank_deterministic}
\end{figure}

\begin{figure}[H]
  \centering
  \includegraphics[width=0.8\linewidth]{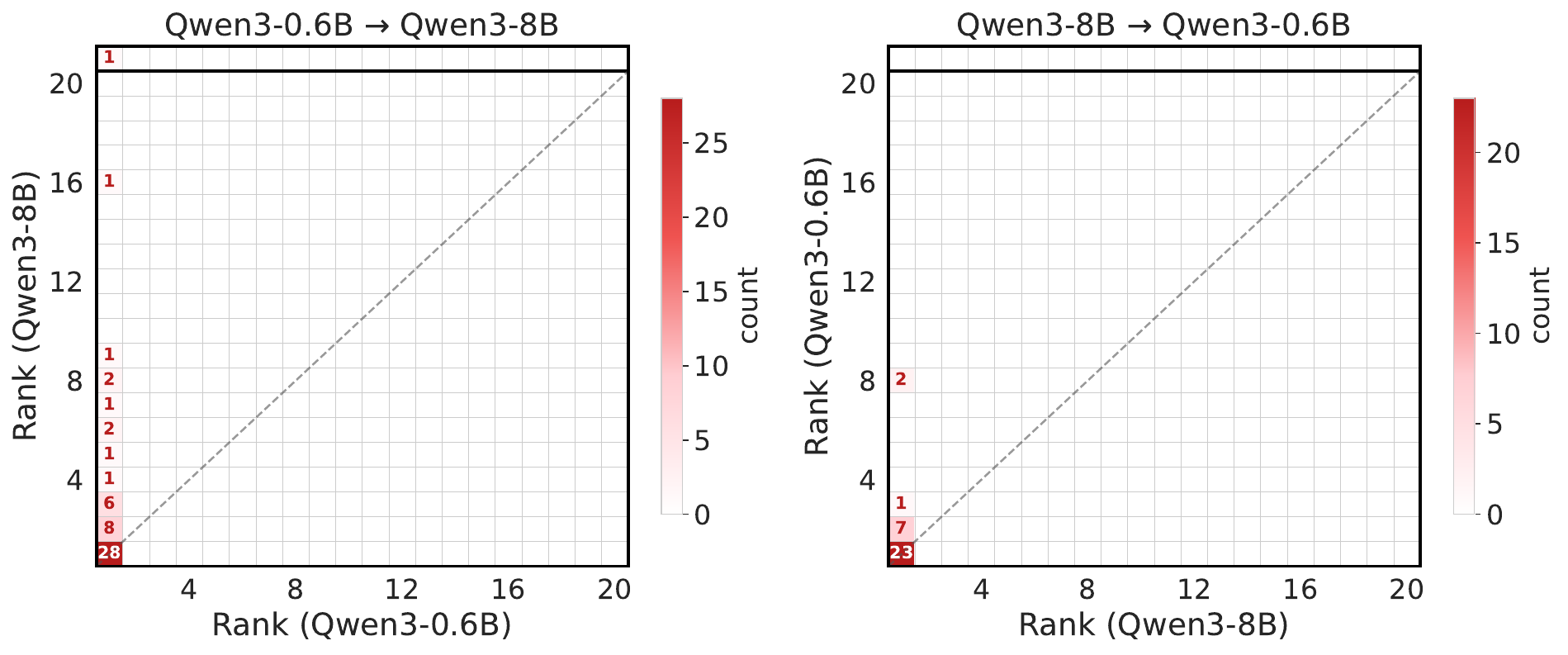}
  \caption{Rank transfer of \emph{uncertain cliffs}. In the \texttt{Qwen3-0.6B} $\rightarrow$ \texttt{Qwen3-8B} direction, $24/52$ cliff tokens shifted away from rank 1, while in the \texttt{Qwen3-8B} $\rightarrow$ \texttt{Qwen3-0.6B} direction, $10/33$ cliff tokens shifted away from rank 1.}
  \label{fig:rank_unertain}
\end{figure}

\begin{figure}[H]
  \centering
  \includegraphics[width=0.8\linewidth]{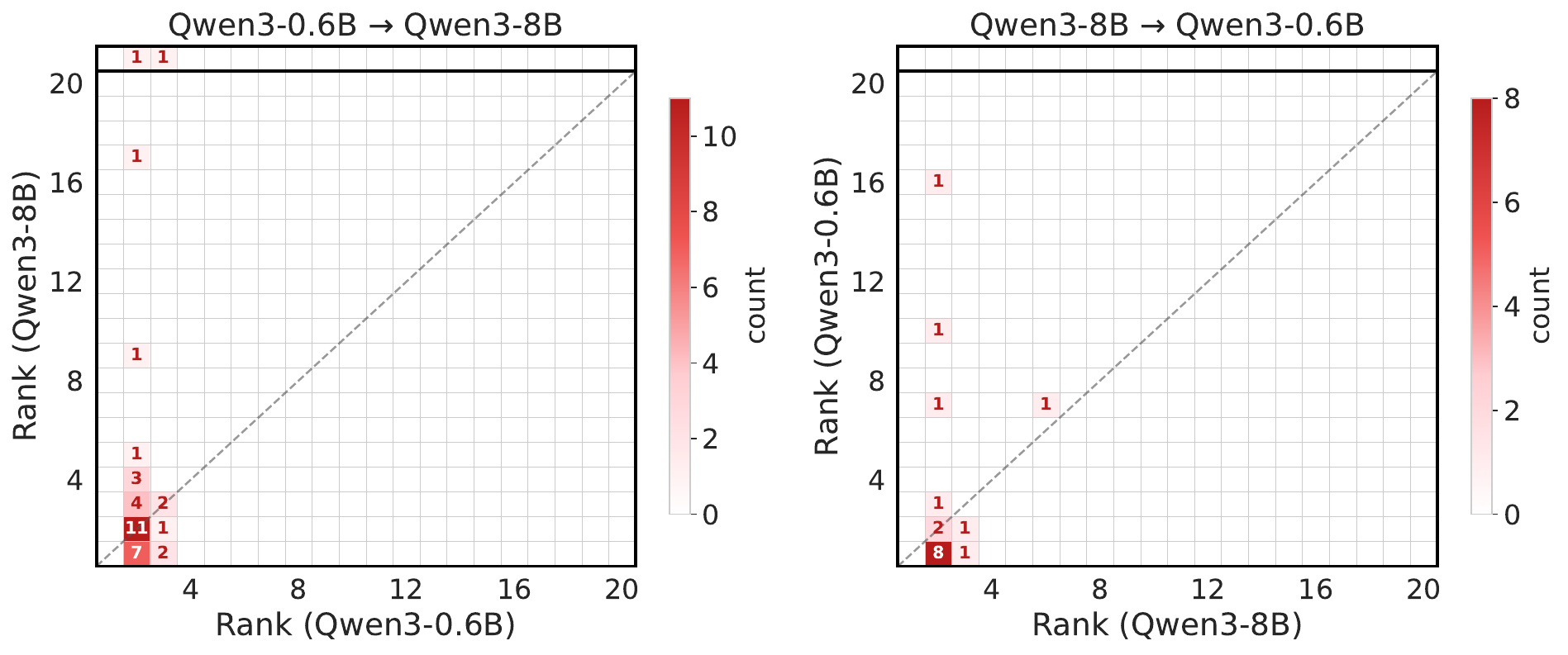}
  \caption{Rank transfer of \emph{sampled-off cliffs}. The rank shifts asymmetrically across transfer directions. In the \texttt{Qwen3-0.6B} $\rightarrow$ \texttt{Qwen3-8B} direction, the rank is preserved, increased, or decreased in roughly comparable proportions, whereas in the \texttt{Qwen3-8B} $\rightarrow$ \texttt{Qwen3-0.6B} direction, $10/17$ cliff tokens move to a lower rank index (higher probability) in the target model.}
  \label{fig:rank_sampled_off}
\end{figure}

This appendix complements \Cref{app:cross_model_transfer} with a token-position-level view of cross-model transfer. For each cliff token $c_{t^*}$ identified by the source model, we feed the cliff-del prefix $\boldsymbol{x} \oplus \boldsymbol{c}_{<t^*}$ to the target model and record the rank of $c_{t^*}$ within the target's top-20 candidates at step $t^*$. Tokens outside the top-20 are placed at rank 21. \Cref{fig:rank_deterministic,fig:rank_unertain,fig:rank_sampled_off} show the resulting rank shifts for deterministic, uncertain, and sampled-off cliffs, respectively.

\newpage

\section{Full results of Cliff-DPO}
\label[appendix]{app:dpo_full}

We trained five variants of Cliff-DPO for each model: \textit{deterministic}, \textit{uncertain}, \textit{sampled-off}, \textit{uncertain + sampled-off}, and \textit{all}. Each name indicates the subset of cliff tokens used during training. As shown in \Cref{tab:cliff_dpo}, for all models, the \textit{uncertain} and \textit{sampled-off} variants outperform the \textit{deterministic} variant across GSM8K, GSM1K, and MATH500, as reported in \Cref{sec:cliff_dpo_results}. Furthermore, the \textit{all} variant, which adds the deterministic subset to the \textit{uncertain + sampled-off} variant, does not consistently outperform the \textit{uncertain + sampled-off} variant. These results suggest that incorporating the deterministic subset provides little additional benefit for post-training.

\begin{table}[H]
  \caption{Cliff-DPO variants and baselines on three models. The table reports mean accuracy or avg@$64$, along with the sample standard error. \textit{Updated tokens} counts gradient-updated token positions. \textbf{Bold} and \underline{underline} mark the best and second-best results within each model on GSM8K, GSM1K, and MATH500. AIME 2025 is not ranked.}
  \label{tab:cliff_dpo}
  \label{tab:cliff_dpo_qwen06}
  \label{tab:cliff_dpo_llama32_1b}
  \label{tab:cliff_dpo_llama31_8b}
  \centering
  \small
  \setlength{\tabcolsep}{6pt}
  \renewcommand{\arraystretch}{1.05}
  \begin{adjustbox}{width=\linewidth}
  \begin{tabular}{l cccc r}
    \toprule
     & \multicolumn{3}{c}{\textbf{Mean accuracy}} & \textbf{avg@64} & \\
    \cmidrule(lr){2-4} \cmidrule(lr){5-5}
    \textbf{Method} & GSM8K & GSM1K & MATH500 & AIME 2025 & \textbf{Updated tokens} \\
    \midrule
    \rowcolor{taxonomyfamilyblue}\multicolumn{6}{l}{\textbf{\texttt{Llama-3.1-8B}}} \\
    Base & $82.2_{\pm 0.20}$ & $78.2_{\pm 0.07}$ & $48.1_{\pm 0.29}$ & $0.7_{\pm 0.20}$ & \textendash \\
    + DPO & $83.2_{\pm 0.25}$ & $78.4_{\pm 0.19}$ & $47.6_{\pm 0.12}$ & $0.6_{\pm 0.03}$ & 1{,}952{,}191 \\
    + cDPO & $\mathbf{84.7}_{\pm 0.35}$ & $\mathbf{81.7}_{\pm 0.34}$ & $49.1_{\pm 1.39}$ & $0.8_{\pm 0.21}$ & 3{,}232{,}251 \\
    \cmidrule(lr){1-6}
    \multicolumn{6}{l}{+ Cliff-DPO} \\
    \rowcolor{rowgray}\quad\textit{deterministic} & $82.1_{\pm 0.31}$ & $78.1_{\pm 0.30}$ & $48.5_{\pm 0.88}$ & $0.6_{\pm 0.09}$ & 5{,}618 \\
    \rowcolor{rowgray}\quad\textit{uncertain} & $\underline{84.6}_{\pm 0.15}$ & $79.6_{\pm 0.07}$ & $\mathbf{50.8}_{\pm 0.72}$ & $0.5_{\pm 0.03}$ & 20{,}848 \\
    \rowcolor{rowgray}\quad\textit{sampled-off} & $83.1_{\pm 0.14}$ & $78.3_{\pm 0.15}$ & $49.1_{\pm 1.09}$ & $0.5_{\pm 0.14}$ & 13{,}236 \\
    \rowcolor{black!10}\quad\textit{uncertain + sampled-off} & $83.9_{\pm 0.24}$ & $\underline{79.8}_{\pm 0.31}$ & $49.8_{\pm 0.42}$ & $0.6_{\pm 0.08}$ & 34{,}084 \\
    \rowcolor{black!15}\quad\textit{all} & $84.5_{\pm 0.25}$ & $\underline{79.8}_{\pm 0.13}$ & $\underline{50.5}_{\pm 0.07}$ & $0.5_{\pm 0.02}$ & 39{,}708 \\
    \midrule[0.8pt]
    \rowcolor{taxonomyfamilyblue}\multicolumn{6}{l}{\textbf{\texttt{Llama-3.2-1B}}} \\
    Base & $39.3_{\pm 0.17}$ & $36.0_{\pm 0.54}$ & $25.4_{\pm 0.42}$ & $0.2_{\pm 0.08}$ & \textendash \\
    + DPO & $40.8_{\pm 0.33}$ & $35.7_{\pm 0.12}$ & $27.5_{\pm 0.44}$ & $0.2_{\pm 0.05}$ & 2{,}264{,}627 \\
    + cDPO & $45.1_{\pm 0.56}$ & $\mathbf{41.4}_{\pm 0.43}$ & $26.3_{\pm 0.18}$ & $0.2_{\pm 0.05}$ & 7{,}264{,}379 \\
    \cmidrule(lr){1-6}
    \multicolumn{6}{l}{+ Cliff-DPO} \\
    \rowcolor{rowgray}\quad\textit{deterministic} & $41.5_{\pm 0.50}$ & $36.2_{\pm 0.59}$ & $26.6_{\pm 0.40}$ & $0.0_{\pm 0.02}$ & 3{,}900 \\
    \rowcolor{rowgray}\quad\textit{uncertain} & $44.5_{\pm 0.28}$ & $37.4_{\pm 0.17}$ & $27.9_{\pm 0.33}$ & $0.1_{\pm 0.05}$ & 14{,}886 \\
    \rowcolor{rowgray}\quad\textit{sampled-off} & $\mathbf{45.6}_{\pm 0.46}$ & $\underline{40.4}_{\pm 0.70}$ & $\mathbf{29.6}_{\pm 0.46}$ & $0.1_{\pm 0.05}$ & 47{,}266 \\
    \rowcolor{black!10}\quad\textit{uncertain + sampled-off} & $\underline{45.3}_{\pm 0.48}$ & $39.4_{\pm 0.41}$ & $\underline{29.2}_{\pm 0.76}$ & $0.0_{\pm 0.03}$ & 62{,}152 \\
    \rowcolor{black!15}\quad\textit{all} & $\mathbf{45.6}_{\pm 0.37}$ & $39.0_{\pm 0.14}$ & $28.1_{\pm 0.24}$ & $0.2_{\pm 0.03}$ & 66{,}388 \\
    \midrule[0.8pt]
    \rowcolor{taxonomyfamilyblue}\multicolumn{6}{l}{\textbf{\texttt{Qwen3-0.6B}}} \\
    Base & $61.5_{\pm 0.27}$ & $57.0_{\pm 0.06}$ & $51.6_{\pm 0.00}$ & $2.0_{\pm 1.04}$ & \textendash \\
    + DPO & $63.0_{\pm 0.60}$ & $56.6_{\pm 0.45}$ & $52.6_{\pm 0.80}$ & $2.4_{\pm 0.10}$ & 2{,}862{,}845 \\
    + cDPO & $\underline{66.4}_{\pm 0.48}$ & $61.4_{\pm 0.53}$ & $54.0_{\pm 0.61}$ & $1.6_{\pm 1.61}$ & 5{,}829{,}052 \\
    \cmidrule(lr){1-6}
    \multicolumn{6}{l}{+ Cliff-DPO} \\
    \rowcolor{rowgray}\quad\textit{deterministic} & $63.7_{\pm 0.18}$ & $57.3_{\pm 0.48}$ & $51.9_{\pm 1.51}$ & $2.7_{\pm 0.09}$ & 5{,}538 \\
    \rowcolor{rowgray}\quad\textit{uncertain} & $\mathbf{66.8}_{\pm 0.07}$ & $62.0_{\pm 0.17}$ & $53.7_{\pm 0.47}$ & $2.7_{\pm 1.35}$ & 18{,}122 \\
    \rowcolor{rowgray}\quad\textit{sampled-off} & $\mathbf{66.8}_{\pm 0.24}$ & $\underline{62.9}_{\pm 0.21}$ & $\underline{54.9}_{\pm 0.93}$ & $3.7_{\pm 0.19}$ & 14{,}794 \\
    \rowcolor{black!10}\quad\textit{uncertain + sampled-off} & $66.3_{\pm 0.47}$ & $\mathbf{63.0}_{\pm 0.28}$ & $\mathbf{56.1}_{\pm 0.59}$ & $4.7_{\pm 0.27}$ & 32{,}916 \\
    \rowcolor{black!15}\quad\textit{all} & $65.6_{\pm 0.07}$ & $61.0_{\pm 0.23}$ & $54.5_{\pm 0.85}$ & $1.3_{\pm 1.34}$ & 38{,}454 \\
    \bottomrule
  \end{tabular}
  \end{adjustbox}
\end{table}



\newpage

\section{Cliff-token probability before and after Cliff-DPO}
\label[appendix]{app:dpo_prob_shift}

To analyze the cliff-probability shift after Cliff-DPO discussed in \Cref{sec:cliff_dpo_results}, we apply the cliff token identification procedure of \Cref{sec:cliff_token_identification} to pre-identified cliff token positions in GSM1K, MATH500, and AIME 2025 traces. For each cliff token at position $t$, we measure its generation probability to evaluate whether the model reduces the probability of selecting it after Cliff-DPO. This procedure is performed on $127$, $111$, and $124$ held-out cliff positions for \texttt{Llama-3.1-8B}, \texttt{Llama-3.2-1B}, and \texttt{Qwen3-0.6B}, respectively. We average the results over the three training seeds for each Cliff-DPO variant. \Cref{tab:dpo_prob_shift} reports the cliff-token probability shifts for all variants across the three cliff types.

For deterministic cliff tokens, the initial probability is $99.9\%$, and no Cliff-DPO variant changes it by more than $0.24$ pp. For uncertain cliff tokens, all Cliff-DPO variants reduce the cliff-token probability. For sampled-off cliff tokens, only the \textit{sampled-off} variant consistently reduces the cliff-token probability.

\begin{table}[H]
  \caption{Average cliff-token probability before and after Cliff-DPO. Rows denote training variants and columns denote evaluation cliff types. Changes from the base model are in percentage points (pp). Darker gray cells mark the diagonal entries reported in Fig.~\ref{fig:cliff_dpo_main_results}.}
  \label{tab:dpo_prob_shift}
  \centering
  \small
  \setlength{\tabcolsep}{6pt}
  \renewcommand{\arraystretch}{1.05}
  \begin{tabular}{l ccc}
    \toprule
     & \multicolumn{3}{c}{\textbf{Average cliff-token probability (\%)}} \\
    \cmidrule(lr){2-4}
    \textbf{Method} & Deterministic & Uncertain & Sampled-off \\
    \midrule
    \rowcolor{taxonomyfamilyblue}\multicolumn{4}{l}{\textbf{\texttt{Llama-3.1-8B}}} \\
    Base & $99.9$ & $56.4$ & $20.1$ \\
    \cmidrule(lr){1-4}
    \multicolumn{4}{l}{+ Cliff-DPO} \\
    \rowcolor{rowgray}\quad\textit{deterministic} & \cellcolor{black!12}$99.9\pdown{-0.00}$ & $56.2\pdown{-0.25}$ & $20.4\pup{+0.26}$ \\
    \rowcolor{rowgray}\quad\textit{uncertain} & $99.9\pdown{-0.05}$ & \cellcolor{black!12}$54.9\pdown{-1.56}$ & $20.4\pup{+0.28}$ \\
    \rowcolor{rowgray}\quad\textit{sampled-off} & $99.9\pdown{-0.01}$ & $55.7\pdown{-0.73}$ & \cellcolor{black!12}$20.0\pdown{-0.09}$ \\
    \midrule[0.8pt]
    \rowcolor{taxonomyfamilyblue}\multicolumn{4}{l}{\textbf{\texttt{Llama-3.2-1B}}} \\
    Base & $99.9$ & $62.2$ & $21.7$ \\
    \cmidrule(lr){1-4}
    \multicolumn{4}{l}{+ Cliff-DPO} \\
    \rowcolor{rowgray}\quad\textit{deterministic} & \cellcolor{black!12}$99.9\pdown{-0.01}$ & $61.9\pdown{-0.35}$ & $21.8\pup{+0.08}$ \\
    \rowcolor{rowgray}\quad\textit{uncertain} & $99.9\pdown{-0.05}$ & \cellcolor{black!12}$59.6\pdown{-2.67}$ & $21.6\pdown{-0.08}$ \\
    \rowcolor{rowgray}\quad\textit{sampled-off} & $99.9\pdown{-0.01}$ & $55.6\pdown{-6.61}$ & \cellcolor{black!12}$19.0\pdown{-2.72}$ \\
    \midrule[0.8pt]
    \rowcolor{taxonomyfamilyblue}\multicolumn{4}{l}{\textbf{\texttt{Qwen3-0.6B}}} \\
    Base & $99.9$ & $63.0$ & $28.6$ \\
    \cmidrule(lr){1-4}
    \multicolumn{4}{l}{+ Cliff-DPO} \\
    \rowcolor{rowgray}\quad\textit{deterministic} & \cellcolor{black!12}$99.8\pdown{-0.06}$ & $62.2\pdown{-0.80}$ & $28.8\pup{+0.16}$ \\
    \rowcolor{rowgray}\quad\textit{uncertain} & $99.6\pdown{-0.24}$ & \cellcolor{black!12}$58.0\pdown{-4.99}$ & $26.9\pdown{-1.74}$ \\
    \rowcolor{rowgray}\quad\textit{sampled-off} & $99.7\pdown{-0.14}$ & $59.1\pdown{-3.86}$ & \cellcolor{black!12}$26.6\pdown{-2.02}$ \\
    \bottomrule
  \end{tabular}
\end{table}

\newpage

\section{Cliff-DPO ablation under matched update-token budget}
\label[appendix]{app:dpo_controlled}

To isolate the effect of cliff type from that of training-data quantity, we subsample each single-type training set to match the number of gradient-updated tokens in the \textit{deterministic} subset: $5{,}618$ for \texttt{Llama-3.1-8B}, $3{,}900$ for \texttt{Llama-3.2-1B}, and $5{,}538$ for \texttt{Qwen3-0.6B}. The \textit{deterministic} subset is fully used, whereas the \textit{uncertain} and \textit{sampled-off} subsets are subsampled using a random seed of $42$. \Cref{tab:cliff_dpo-controlled} reports the results.

Under this matched-token setting, the main finding of \Cref{sec:cliff_dpo_results} continues to hold for \texttt{Llama-3.1-8B} and \texttt{Qwen3-0.6B}: both the \textit{uncertain} and \textit{sampled-off} variants outperform the \textit{deterministic} variant on GSM8K, GSM1K, and MATH500. \texttt{Llama-3.2-1B} is an exception: the \textit{deterministic} variant performs best on GSM8K and GSM1K. This may reflect its small number of training tokens ($3{,}900$ tokens). At their full budgets in \Cref{sec:cliff_dpo_results}, the \textit{uncertain} and \textit{sampled-off} variants gain a further $3.8$ and $5.4$ points on GSM8K, respectively.

\begin{table}[H]
  \caption{Token-matched ablation of Cliff-DPO variants on three models. Each variant is trained on a single cliff type, with the update-token budget subsampled within each model to match the \textit{deterministic} subset. Notation follows \Cref{tab:cliff_dpo}.}
  \label{tab:cliff_dpo-controlled}
  \centering
  \small
  \setlength{\tabcolsep}{6pt}
  \renewcommand{\arraystretch}{1.05}
  \begin{adjustbox}{width=0.893\linewidth}
  \begin{tabular}{l cccc r}
    \toprule
     & \multicolumn{3}{c}{\textbf{Mean accuracy}} & \textbf{avg@64} & \\
    \cmidrule(lr){2-4} \cmidrule(lr){5-5}
    \textbf{Method} & GSM8K & GSM1K & MATH500 & AIME 2025 & \textbf{Updated tokens} \\
    \midrule
    \rowcolor{taxonomyfamilyblue}\multicolumn{6}{l}{\textbf{\texttt{Llama-3.1-8B}}} \\
    Base & $\underline{82.2}_{\pm 0.20}$ & $\underline{78.2}_{\pm 0.07}$ & $48.1_{\pm 0.29}$ & $0.7_{\pm 0.20}$ & \textendash \\
    \cmidrule(lr){1-6}
    \multicolumn{6}{l}{+ Cliff-DPO} \\
    \rowcolor{rowgray}\quad\textit{deterministic} & $82.1_{\pm 0.31}$ & $78.1_{\pm 0.30}$ & $48.5_{\pm 0.88}$ & $0.6_{\pm 0.09}$ & 5{,}618 \\
    \rowcolor{rowgray}\quad\textit{uncertain} & $\mathbf{83.0}_{\pm 0.24}$ & $\mathbf{78.5}_{\pm 0.10}$ & $\underline{48.7}_{\pm 0.37}$ & $0.5_{\pm 0.31}$ & 5{,}618 \\
    \rowcolor{rowgray}\quad\textit{sampled-off} & $\mathbf{83.0}_{\pm 0.14}$ & $\underline{78.2}_{\pm 0.18}$ & $\mathbf{48.9}_{\pm 0.74}$ & $0.5_{\pm 0.27}$ & 5{,}618 \\
    \midrule[0.8pt]
    \rowcolor{taxonomyfamilyblue}\multicolumn{6}{l}{\textbf{\texttt{Llama-3.2-1B}}} \\
    Base & $39.3_{\pm 0.17}$ & $\underline{36.0}_{\pm 0.54}$ & $25.4_{\pm 0.42}$ & $0.2_{\pm 0.08}$ & \textendash \\
    \cmidrule(lr){1-6}
    \multicolumn{6}{l}{+ Cliff-DPO} \\
    \rowcolor{rowgray}\quad\textit{deterministic} & $\mathbf{41.5}_{\pm 0.50}$ & $\mathbf{36.2}_{\pm 0.59}$ & $26.6_{\pm 0.40}$ & $0.0_{\pm 0.02}$ & 3{,}900 \\
    \rowcolor{rowgray}\quad\textit{uncertain} & $\underline{40.7}_{\pm 0.24}$ & $34.7_{\pm 0.05}$ & $\mathbf{28.4}_{\pm 0.70}$ & $0.1_{\pm 0.03}$ & 3{,}900 \\
    \rowcolor{rowgray}\quad\textit{sampled-off} & $40.2_{\pm 0.21}$ & $35.6_{\pm 0.20}$ & $\underline{27.6}_{\pm 0.92}$ & $0.1_{\pm 0.03}$ & 3{,}900 \\
    \midrule[0.8pt]
    \rowcolor{taxonomyfamilyblue}\multicolumn{6}{l}{\textbf{\texttt{Qwen3-0.6B}}} \\
    Base & $61.5_{\pm 0.27}$ & $57.0_{\pm 0.06}$ & $51.6_{\pm 0.00}$ & $2.0_{\pm 1.04}$ & \textendash \\
    \cmidrule(lr){1-6}
    \multicolumn{6}{l}{+ Cliff-DPO} \\
    \rowcolor{rowgray}\quad\textit{deterministic} & $63.7_{\pm 0.18}$ & $57.3_{\pm 0.48}$ & $\underline{51.9}_{\pm 1.51}$ & $\underline{2.7}_{\pm 0.09}$ & 5{,}538 \\
    \rowcolor{rowgray}\quad\textit{uncertain} & $\mathbf{64.5}_{\pm 0.15}$ & $\underline{58.3}_{\pm 0.47}$ & $\mathbf{53.1}_{\pm 0.24}$ & $\underline{2.7}_{\pm 1.41}$ & 5{,}538 \\
    \rowcolor{rowgray}\quad\textit{sampled-off} & $\underline{64.0}_{\pm 0.05}$ & $\mathbf{58.6}_{\pm 0.29}$ & $\mathbf{53.1}_{\pm 0.18}$ & $\mathbf{3.1}_{\pm 1.83}$ & 5{,}538 \\
    \bottomrule
  \end{tabular}
  \end{adjustbox}
\end{table}

\newpage

\section{Training and evaluation configuration}
\label[appendix]{app:appendix_training}

Experiments in \Cref{sec:cliff_dpo} use \texttt{Llama-3.1-8B} \citep{llama3}, \texttt{Llama-3.2-1B} \citep{llama3}, and \texttt{Qwen3-0.6B} \citep{Qwen3} as base models, and all preference data is constructed from the GSM8K \citep{gsm8k} training set ($7{,}473$ problems). All training is conducted in BF16 with seeds \texttt{42}, \texttt{43}, and \texttt{44}.

\subsection{DPO}
\label[appendix]{app:appendix_dpo}

The DPO baseline follows \citet{DPO}. We sample $64$ traces per problem from the base model on the GSM8K training set and pair, for each problem, one correct trace (chosen) with one incorrect trace (rejected). This produces $4{,}006$ trace-level preference pairs for \texttt{Llama-3.1-8B}, $6{,}797$ for \texttt{Llama-3.2-1B}, and $5{,}566$ for \texttt{Qwen3-0.6B}. The remaining problems whose $64$ traces are all correct or all incorrect are excluded from the dataset.

\paragraph{Training hyperparameters.} Sigmoid DPO loss with $\beta = 0.1$ and no label smoothing. Training runs for $1$ epoch with learning rate $1\mathrm{e}{-6}$, cosine decay, and warmup ratio $0.1$. The batch size is $64$. Gradient clipping uses a max norm of $1.0$; weight decay is disabled. LoRA: rank $r=32$, $\alpha=64$, dropout $0.05$, applied to all seven attention and feed-forward projection modules: \texttt{q\_proj}, \texttt{k\_proj}, \texttt{v\_proj}, \texttt{o\_proj}, \texttt{gate\_proj}, \texttt{up\_proj}, \texttt{down\_proj}. We additionally verified that aligning DPO's training configuration with that of Cliff-DPO degrades its performance; we therefore report DPO under configuration recommended by \citet{DPO}.

\subsection{cDPO}
\label[appendix]{app:appendix_cdpo}

We reproduce the five-step pipeline of \citet{CriticalTokens} without modification, except that the models are replaced with the three base models used in this work, for consistency with our other baselines. The contrastive logit formulation, $\beta=1.0$ in the Contrastive Estimation (CE) score $s_t$, the \texttt{only\_neg=True} setting, and the per-token $(1-s_t)$ weighting on the rejected side all follow the exact configurations detailed in \citet{CriticalTokens}.

\paragraph{Auxiliary contrastive SFT models.}
The positive SFT adapter $\pi_+$ is trained on one correct trace per problem. The negative SFT adapter $\pi_-$ is trained on all incorrect traces from the problems that have both correct and incorrect rollouts, since multiple distinct failure modes exist for the same problem. For $\pi_+$, this gives $4{,}006$ examples for \texttt{Llama-3.1-8B}, $6{,}797$ for \texttt{Llama-3.2-1B}, and $5{,}566$ for \texttt{Qwen3-0.6B}. For $\pi_-$, it gives $6{,}062$, $19{,}051$, and $9{,}963$ examples for the same three models. Both share the same hyperparameters: learning rate $3\mathrm{e}{-4}$, $1$ epoch, batch size $18$, $100$ warmup steps, max sequence length $2{,}048$. LoRA: rank $r=8$, $\alpha=16$, dropout $0.1$, targeting only \texttt{gate\_proj}, \texttt{down\_proj}, and \texttt{up\_proj}.

\paragraph{cDPO fine-tuning.} The final preference dataset contains $6{,}062$ pairs for \texttt{Llama-3.1-8B}, $19{,}051$ for \texttt{Llama-3.2-1B}, and $9{,}963$ for \texttt{Qwen3-0.6B}, each annotated with per-token CE probabilities. Sigmoid DPO loss with $\beta = 1.0$ and no label smoothing. Training runs for $3$ epochs (the setting reported as best in the original paper), with learning rate $4\mathrm{e}{-5}$, cosine decay, warmup ratio $0.1$, and weight decay $0.01$. Batch size $8$. LoRA: rank $r=16$, $\alpha=32$, no dropout, targeting \texttt{q\_proj}, \texttt{k\_proj}, \texttt{v\_proj}, \texttt{o\_proj}, \texttt{gate\_proj}, \texttt{up\_proj}, \texttt{down\_proj}.

\subsection{Cliff-DPO}

All Cliff-DPO variants reported in \Cref{fig:cliff_dpo_main_results}, \Cref{tab:cliff_dpo}, and \Cref{tab:cliff_dpo-controlled} share an identical training configuration; they differ only in which subset of cliff tokens is included in training.

\paragraph{Training hyperparameters.} Sigmoid DPO loss with $\beta = 0.1$ and no label smoothing. Learning rate $1\mathrm{e}{-6}$ for \texttt{Llama-3.1-8B}, and $5\mathrm{e}{-6}$ for \texttt{Llama-3.2-1B} and \texttt{Qwen3-0.6B}; cosine decay, warmup ratio $0.1$. Training runs for $1$ epoch with an effective batch size of $64$ (batch size $4$, gradient accumulation $16$). Gradient clipping at max norm $1.0$, no weight decay. LoRA: rank $r=32$, $\alpha=64$, dropout $0.05$, targeting \texttt{q\_proj}, \texttt{k\_proj}, \texttt{v\_proj}, \texttt{o\_proj}, \texttt{gate\_proj}, \texttt{up\_proj}, \texttt{down\_proj}.

\subsection{Evaluation configuration}
\label[appendix]{app:evaluation_config}

GSM8K, GSM1K, and MATH500 are evaluated using greedy decoding, with accuracy as the evaluation metric. AIME 2025 is evaluated using the model-specific sampling configurations detailed in \Cref{app:sampling_hyperparameters}, using avg@64 as the metric. For each problem, correctness is averaged over 64 sampled responses, and the final scores are averaged across the 30 problems.

All reported scores are averaged over three independent evaluation runs. For trained models, this corresponds to one evaluation per training seed. For base models, it corresponds to three repeated evaluations of the same base model. While variations in AIME 2025 naturally stem from sampling, repeated evaluations of base models on the other benchmarks may also differ slightly despite greedy decoding, due to numerical nondeterminism introduced by continuous batching and BF16 arithmetic. \Cref{fig:cliff_dpo_main_results} displays $95\%$ confidence intervals computed over these three runs. \Cref{tab:cliff_dpo,tab:cliff_dpo-controlled} instead report the sample standard error, $s/\sqrt{3}$.

The \textit{Updated tokens} column in \Cref{tab:cliff_dpo,tab:cliff_dpo-controlled} reports the total number of token positions assigned a non-zero loss during fine-tuning. For DPO and cDPO, this count includes all response-token positions supervised by the preference objective. For Cliff-DPO, it includes only the selected cliff-token positions.




\section{Existing assets and licenses}
\label[appendix]{app:assets_licenses}

\begin{table}[H]
    \centering
    \small
    \caption{Existing assets used in this work and their licenses or terms of use.}
    \label{tab:assets_licenses}
    \centering
    \small
    \setlength{\tabcolsep}{4pt}
    \resizebox{\linewidth}{!}{%
    \begin{tabular}{l l l}
        \toprule
        \textbf{Asset} & \textbf{Use in this work} & \textbf{License / Terms} \\
        \midrule
        \texttt{Qwen3-8B} & Evaluation model & Apache-2.0 \\
        \texttt{Qwen3-4B} & Evaluation model & Apache-2.0 \\
        \texttt{Qwen3-0.6B} & Training and evaluation model & Apache-2.0 \\
        \texttt{Llama-3.1-8B-Instruct} & Training and evaluation model & Llama 3.1 Community License \\
        \texttt{Llama-3.2-3B-Instruct} & Evaluation model & Llama 3.2 Community License \\
        \texttt{Llama-3.2-1B-Instruct} & Training and evaluation model & Llama 3.2 Community License \\
        \texttt{Gemma-3-4B-it} & Evaluation model & Gemma Terms of Use \\
        GSM8K & Training and evaluation benchmark & MIT License \\
        GSM1K & Evaluation benchmark & MIT License \\
        MATH500 & Evaluation benchmark & MIT License \\
        AIME 2025 & Evaluation benchmark & MIT License \\
        vLLM & Inference engine & Apache-2.0 \\
        PEFT & Fine-tuning implementation & Apache-2.0 \\
        \bottomrule
    \end{tabular}}
\end{table}

\end{document}